%% file: main.tex
\pdfoutput=1
\documentclass[11pt]{article}

\usepackage{acl}

\usepackage{times}
\usepackage{latexsym}
\usepackage[T1]{fontenc}
\usepackage[utf8]{inputenc}
\usepackage{microtype}

\usepackage{amsmath}
\usepackage{amssymb}
\usepackage{graphicx}
\usepackage{booktabs}
\usepackage{placeins}
\usepackage{stfloats}
\usepackage{enumitem}

\input{figures/styles}

\newcommand{\asilobj}[1]{\textsc{#1}}

\title{ASIL: Replacing Screenshot-and-Click\\with Structured State and Semantic Actions}

\author{Rui Xie$^{1,2}$ \quad Lu Chen$^{1}$ \\
  $^1$X-LANCE Lab, School of Computer Science, Shanghai Jiao Tong University \\
  $^2$State Key Laboratory for General Artificial Intelligence, BIGAI \\
  \texttt{sharryXR@sjtu.edu.cn} \quad \texttt{chenlusz@sjtu.edu.cn} \\
  Project page: \url{https://sharryxr.github.io/ASIL/}}

\begin{document}
\maketitle

\begin{abstract}
Powerful code agents can execute scripts, call tools, and manage files, yet many important applications remain accessible primarily through graphical user interfaces. We argue that screenshot-and-click is an inefficient interface for software-operating agents: screenshots are state-incomplete, and GUI actions are brittle, semantically weak, and poorly matched to long-horizon planning. We introduce ASIL (Agent-Software Interaction Layer), an agent-native interface that exposes software through structured JSON observations and code-executable semantic actions, realized through the deepest feasible access path for each application. We instantiate ASIL across 15 applications and a benchmark of 300 single-application and 80 multi-application tasks. ASIL reaches above 80 with closed models while executing fewer than five actions per task. Under a repaired runtime and a 50-step screenshot budget, the same tasks yield 6.6 and 26.6 strict success under screenshot-and-click control, rising to 15.0 and 53.3 on an easier OSWorld-comparable band. Against application-native interfaces on matched tasks, ASIL exceeds LibreOffice's UNO API by 28--38 strict points but only matches draw.io's MCP content contract. The structured modality also suits training: small-scale SFT raises Qwen3.5-2B from 58.0 to 72.1 and Qwen3.5-9B from 66.6 to 80.4, and resource-limited on-policy RL further raises them to 74.4 and 82.2.
\end{abstract}

\section{Introduction}

Code agents have rapidly evolved from code-completion assistants into execution-oriented software engineering agents. Systems such as Claude Code and Codex operate in terminals, IDEs, CI/CD pipelines, and repository workflows, where they read and modify codebases, run commands and tests, and verify changes through tests or state checks~\cite{sweagent,claudecode,openaicodex}. These agents succeed because they consume machine-readable project context and act through executable tools. In many practical settings, however, the last barrier to fully agentic execution is not text or code reasoning but operation of graphical software: creative tools, productivity suites, desktop utilities, and service-backed applications still expose most of their functionality through GUIs, even when the underlying systems already contain rich internal state, structured artifacts, and executable logic.

\input{figures/method_overview}
\vspace{-6pt}

Current computer-use agents approach this barrier by imitating a human operator: they observe screenshots, infer the rendered interface state, and emit low-level clicks, drags, key presses, and text entries~\cite{osworld,agents,agents2,ufo,cradle,uitars,opencua,cogagent,seeclick,showui}. This screenshot-and-click loop bakes a human-native interface into the core of the agent. The observation side is misaligned because a screenshot is not software state but only its visible projection, omitting hidden panels, background processes, document structure, and internal metadata, and forcing repeated multimodal inference over a presentation layer. The action side is misaligned because GUI events are coordinate-sensitive motor primitives: a single semantic intent (rename a layer, modify a property, trigger an export) expands into long brittle sequences that depend heavily on grounding and break under layout or theme changes. Recent improvements to visual grounding and UI parsing~\cite{setofmark,seeclick,omniparser,showui} still operate within this same low-semantic action space. The central question is therefore not how to make models better at looking at screens, but what observation and action interfaces a software-operating agent should natively use.

We address this mismatch with \textbf{ASIL} (Agent-Software Interaction Layer), an agent-native interface in which the native observation is a structured JSON representation of software state and the native action is a code-executable semantic operation. ASIL is not tied to a single backend: the same agent form is realized through whichever interface most directly exposes stable state and meaningful operations -- structured file formats, native scripting runtimes, or service APIs. We instantiate ASIL across 15 applications, implement a benchmark with 300 single-application and 80 multi-application tasks, and render per-step GUI snapshots from the underlying software state so that ASIL and screenshot-driven runs can be compared under the same task definitions and validators. Figure~\ref{fig:method-overview} summarizes this transition and its realization pathways.

This paper makes five contributions:
\begin{itemize}[topsep=2pt,itemsep=2pt,parsep=0pt,leftmargin=1.4em]
    \item We formulate GUI-agent failure as an interface problem and define ASIL as a software-operating agent form that replaces screenshot observations and GUI events with structured JSON observations and code-executable semantic actions.
    \item We present a realization methodology that maps this form onto heterogeneous software through file, scripting, and service access paths, together with a semi-automatic onboarding framework that compiles reviewed interface profiles into validated adapter contracts.
    \item We provide an implemented 380-task benchmark with shared validators and visual renderings that supports direct ASIL/GUI comparison across 15 applications.
    \item We add repaired GUI / 50max results, matched native-interface baselines against LibreOffice UNO and draw.io MCP, and independent validity checks that bound the comparison.
    \item We show that the same modality is practical for training: small-scale SFT plus resource- and time-limited on-policy RL yield double-digit gains on Qwen3.5-2B and Qwen3.5-9B, and we confirm these findings under a curated 80-task hard suite and a realization-pattern ablation.
\end{itemize}

\section{Related Work}

GUI agents and computer-use benchmarks have demonstrated that large multimodal models can operate real software through human-facing interfaces~\cite{guisurvey,osworld,agents,agents2,ufo,cradle,uitars,opencua,guide}. These systems typically frame software use as a screenshot-centered observation problem and a GUI-event action problem: the agent interprets rendered screens, grounds controls, and emits clicks, keystrokes, drags, or other human-like operations. Recent work also reduces domain bias in GUI agents by retrieving web tutorial videos and injecting plug-and-play planning and grounding annotations~\cite{guide}. Our work shares the goal of making agents operate real software, but studies a different interface question: whether the screenshot-and-click loop should remain the default substrate for software-operating agents.

Several recent systems improve GUI-agent capability by adding scripts, tools, APIs, or compound action layers. OS-Copilot, UFO2, and DynaSaur introduce OS-level skills, hybrid GUI--API control, or dynamically generated programmatic actions~\cite{oscopilot,ufo2,dynasaur}. AXIS and API-based web agents show that routing intent through APIs rather than pixels improves efficiency in their domains, and declarative-interface analyses make the same argument for computer-use agents more broadly~\cite{axis,beyondbrowsing,goi}. CLI-Anything and OpenCLI expose higher-level command or DOM interfaces through agent-facing CLIs and adapters~\cite{clianything,opencli}; application-specific MCPs and native interfaces such as draw.io MCP and LibreOffice UNO provide additional programmatic surfaces. ASIL shares this goal of making software callable, but its evaluated contract couples normalized observations, schema-constrained semantic actions, stable identifiers where exposed, final-state validators, serialized trajectories, and reusable traces/rewards across inference, SFT, and RL. Table~\ref{tab:feature-matrix} in Appendix~\ref{sec:evaluation-details-appendix} summarizes the contract-level distinction, and Section~\ref{sec:native-validity} reports matched baselines rather than claiming universal dominance over application-specific interfaces.

Code agents provide the closest positive precedent. SWE-agent shows that a purpose-built agent-computer interface improves repository navigation, file editing, and test execution~\cite{sweagent}, and CodeAct shows that executable Python can serve as a unified action space for LLM agents~\cite{codeact}. Claude Code and Codex further show that agents become powerful inside environments with readable project state, executable commands, and verifiable feedback~\cite{claudecode,openaicodex}. ASIL transfers this lesson from code repositories to existing GUI software, reconstructing the agent-software interface around each application's file formats, scripting runtimes, service APIs, and verifiable state, and thereby lowering the barrier between agents and the large body of software functionality currently trapped behind human-oriented GUIs.

\section{Interface Mismatch and ASIL}
\label{sec:design}

\subsection{Why Screenshot-and-Click is Inefficient}

The coupling of screenshot observation with low-level GUI events is doubly inefficient. Every step pays for visual encoding and multimodal reasoning before any task-level planning, while the low-level action space expands a single semantic goal into many atomic events. Current OSWorld-Verified results make this pressure visible: many simple tasks are evaluated with budgets above 50 steps and close to 100 steps~\cite{osworld,osworldverifiedweb}. At inference time, high per-step latency multiplied by long trajectories makes even simple desktop tasks slow. At training time, RL rollouts repeatedly pay for screenshot acquisition, model calls, GUI execution, state waiting, and validation across many turns. Long trajectories also lower reliability: GUI actions are highly dependent across steps, so one grounding, focus, layout, or timing error corrupts every state that follows, and recovery requires re-grounding in a changed interface before repair.

\subsection{ASIL Protocol Objects}

ASIL replaces this loop with a structured environment. The agent receives an \asilobj{Observation} object that organizes the current software state into task metadata, application state, interactive elements, environment context, navigation structure, and a concise textual summary; this preserves exactly the information that screenshot-based agents struggle to recover (which document is active, which entities are editable, what background conditions affect correctness). The agent then emits an \asilobj{Action} object whose type, target, and parameters are defined by the application's action schema. Depending on the software, an action may modify a structured file, execute a native script, call a service endpoint, navigate within an internal topology, or trigger a batch operation -- moving the action space from ``click here'' to ``set this value'' or ``invoke this function.'' A single semantic action can encapsulate a long GUI sequence, and its post-action state is checked directly from the next structured observation.

We develop ASIL around four design principles. \emph{Completeness} exposes relevant software state beyond the visible surface. \emph{Semanticity} aligns actions with meaningful software operations rather than GUI mechanics. \emph{Stability} keeps identifiers and targets valid under presentation changes. \emph{Composability} expresses complex tasks through a small number of higher-level actions. Figure~\ref{fig:schema-anatomy} shows the protocol objects and the resulting observe-act-check loop.

Formally, ASIL substitutes the screenshot-and-click interface $(\mathrm{Pix}, \mathcal{M}^k)$ with $(\Phi, \mathcal{A})$ in a single observation--action--transition loop:
\begin{equation}
o_t = \Phi(s_t), \;\; a_t \sim \pi_\theta(\cdot \mid o_t), \;\; s_{t+1} = T(s_t, a_t), \label{eq:asil-loop}
\end{equation}
where $\mathcal{S}$ is the latent software state, $\mathcal{O}$ the structured observation space, and $\mathcal{A}$ the schema-constrained semantic action space. On the observation side, $\mathrm{Pix}$ collapses task-determining attributes outside the rendered viewport (hidden panels, document properties, background state) into indistinguishable equivalence classes, whereas $\Phi$ is engineered to be near-injective on the task-relevant subspace by construction. On the action side, one semantic $a \in \mathcal{A}$ realizes the same transition as a sequence $(m_1, \ldots, m_k) \in \mathcal{M}^k$ with $k \gg 1$, lowering per-step cost and shortening the RL credit-assignment horizon.

\input{figures/schema_anatomy}
\vspace{-6pt}

\section{Realizing ASIL in Real Software}
\label{sec:realization}

\subsection{ASILization Pipeline and Realization Patterns}

ASIL is realized through a semi-automatic ASILization pipeline. For each application, we first identify the deepest feasible access path: the interface that exposes stable state and semantic operations closest to the software's real transition system while remaining practical to run and evaluate. The resulting application adapter implements a shared observe-execute-validate contract, so heterogeneous applications expose the same ASIL interface. The current system uses three recurring realization patterns -- file-backed execution (e.g., SVG, ODF, notebooks), native scripting execution (e.g., Blender Python), and service or API execution (e.g., REST, WebSocket) -- as implementation pathways for the same protocol rather than separate agent types. JSON is the normalized agent-facing representation, not a requirement that software natively stores JSON: reviewed file parsers, scripts/commands, and APIs all map state into the same schema. Across the fifteen implementations, six are file-backed, four native-script, and five service/API realizations.

\paragraph{Eligibility and onboarding.} An application qualifies when it exposes at least one open read path plus a semantic-action path. Coverage is therefore best read at the level of software \emph{functions}: for most common functions there exists at least one qualifying application, even when a particular closed product does not qualify. Functions available only through opaque software with no parseable file, scripting, structured-command, or service surface remain out of scope. The repeatable onboarding work is automated after a reviewed interface profile: one GPT-5.4 call compiled a 97-line Gitea API profile in 24.8 seconds into one observation view and two semantic actions with zero audit errors, then passed 3/3 host and 3/3 Docker probes. Interface discovery, task/evaluator design, application-specific bridges, and GUI synchronization remain human-reviewed; Appendix~\ref{sec:asil-implementation-appendix} separates these stages.

\subsection{Benchmark Construction}

Figure~\ref{fig:benchmark-effect} summarizes the benchmark's application coverage and the observed ASIL-vs-GUI interface effect using representative real screenshots from the evaluated runs.

\input{figures/benchmark_effect}
\vspace{-6pt}

We implement a single benchmark that contains 300 single-application tasks and 80 multi-application tasks. The single-application portion covers 15 software domains with 20 tasks each, including creative tools, productivity applications, service-backed systems, code and workspace environments, and desktop utilities. Representative environments include Inkscape, Blender, GIMP, Audacity, Kdenlive, LibreOffice Calc, Writer, and Impress, together with OBS, Gitea, code-server, JupyterLab, Thunderbird, and Nautilus. The multi-application portion requires information or artifacts to move across application boundaries. This coverage tests whether ASIL is a general software-operating agent form rather than a technique specialized to one application family.

The benchmark is implemented as a shared evaluation system rather than as separate ASIL and GUI task suites. The same task definitions, initial artifacts, state validators, and result directories are used across interface modes. In ASIL mode, the participant receives structured observations and emits JSON semantic actions; in GUI mode, the participant receives screenshots and emits GUI-event actions in real software sessions. ASIL runs also render per-step GUI or page snapshots from the underlying software state, producing visual artifacts that can be inspected alongside screenshot-driven GUI-agent runs. File-level artifact details are given in Appendix~\ref{sec:asil-implementation-appendix}.

The evaluator checks final software state rather than surface-level action histories. Because the same evaluator is used for deterministic execution, ASIL agent runs, GUI runs, SFT filtering, and RL rewards, the benchmark keeps task success criteria software-aware and replayable across all reported settings. ASIL prompts in the main benchmark receive evaluator-derived textual success hints, whereas GUI prompts receive only the task instruction; we therefore do not claim that the original main-table comparison isolates the interface variable alone. Every comparison added in this camera-ready version -- matched native-interface baselines, the visual-supplement ablation, and the hint-off arm of the balanced audit -- is run hint-off on both sides, and Section~\ref{sec:native-validity} quantifies the hint effect.

\subsection{Training Setup under the ASIL Modality}

The original motivation for ASIL also suggests a practical training direction: structured observations, semantic actions, and evaluator traces are easier to serialize, verify, and reuse than screenshot-and-click traces. We therefore run supervised and reinforcement-learning studies under the ASIL modality, using the same evaluator-backed runtime as the benchmark. Figure~\ref{fig:training-pipeline} summarizes this pipeline as a single closed loop: verified ASIL trajectories supply both the replay-based SFT datasets and the on-policy rollout data, the policy is updated against rewards produced by the same evaluator used at inference time, and the resulting checkpoints are then scored on the same 380-task benchmark, with the 9B family's training deltas amplifying by an additional 3--4 points on the hard suite of Section~\ref{sec:hard-results}.

\input{figures/training_pipeline}
\vspace{-6pt}

The setup has three components: a low-overlap training task pool generated separately from the evaluation benchmark (a learnable 320/80 RL subset is drawn from a final 512/128 v3 pool covering all 15 domains); step-level SFT traces from known-correct ASIL actions and verified GPT-5.4 rollouts; and evaluator-backed on-policy RL through an external ASIL AgentService, so rewards and rollouts share the same observations, schemas, adapters, and validators as inference-time evaluation.

Appendices~\ref{sec:training-details-appendix} and \ref{sec:taskgen-appendix} give the full SFT/RL settings and the task-generation overlap audit.

\section{Experiments}
\label{sec:results}

\subsection{Main Benchmark}

Table~\ref{tab:main-results} reports the main 380-task benchmark. The single-application columns group the 15 software environments (20 tasks each) by domain region, and Multi-App is reported as a peer region; Overall aggregates all 380 tasks. ASIL and GUI share the same task definitions, initial states, and validators. ASIL uses the default 15-step budget and averages fewer than five executed actions; the repaired GUI rows use a 50-step native-computer-use budget, with one additional row truncating the same sonnet4.6 trajectories at 15 actions.

\begin{table*}[t]
\centering
\small
\setlength{\tabcolsep}{4pt}
\renewcommand{\arraystretch}{1.0}
\resizebox{\textwidth}{!}{%
\begin{tabular}{@{}llcccccc@{}}
\toprule
Model setting & Mode & Creative (120) & Office (80) & Developer (60) & Files (40) & Multi-App (80) & Overall (380) \\
\midrule
GPT-5.4 & ASIL / 15max & 75.8 & \textbf{85.0} & 91.7 & 92.5 & 73.8 & 81.6 \\
 & GUI / 50max$^{\dagger}$ & 11.7 & 1.2 & 6.7 & 12.5 & 1.2 & 6.6 \\
Qwen3.6-plus & ASIL / 15max & \underline{82.5} & 78.8 & 91.7 & 87.5 & 70.0 & 81.1 \\
Kimi K2.5 & ASIL / 15max & \textbf{84.4} & \underline{79.5} & \textbf{95.6} & 86.2 & 81.7 & \textbf{84.8} \\
sonnet4.6 & ASIL / 15max & 81.2 & 74.3 & 89.8 & 94.4 & 75.1 & 81.2 \\
 & GUI / 50max$^{\dagger}$ & 29.2 & 7.5 & 51.7 & 72.5 & 0.0 & 26.6 \\
 & GUI / 15max$^{\dagger}$ & 23.3 & 6.2 & 36.7 & 32.5 & 0.0 & 17.9 \\
\midrule
Qwen3.5-27B & ASIL / 15max & 79.1 & 72.7 & 92.1 & \underline{98.3} & 50.2 & 75.8 \\
Qwen3.5-2B Base & ASIL / 15max & 49.5 & 48.7 & 68.4 & 52.7 & 75.1 & 58.0 \\
Qwen3.5-2B SFT & ASIL / 15max & 67.8 (+18.3) & 68.9 (+20.2) & 70.6 (+2.2) & 92.5 (+39.8) & 72.5 (-2.6) & 72.1 (+14.1) \\
Qwen3.5-2B RL & ASIL / 15max & 69.7 (+20.2) & 69.1 (+20.4) & 79.8 (+11.4) & 77.7 (+25.0) & 81.0 (+5.9) & 74.4 (+16.4) \\
Qwen3.5-9B Base & ASIL / 15max & 67.5 & 48.3 & 87.3 & 97.5 & 52.7 & 66.6 \\
Qwen3.5-9B SFT & ASIL / 15max & 72.1 (+4.6) & 75.5 (+27.2) & 91.4 (+4.1) & 83.8 (-13.8) & \underline{87.9} (+35.2) & 80.4 (+13.8) \\
Qwen3.5-9B RL & ASIL / 15max & 72.2 (+4.7) & 73.5 (+25.3) & \underline{92.6} (+5.3) & \textbf{98.8} (+1.2) & \textbf{89.6} (+36.9) & \underline{82.2} (+15.5) \\
\bottomrule
\end{tabular}%
}
\caption{Main 380-task benchmark results. Column heads show task counts; full region membership is in Appendix~\ref{sec:evaluation-details-appendix}. $^{\dagger}$GUI / 50max rows are this round's repaired native-computer-use re-test; GUI / 15max truncates the same trajectories. Parentheses are point changes over the corresponding ASIL base model; bold and underline mark the best and second-best scores per column.}
\label{tab:main-results}
\end{table*}
\vspace{-5pt}

The gap remains large after repair and after tripling the GUI budget. GPT-5.4 reaches 81.6 under ASIL versus 6.6 under GUI; sonnet4.6 reaches 81.2 versus 26.6, and 17.9 when the same GUI trajectories are restricted to ASIL's 15-step budget. The repaired GUI rows also differ sharply from each other, so the GUI side is not uniformly weak; a strong native computer-use model can recover many single-application tasks. The aggregate still supports the central claim: structured observations and semantic actions make software operation shorter, more stable, and more verifiable than screenshot-and-click control. Figure~\ref{fig:success-case} in Appendix~\ref{sec:evaluation-details-appendix} illustrates this failure mode on a LibreOffice task, where GUI control spends its budget in a repeated keypress paste loop while ASIL writes the spreadsheet state with one \texttt{modify\_file} action.

\subsection{Repaired GUI Bands, Native Baselines, and Validity}
\label{sec:native-validity}

A single GUI aggregate hides strong dependence on task shape. On \emph{easy60}, a single-application band drawn from the same 380 tasks and calibrated by structural proxies against OSWorld-369, the repaired GUI / 50max rows rise from 6.6 to 15.0 strict for GPT-5.4 and from 26.6 to 53.3 for sonnet4.6 (mean scores 18.3 and 54.2). We treat this as an OSWorld-comparable reference band, not an exact difficulty match; the full benchmark is deliberately harder because it includes complete workflows with no pre-staged intermediate progress.

\begin{table}[t]
\centering
\footnotesize
\setlength{\tabcolsep}{3pt}
\renewcommand{\arraystretch}{1.05}
\resizebox{\columnwidth}{!}{%
\begin{tabular}{@{}llrr@{}}
\toprule
Setting & Interface & Strict & Mean \\
\midrule
GPT-5.4 full 380 & GUI / 50max & 6.6 & 11.6 \\
sonnet4.6 full 380 & GUI / 50max & 26.6 & 30.3 \\
GPT-5.4 easy60 & GUI / 50max & 15.0 & 18.3 \\
sonnet4.6 easy60 & GUI / 50max & \textbf{53.3} & \textbf{54.2} \\
\midrule
GPT-5.4 LibreOffice & ASIL / UNO & 95.0 / 66.7 & 97.5 / 72.8 \\
sonnet4.6 LibreOffice & ASIL / UNO & 98.3 / 60.0 & 99.5 / 69.2 \\
GPT-5.4 draw.io & ASIL / MCP & 55.0 / 55.0 & 80.9 / 80.7 \\
sonnet4.6 draw.io & ASIL / MCP & 25.0 / 55.0 & 46.7 / 81.9 \\
\bottomrule
\end{tabular}%
}
\caption{Camera-ready calibration results (\%). Top: repaired GUI / 50max by task band. Bottom: matched native-interface baselines, run on identical tasks with the same evaluator, model, 15-step budget, and no success hint.}
\label{tab:camera-ready-audits}
\end{table}

Some applications also expose native programmatic interfaces, so we compare against those directly in Table~\ref{tab:camera-ready-audits}. On 60 LibreOffice tasks, ASIL exceeds native UNO by 28--38 strict points even though UNO exposes the full automation surface, indicating that a normalized observation and contracted action layer can be easier for agents to use than a low-level native API. On 20 draw.io tasks, ASIL is level with draw.io's MCP content contract for GPT-5.4 and behind it for sonnet4.6. The claim is therefore compositional rather than per-application dominance: ASIL contributes one cross-application contract that couples observation, action, validation, and reusable trajectories. Where a mature native agent interface already exists, the right use of ASIL is to ingest it as an access path.

We also check measurement validity. A fail-closed raw validator that does not call the ASIL observation builder agrees with the official evaluator on 60/60 balanced final states. The prompt success hint is measurable: on balanced-30, GPT-5.4 scores 29/30 strict with evaluator hints and 26/30 without them. A GIMP visual-supplement ablation (44 tasks, no hint, 50 steps) gives 44.1 mean for ASIL-only and 43.7 with per-step screenshots, suggesting the current bottleneck is perceptual action vocabulary rather than missing screenshot observations.

\subsection{Training Results under ASIL}
\label{sec:training-results}

The open-model rows provide a second result: ASIL is not only a stronger inference interface, but also a more sample- and compute-efficient training modality. Empirically, GUI-agent training is comparatively harder and more expensive because screenshot observations and low-level GUI actions make each rollout pay for repeated visual processing, long action horizons, and brittle state recovery. Under ASIL, Qwen3.5-2B improves from 58.0 to 72.1 with SFT and to 74.4 with resource- and time-limited on-policy RL. Qwen3.5-9B improves from 66.6 to 80.4 with SFT and to 82.2 with RL. These double-digit gains are obtained from a deliberately small training setup: the final SFT stage uses only thousands of step-level ASIL samples, and the final RL runs use 320 training tasks and 80 validation tasks with 4 A800 GPUs for 2B and 8 A800 GPUs for 9B. Appendix~\ref{sec:training-details-appendix} gives the full training settings.

The gains are meaningful but not uniform. For Qwen3.5-2B, SFT improves Files \& Communication by 39.8 points but slightly trails the base model on Multi-App; RL recovers Multi-App and reaches the best 2B overall score. For Qwen3.5-9B, SFT contributes the largest improvements in Multi-App (+35.2) and Office \& Diagrams (+27.2), while RL further improves the overall score to 82.2 and reaches the best Multi-App score among all reported rows. We therefore read the training results as evidence for the training advantage predicted by the interface design: small-scale SFT and RL become effective under ASIL with much less rollout burden than screenshot-and-click training would require.

\subsection{Hard-Task Evaluation}
\label{sec:hard-results}

We further evaluate the selected checkpoints on a held-out 80-task hard suite emphasizing long-horizon, multi-constraint, deliverable-grade workflows: real-photo editing and annotation in GIMP, cross-application artifact transfer, office and diagram deliverables, developer and service workflows, and file/email operations. Each task requires 6 to 15 meaningful semantic actions. The suite is ASIL-only because Table~\ref{tab:main-results} already shows repaired GUI / 50max control far below ASIL on the easier 380-task benchmark, and preliminary GPT-5.4 GUI trials confirmed the same collapse on the hard suite. Appendix~\ref{sec:evaluation-details-appendix} gives the task composition and evaluation setup.

\begin{table*}[t]
\centering
\small
\setlength{\tabcolsep}{4pt}
\renewcommand{\arraystretch}{1.0}
\resizebox{\textwidth}{!}{%
\begin{tabular}{@{}llcccccc@{}}
\toprule
Model setting & Mode & Creative (24) & Office (12) & Developer (8) & Files (6) & Multi-App (30) & Overall (80) \\
\midrule
GPT-5.4 & ASIL / 15max & \underline{14.7} & \textbf{85.9} & \textbf{91.7} & \textbf{100.0} & \underline{73.9} & \textbf{61.7} \\
Qwen3.5-2B Base & ASIL / 15max & 13.3 & 40.8 & 41.1 & 20.8 & 72.8 & 43.1 \\
Qwen3.5-2B SFT & ASIL / 15max & 8.0 (-5.3) & 50.3 (+9.5) & 66.5 (+25.4) & \textbf{100.0} (+79.2) & 66.7 (-6.1) & 49.1 (+6.0) \\
Qwen3.5-2B RL & ASIL / 15max & 10.1 (-3.2) & 54.2 (+13.4) & 64.0 (+22.9) & \underline{83.3} (+62.5) & 61.1 (-11.7) & 46.7 (+3.6) \\
Qwen3.5-9B Base & ASIL / 15max & 12.4 & 28.3 & \underline{84.4} & \textbf{100.0} & 33.9 & 36.6 \\
Qwen3.5-9B SFT & ASIL / 15max & 14.5 (+2.1) & \underline{57.5} (+29.2) & 80.3 (-4.1) & 66.7 (-33.3) & \underline{73.9} (+40.0) & 53.7 (+17.1) \\
Qwen3.5-9B RL & ASIL / 15max & \textbf{15.2} (+2.8) & 41.4 (+13.1) & \textbf{91.7} (+7.3) & \textbf{100.0} (+0.0) & \textbf{77.2} (+43.3) & \underline{56.4} (+19.8) \\
\bottomrule
\end{tabular}%
}
\caption{Held-out 80-task hard-suite results under ASIL / 15max mode. Region breakdown (24 GIMP, 12 Draw.io+LibreOffice, 8 code-server+Gitea+JupyterLab, 6 Nautilus+Thunderbird, 30 multi-app) and full setup in Appendix~\ref{sec:evaluation-details-appendix}; parentheses are point changes over the corresponding ASIL base model.}
\label{tab:hard-results}
\end{table*}
\vspace{-5pt}

The hard suite sharpens the training picture. For Qwen3.5-9B, the SFT and RL gains over the base model increase to +17.1 and +19.8 points, larger than the corresponding +13.8 and +15.5 gains on the main benchmark. The strongest improvement appears in Multi-App, which rises from 33.9 to 73.9 to 77.2 across Base, SFT, and RL. This is the regime where semantic actions should matter most: long workflows make low-level GUI traces fragile, while ASIL exposes higher-level operations and evaluator-backed rewards.

For Qwen3.5-2B, the picture is more limited. SFT improves the base model by +6.0 on the hard suite, while the selected RL checkpoint reaches only +3.6 and trails the 2B SFT checkpoint. This indicates that the 2B RL operating point that is best on the broader benchmark does not transfer uniformly to the difficulty tail. Creative real-photo editing remains the main bottleneck for all rows, suggesting a residual access-path and data-coverage limitation rather than a pure interface failure.

\subsection{Realization-Pattern Ablation}
\label{sec:pattern-ablation}

Finally, we test whether the result depends on a single adapter style by probing the deepest-feasible access principle of Section~\ref{sec:realization}. Table~\ref{tab:pattern-ablation} reports two probes: a cross-application pattern audit over the 300 single-application tasks, and a within-application path-restriction study on LibreOffice.

\begin{table*}[t]
\centering
\small
\renewcommand{\arraystretch}{1.0}

\resizebox{\textwidth}{!}{%
\setlength{\tabcolsep}{4pt}
\begin{tabular}{@{}llccc@{\hspace{1.6em}}c@{\hspace{1.6em}}lcccc@{}}
\toprule
\multicolumn{5}{c}{\textbf{(a) Cross-application pattern audit}} & & \multicolumn{5}{c}{\textbf{(b) LibreOffice path-restriction study}} \\
\midrule
Participant & Mode & A (file) & B (script) & C (api) & & Participant / realization & Calc & Writer & Impress & Overall \\
\midrule
GPT-5.4 & ASIL / 15max & \textbf{91.2} & \textbf{89.5} & \textbf{95.1} & & GPT-5.4 / file & \textbf{100.0} & \underline{98.4} & \underline{97.4} & \textbf{98.6} \\
GPT-5.4 & GUI / 50max & 1.7 & 5.0 & 18.0 & & GPT-5.4 / script-dispatch & \textbf{100.0} & \underline{98.4} & \underline{97.4} & \textbf{98.6} \\
sonnet4.6 & GUI / 50max & 10.8 & 21.2 & 71.0 & & Qwen3.5-9B RL / file & 35.6 & \textbf{100.0} & \textbf{100.0} & 78.5 \\
Qwen3.5-9B Base & ASIL / 15max & 53.0 & \underline{78.8} & 84.4 & & Qwen3.5-9B RL / script-dispatch & \underline{40.6} & \textbf{100.0} & 99.0 & \underline{79.9} \\
Qwen3.5-9B RL & ASIL / 15max & \underline{71.8} & 77.9 & \underline{92.1} & & & & & & \\
\bottomrule
\end{tabular}%
}

\caption{Realization-pattern ablation. Panel (a) groups single-application tasks by Pattern A file-backed, Pattern B native-scripting, and Pattern C service/API realizations; GUI / 50max rows are recomputed from this round's repaired re-test. Panel (b) compares file-backed and script-dispatch access paths on the same 60 LibreOffice tasks.}
\label{tab:pattern-ablation}
\end{table*}
\vspace{-5pt}

The interface effect is large in every realization pattern: for GPT-5.4, ASIL beats repaired GUI / 50max by 89.5, 84.5, and 77.1 points in the file-backed, scripting, and service/API buckets. The sonnet4.6 GUI row shows that service/API tasks are the easiest band for screenshot control, but still below the ASIL row. Training gains are more pattern-dependent. Qwen3.5-9B RL improves most in the file-backed bucket (+18.8) and in service/API (+7.7), but is nearly unchanged in native scripting ($-0.9$), where the base model is already strong. We therefore do not claim that RL improves every access pattern equally.

The LibreOffice path-restriction study separates realization pattern from application identity. GPT-5.4 reaches exactly 98.6 under both the file-backed and script-dispatch paths, and Qwen3.5-9B RL differs by only 1.3 points. Since GUI control is near zero on the same LibreOffice family in Table~\ref{tab:main-results}, both deep access paths sit far above the GUI baseline. The evidence supports the deepest-feasible access principle without implying that one backend pattern is universally optimal.

\subsection{Failure Analysis}
\label{sec:failure-analysis}

The remaining failures fall into two classes. GUI runs mainly fail from missing hidden state, grounding errors, and brittle event sequences. These errors explain the low repaired GUI rows in Table~\ref{tab:main-results} and the GPT-5.4 GUI / 50max scores of 1.7, 5.0, and 18.0 across realization patterns in Table~\ref{tab:pattern-ablation}. ASIL removes this class by changing the interface.

ASIL runs instead fail from residual modeling and coverage limits. The clearest example is Creative real-photo editing in the hard suite: even the strongest row averages only 15.2 on GIMP-heavy tasks, because some target conditions depend on pixel-level image semantics that the current semantic action vocabulary does not fully cover. Other failures come from incorrect semantic-action choice or from underrepresented application families in the training traces, and the Thunderbird regression shows that supervised training can also disturb a behavior pattern the base model already handles. Figure~\ref{fig:failure-cases} in Appendix~\ref{sec:evaluation-details-appendix} illustrates these three failure modes. The practical implication is that after the interface bottleneck is removed, progress should come from richer adapters, broader trace coverage, and stronger training, not from further optimizing screenshot-and-click control.

\section{Conclusion}

This paper argues that the central obstacle in operating GUI software is not task difficulty but an interface mismatch between human-native screenshot-and-click and agent-native semantic operation. ASIL replaces this loop with structured software state and code-executable semantic actions, realized through the deepest feasible access path for each application -- a more software-native agent form, not an extra tool layer.

On a 380-task benchmark across 15 applications, ASIL obtains strong inference results with short semantic trajectories, and the same modality enables practical training: small-scale SFT and on-policy RL yield double-digit gains on Qwen3.5-2B and Qwen3.5-9B. The comparative claim is deliberately bounded. A repaired 50-step GUI re-test reaches 6.6 and 26.6 strict success on the full benchmark and 15.0 and 53.3 on an easier OSWorld-comparable band; against native interfaces, ASIL clearly exceeds LibreOffice UNO but only matches draw.io MCP. Software operation is best treated as interaction over state and verifiable artifacts, not over pixels and motor traces.

\clearpage

\section*{Limitations}

We discuss four limitations tied to the paper's core claims: prompt asymmetry in the original comparison, small-model RL stability on long-horizon tasks, realization coverage gaps for fully opaque applications, and tasks whose success criteria are intrinsically perceptual.

\paragraph{Prompt asymmetry and evaluator reuse.} The submitted main ASIL prompts include evaluator-derived success hints, while GUI prompts do not. The hint has a measurable effect (29/30 versus 26/30 strict on balanced-30), and the same evaluator also filters SFT data and supplies RL rewards. We therefore do not claim that the original main table isolates only the interface variable. The camera-ready baselines added in Section~\ref{sec:native-validity} use hint-off comparisons and an independent raw-state validator, but broader independent validation across all 380 tasks remains future work.

\paragraph{2B training stability on long-horizon tasks.} The Qwen3.5-2B training gradient amplifies cleanly on the main 380-task benchmark (+14.1 and +16.4 points for SFT and RL over the base model) but only partially on the 80-task hard suite, where the gains shrink to +6.0 and +3.6 and the selected 2B RL checkpoint falls 2.4 points behind the 2B SFT checkpoint. We read this as the small-model RL operating point rather than an interface problem: \texttt{global\_step\_8} maximizes the aggregate ASIL-380 score but does not consistently improve long-horizon behavior on Multi-App and Creative. Finding a 2B RL setting that retains the main-benchmark gain on long-horizon tasks -- for example, a longer schedule, a hard-task-weighted curriculum, or a different checkpoint aggregation -- remains future work. The Qwen3.5-9B family does not show this pattern.

\paragraph{Coverage gaps for fully opaque applications.} ASIL is realized through the semi-automatic ASILization pipeline of Section~\ref{sec:realization}, which substantially lowers per-application setup cost. The main remaining limitation is extending ASIL to applications that are simultaneously \emph{closed-source}, use \emph{file formats that are hard to unpack}, and \emph{lack rich external scripting or service interfaces}: the three realization patterns (file, scripting, service) all require at least one such access door to be open, so applications with all three closed cannot expose a deep feasible access path under our current methodology. The 15-application instantiation reported here focuses on software where at least one access door is open; extending ASIL to fully opaque applications, which would require fundamentally different access strategies, is left to future work.

\paragraph{Residual perceptual tasks.} ASIL exposes structured software state, but some task goals are intrinsically perceptual. The clearest example is the Creative region of the hard suite, where GIMP real-photo editing requires judgments about visual composition that the current semantic action vocabulary cannot fully express -- even GPT-5.4 ASIL averages only 14.7 on this region and obtains zero full passes across the 24 tasks. A naive visual supplement does not fix this alone (44.1 mean ASIL-only versus 43.7 with screenshots on 44 no-hint GIMP tasks), suggesting that richer perceptual action primitives are also needed. ASIL is most valuable when success criteria are checkable against structured state; tasks dominated by aesthetic or perceptual criteria need a hybrid interface.

\section*{Ethical Considerations}

ASIL is an interface and evaluation framework rather than a user-facing autonomous deployment. The benchmark tasks are synthetic software-operation tasks and do not contain personally identifying or offensive content. The main risks are misuse of more capable software-operating agents, overclaiming generality for software without open access paths, and hidden evaluator bias if downstream users treat ASIL scores as universal GUI capability. We mitigate these risks by releasing task definitions, validators, adapter code, training data, and failure summaries; by reporting repaired GUI and native-interface baselines; and by scoping claims to software with open file, scripting, structured-command, or service access paths. Deployments that connect ASIL-style agents to real accounts or files should add permission boundaries, audit logging, confirmation gates for destructive actions, and application-specific safety policies.

\section*{Acknowledgments}

We thank the anonymous reviewers and area chairs for their constructive feedback. This work is funded by the China NSFC Projects (92370206, 62120106006, U23B2057, 62576212) and Frontier Technologies R\&D Program of Jiangsu (BF2025029).

\bibliography{refs}

\newpage

\appendix

\section{Evaluation Details}
\label{sec:evaluation-details-appendix}

\paragraph{Main benchmark regions.} The main benchmark contains 300 single-application tasks and 80 multi-application tasks. The 300 single-application tasks are grouped into four regions in Table~\ref{tab:main-results}: Creative Media contains Audacity, Blender, GIMP, Inkscape, Kdenlive, and OBS; Office \& Diagrams contains Draw.io, LibreOffice Calc, LibreOffice Impress, and LibreOffice Writer; Developer Workflows contains code-server, Gitea, and JupyterLab; Files \& Communication contains Nautilus and Thunderbird. Each single-application environment contributes 20 tasks. Multi-App is reported separately because these tasks require information or artifacts to move across application boundaries. The default paired budget is 15 interaction steps for both ASIL and GUI. This is a middle-ground setting: ASIL generally needs far fewer steps, whereas GUI-agent evaluations often use 50-step budgets or larger. For complex tasks whose GUI execution would otherwise have too little room for retries, the GUI run is allowed up to 50 steps; these exceptions give GUI extra recovery opportunities. Across reported ASIL runs, the average executed trajectory length remains below five steps.

\paragraph{Hard-task suite.} The held-out hard suite contains 80 ASIL-only tasks designed to overweight long-horizon, multi-constraint, deliverable-grade workflows. It contains 24 GIMP real-photo editing and annotation tasks, 30 multi-application workflows, 12 office and diagram tasks, 8 developer tasks, and 6 files and communication tasks. Each task is 6 to 15 meaningful semantic actions long and is checked against final artifacts and visible application state. The suite is held out from the main 380-task benchmark and passes deterministic ASIL execution before agentic evaluation. We do not report formal GUI rows on this suite because repaired GUI / 50max control already remains far below ASIL on the easier main benchmark. We also ran preliminary GPT-5.4 GUI trials on the hard suite and observed the same collapse, so a full GUI sweep would mostly add expensive low-score rows rather than change the interpretation.

\paragraph{Interface-effect case study.} Figure~\ref{fig:success-case} grounds the aggregate interface effect of Section~\ref{sec:results} in one concrete LibreOffice payroll task from the main benchmark. The figure is retained as a qualitative failure-mode illustration from the submitted paired run: GPT-5.4 GUI control exhausts 15 keypress events in a repeated literal-tab paste loop, whereas GPT-5.4 ASIL uses a single \texttt{modify\_file} semantic action to satisfy all six evaluator checkpoints.

\input{figures/success_case}

\paragraph{Failure case studies.} Figure~\ref{fig:failure-cases} visualizes the three residual failure modes named in Section~\ref{sec:failure-analysis} using representative tasks from the hard suite: an access-path limit on GIMP real-photo editing, a data-coverage gap on LibreOffice Calc multi-cell edits, and a Qwen3.5-9B SFT regression on Thunderbird that RL later recovers. Each panel pairs a task instruction with a canonical thought/action sketch and the aggregate pass counts observed in the seven-model hard-suite evaluation.

\input{figures/failure_cases}

\paragraph{Evaluator design.} The evaluator is a shared component used identically across deterministic execution, ASIL agent runs, GUI agent runs, SFT trace filtering, and RL reward computation. Each task defines one or more \emph{success paths}, where each path is an ordered conjunction of typed \emph{checkpoints} (e.g., spreadsheet cell value, file existence, regex match on document text, XPath structure on SVG, image pixel statistics, REST resource state). At evaluation time the evaluator inspects the current software state through the same adapter the agent uses, scores each checkpoint independently with a typed comparator, and returns the maximum path score in $[0, 1]$ as the task's continuous score; binary success requires every required checkpoint along the matched path to pass. Because the same evaluator is invoked for SFT filtering, RL reward, and benchmark scoring, training signal is identical to evaluation signal by construction; Section~\ref{sec:native-validity} reports an independent raw-state check on a balanced subset.

\paragraph{Realization-pattern audit.} Pattern A is file-backed and covers Inkscape, LibreOffice Calc, LibreOffice Writer, LibreOffice Impress, Draw.io, and JupyterLab. Pattern B is native-scripting and covers Blender, GIMP, Kdenlive, and Audacity. Pattern C is service or API based and covers OBS, Gitea, code-server, Thunderbird, and Nautilus. The audit in Table~\ref{tab:pattern-ablation} excludes multi-application tasks because they can traverse multiple patterns within one task.

\paragraph{LibreOffice path-restriction study.} The default LibreOffice path uses file-backed ODF and spreadsheet edits. The script-dispatch variant routes the same semantic actions through a generated Python script process while reusing the same observation builder, task set, and evaluator. Both variants are evaluated on the same 60 LibreOffice tasks used by the main benchmark.

\begin{table*}[t]
\centering
\small
\setlength{\tabcolsep}{4pt}
\renewcommand{\arraystretch}{1.04}
\resizebox{\textwidth}{!}{%
\begin{tabular}{@{}lccccc@{}}
\toprule
Dimension & ASIL & CLI-Anything & OpenCLI & draw.io MCP & LibreOffice UNO/CLI \\
\midrule
Access path & file+script+API & generated CLIs & browser/DOM & draw.io only & headless/UNO \\
Structured observation & typed JSON & command output & DOM snapshot & app-specific diagram & file/UNO objects \\
Semantic actions & typed vocabulary & CLI subcommands & browser commands & diagram tools & native API calls \\
Stable IDs & yes & no & partial & partial & partial \\
Final-state validator & yes & no & no & no & no \\
Cross-app evaluation & 15 apps & no & no & no & no \\
Trajectory reuse for SFT/RL & yes & no & no & no & no \\
Onboarding & semi-auto gen+audit & harness generation & adapter primitives & ready interface & native interface \\
\bottomrule
\end{tabular}%
}
\caption{Contract-level comparison with nearby agent-native or programmatic software interfaces. The table compares documented interface dimensions, not task performance; matched task results for LibreOffice UNO and draw.io MCP appear in Table~\ref{tab:camera-ready-audits}.}
\label{tab:feature-matrix}
\end{table*}

\section{ASIL Adapter and Trace Implementation Details}
\label{sec:asil-implementation-appendix}

Figure~\ref{fig:realization-patterns} illustrates the three realization pathways as implementations of one shared protocol and shows the normalized trace artifacts they produce.

\input{figures/realization_patterns}

\begin{table}[t]
\centering
\footnotesize
\setlength{\tabcolsep}{4pt}
\renewcommand{\arraystretch}{1.05}
\resizebox{\columnwidth}{!}{%
\begin{tabular}{@{}lc@{}}
\toprule
Onboarding stage & Automated \\
\midrule
Access-path discovery, profile authoring/review & no \\
Interface plan (typed observation/action) & yes \\
Observation schema and action schema & yes \\
Adapter wrapper and extension bundle & yes \\
Direct vs.\ bridge-assisted classification & yes \\
Static, host, and Docker validation gates & yes \\
Application-specific bridges & no \\
GUI synchronization and rendering & no \\
Task and evaluator design & no \\
\bottomrule
\end{tabular}%
}
\caption{Manual versus automated stages of the ASILization pipeline. The automated block is the repeated interface-to-contract work; the manual stages are reviewed setup and evaluation work.}
\label{tab:onboarding-split}
\end{table}

\paragraph{Adapter contract.} Each ASIL adapter implements three required methods: \texttt{observe()} extracts the current structured state, \texttt{execute(action)} applies one semantic action and returns the next observation, and \texttt{validate\_action(action)} checks whether an action is valid for the application. The shared adapter base also supports optional hooks for source cloning, task context, real-GUI launch specifications, GUI-to-canonical-state synchronization, and rendering. These hooks allow the same application backend to support ASIL execution, GUI comparison, evaluator replay, and visual inspection without changing the agent-facing protocol.

\paragraph{Observation schema.} Each ASIL observation is normalized into a typed JSON object with six main fields: metadata, application state, interactive elements, environment context, navigation structure, and a textual data summary. Metadata records the application and observation source, such as file parsing, native script output, REST state, or DOM snapshots. Application state records the current view, active document, and document path. Interactive elements carry stable IDs, type, label, value, editability, data type, constraints, available actions, child links, and metadata. Environment and navigation fields expose background conditions and reachable views that are usually absent from a screenshot.

\paragraph{Representative adapters.} The concrete parsers are application-specific but follow the same normalization rule. The Inkscape adapter parses SVG XML with namespace-aware XPath, exposes shapes, text, images, groups, and layers as elements, and preserves geometry, style, transform, parent, and canvas metadata. Its actions mutate SVG nodes or attributes and then write the file back. The LibreOffice Calc adapter reads \texttt{content.xml} inside an ODS archive, exposes cells using stable IDs such as \texttt{Sheet1!A1}, records value types and formulas, and executes batched cell edits by updating the underlying ODF XML while extending rows or columns when needed. The Blender adapter generates Python code that runs inside Blender to dump scene objects, transforms, materials, modifiers, animation data, render settings, and timeline settings; actions are realized as generated \texttt{bpy} scripts. Service-backed adapters follow the same pattern through APIs: for example, the Gitea adapter exposes repositories, issues, pull requests, milestones, and labels with stable resource IDs and executes REST calls against the corresponding endpoints.

\paragraph{Action schemas and traces.} An ASIL action contains an \texttt{action\_type}, a \texttt{target}, and a \texttt{params} object, with supported types including \texttt{set\_value}, \texttt{invoke\_function}, \texttt{modify\_file}, \texttt{api\_call}, \texttt{navigate}, and \texttt{batch}. Each software environment has a JSON action schema that specifies allowed action types, target format, parameter schema, examples, a \texttt{done} action, and software-specific tips. Agentic ASIL runs record a \texttt{traj.jsonl} file with per-step observations, thoughts, actions, execution status, latency fields, render metadata, and evaluator scores. GUI runs write the same result structure with screenshot observations and GUI actions. Final scores are written to \texttt{result.txt}, while per-step visual artifacts are stored as \texttt{step\_N.png} and \texttt{step\_N.render.json}.

\section{Training Details under the ASIL Modality}
\label{sec:training-details-appendix}

\subsection{SFT Trajectory Sources}
\label{sec:sft-data}

We construct two step-level SFT datasets. SFT-v0 replays the known-correct ASIL actions stored in each task definition: for every task, the pipeline re-executes the annotated correct actions in the ASIL environment and collects per-step (observation, action, post-action observation) tuples, while the teacher model only annotates a short thought before each predetermined action. The supervision target is therefore the (thought, action) pair, but the action itself is not model-generated. The agentic-guided-v2 set instead records verified GPT-5.4 ASIL rollouts in which the model produces its own (thought, action) at each step. For tasks that failed under earlier rollout rounds, the agentic-guided-v2 generator also runs guided recovery using either a compact action hint that names the next operation or a detailed action hint that provides the full action JSON; only evaluator-verified steps with valid action JSON and zero execution errors are written into the dataset. For the final 9B SFT run, v0 and agentic-guided-v2 are merged and exactly deduplicated, producing the merged split summarized at the bottom of Table~\ref{tab:sft-data}.

\begin{table}[!htbp]
\centering
\small
\setlength{\tabcolsep}{4pt}
\renewcommand{\arraystretch}{1.15}
\begin{tabular}{@{}lrrrr@{}}
\toprule
Split & Rows & Tasks & Apps & Hints \\
\midrule
v0 train & 556 & 512 & 15 & 0 \\
v0 valid & 138 & 128 & 15 & 0 \\
\addlinespace[2pt]
guided-v2 train & 2{,}330 & 500 & 15 & 148 \\
guided-v2 valid & 594 & 128 & 15 & 60 \\
\addlinespace[2pt]
merged train (9B) & 2{,}886 & 512 & 15 & 148 \\
merged valid (9B) & 732 & 128 & 15 & 60 \\
\bottomrule
\end{tabular}
\caption{SFT data construction. All retained rows have valid action JSON, zero execution or annotation errors, and non-empty thought fields; the merged 9B split is the exact deduplication of v0 and agentic-guided-v2.}
\label{tab:sft-data}
\end{table}

\subsection{SFT Hyperparameters and Selected Checkpoints}
\label{sec:sft-hparams}

Both 2B and 9B runs use FSDP, bfloat16 precision, gradient checkpointing, AdamW with $\beta = (0.9, 0.95)$ and weight decay $0.01$, warmup ratio $0.1$, gradient clipping at $1.0$, and left truncation. Inputs are stored as ASIL chat-format \texttt{prompt\_messages} and supervised targets are the corresponding (thought + action) assistant response. The final 2B SFT continues from a short v0-pretrained checkpoint for three additional epochs on the agentic-guided-v2 split (111 total steps), and the final 9B SFT trains for six epochs on the merged v0+v2 deduplicated split (1{,}086 total steps). For both models the selected checkpoint is chosen by downstream ASIL-380 benchmark performance rather than training loss alone, which leads the 9B run to select \texttt{global\_step\_543} (epoch 3 of 6) over the later epochs. Table~\ref{tab:sft-runs} reports the per-model settings.

\begin{table}[!htbp]
\centering
\footnotesize
\renewcommand{\arraystretch}{1.12}
\resizebox{\columnwidth}{!}{%
\setlength{\tabcolsep}{4pt}
\begin{tabular}{@{}lll@{}}
\toprule
Setting & 2B SFT & 9B SFT \\
\midrule
Init model & v0-pretrained cont. & \texttt{Qwen3.5-9B} \\
Data split & guided-v2 & merged v0+v2 \\
Train / valid rows & 2{,}330 / 594 & 2{,}886 / 732 \\
Global batch & 64 & 16 \\
Micro batch / GPU & 1 & 1 \\
Max seq length & 8{,}192 & 12{,}288 \\
Learning rate & 5e-6 & 5e-6 \\
Epochs / total steps & 3 / 111 & 6 / 1{,}086 \\
Saved ckpts (step) & 37, 74, 111 & 181, \ldots, 1086 (6 ckpts) \\
Selected step & 111 & 543 \\
GPUs (A100) & 4 & 8 \\
380-task score & 72.1 & 80.4 \\
\bottomrule
\end{tabular}%
}
\caption{Final SFT settings. The selected step is chosen by downstream ASIL-380 benchmark performance; for the 9B run this favors an intermediate epoch over the last epoch.}
\label{tab:sft-runs}
\end{table}

\subsection{RL Curriculum and Rollout Service}
\label{sec:rl-service}

For reinforcement learning, we use the learnable v4 curriculum with 320 training tasks and 80 validation tasks, exported as Verl-compatible rows. A training step is one on-policy GRPO-style update over \texttt{TRAIN\_BATCH\_SIZE} task prompts. For every prompt, the trainer requests \texttt{ROLLOUT\_N} trajectories from a vLLM policy server, which routes ASIL tool calls through an external rollout service. Each rollout job restores the task state, runs an ASIL agent loop for up to 20 action turns inside a managed Singularity ASIL worker, computes a scalar reward from the same evaluator used at benchmark time, and returns the reward together with the full trajectory messages. The trainer, the vLLM policy server, the rollout service, and the Singularity workers run as decoupled processes, and all RL evaluations set \texttt{ASIL\_DISABLE\_RENDER=1} so that the training environment matches the no-render benchmark environment. Both 2B and 9B runs use a reference model for KL regularization, one PPO epoch per update, and an overlength shaping reward with coefficient $0.1$ and a buffer of 256 tokens to discourage invalid or excessively long responses.

\subsection{RL Hyperparameters and Two-Stage Schedule}
\label{sec:rl-hparams}

The 9B run uses a two-stage schedule. The first stage initializes from the 9B SFT checkpoint \texttt{global\_step\_543}, runs to \texttt{global\_step\_80}, and saves every 40 steps near step 80 and 120. The resume stage loads the full Verl checkpoint at step 80, lowers the actor learning rate from $5\text{e-}7$ to $3\text{e-}7$, saves every 20 steps, and produces checkpoints at 100, 120, 140, 160, and 180. Downstream ASIL-380 benchmark performance selects \texttt{global\_step\_140} from the resume stage. The 2B run uses a single 40-step schedule with checkpoints saved every 8 steps; \texttt{global\_step\_8} maximizes the aggregate ASIL score, and later updates do not improve it further, indicating that the small-model policy reaches its useful operating point early. Table~\ref{tab:rl-runs} reports the per-model RL settings; both models reuse identical KL, entropy, advantage-normalization, and overlength-reward configurations.

\begin{table}[!htbp]
\centering
\footnotesize
\renewcommand{\arraystretch}{1.08}
\resizebox{\columnwidth}{!}{%
\setlength{\tabcolsep}{4pt}
\begin{tabular}{@{}lll@{}}
\toprule
Setting & 2B RL & 9B RL \\
\midrule
Init model & 2B SFT step 27 & 9B SFT step 543 \\
Curriculum train / valid & 320 / 80 & 320 / 80 \\
Hardware & 4 A800 & 8 A800 \\
Max ASIL steps / traj. & 20 & 20 \\
\texttt{NUM\_ENVS} & 12 & 8 \\
\texttt{ROLLOUT\_N} / \texttt{OVER\_SAMPLING} & 4 / 4 & 2 / 2 \\
\texttt{TRAIN\_BATCH} & 32 & 16 \\
\texttt{PPO\_MINI\_BATCH} & 4 & 8 \\
PPO epochs / update & 1 & 1 \\
Actor lr & 1e-6 & 5e-7 $\to$ 3e-7 \\
KL / entropy coef & 0.001 / 0.0 & 0.001 / 0.0 \\
Overlength (coef / buf) & 0.1 / 256 & 0.1 / 256 \\
Max prompt / resp. & 12{,}288 / 1{,}024 & 12{,}288 / 1{,}024 \\
vLLM max batch tokens & 8{,}192 & 4{,}096 \\
vLLM GPU mem util. & 0.55 & 0.35 \\
FSDP actor / ref offload & off / on & on / on \\
Total RL steps & 40 & 80 + 100 \\
Save freq. & every 8 & 40 / 20 \\
Selected ckpt & step 8 & step 140 \\
380-task score & 74.4 & 82.2 \\
\bottomrule
\end{tabular}%
}
\caption{Final RL settings. The 9B run uses a two-stage schedule with reduced learning rate during the resume stage; the 2B run uses a single short schedule whose best aggregate-score checkpoint occurs early.}
\label{tab:rl-runs}
\end{table}

\FloatBarrier

\section{Training Task Generation Overlap Audit}
\label{sec:taskgen-appendix}

\paragraph{Generation pipeline.} The RL curriculum reuses the ASIL full15 software environments and evaluators, but generates new train and validation tasks by rewriting task instructions and literal slots. The generator first builds a template catalog with action skeletons, slot categories, evaluator keys, and non-textual risk fingerprints. It does not expose held-out raw final-test instructions to the model. A task-generation model then proposes structured task specs that change workplace scenarios, target artifact names, labels, paths, layer names, track names, cell names, and other slot values. Deterministic slot replacement turns each spec back into executable ASIL task JSON, after which static audits check parser validity, evaluator consistency, and quota coverage.

\paragraph{Overlap scoring.} To reduce overlap with held-out tasks, each candidate is converted into a fingerprint containing software domain, normalized instruction tokens, character n-grams, action-operation signatures, evaluator-rule signatures, literal tokens, target artifacts, asset IDs, and source tags. The overlap score combines lexical, character n-gram, structural, and semantic similarities with weights 0.36, 0.18, 0.28, and 0.18, respectively. Hard rejects cover identical task IDs, identical sources, same target artifacts with high textual overlap, and forbidden held-out GIMP assets. For each candidate, a GPT-5.4 judge inspects the top heuristic-risk pairs and labels them as exact duplicate, near duplicate, same template, same domain but distinct, or unrelated. We reject candidates with hard rejects or maximum risk at least 0.72, warn on risk between 0.55 and 0.72, and accept below 0.55.

\paragraph{Selected curriculum.} The final v3 generated pool selects 512 training tasks and 128 validation tasks across all 15 software domains, with strict quotas of 32 train tasks per non-GIMP application and 64 for GIMP, and 8 validation tasks per non-GIMP application and 16 for GIMP. The selected sets contain zero rejected or hard-rejected tasks. The selected v3 train set has 91 strict-accept tasks, a 17.8\% strict-accept ratio, and maximum selected overlap risk 0.690; the selected v3 validation set has 19 strict-accept tasks, a 14.8\% strict-accept ratio, and maximum selected overlap risk 0.650. We treat this as an engineering-valid low-leakage source pool rather than as a fully de-templateized task generator, because many selected tasks remain in the warning band. The final RL runs use a learnable v4 subset with 320 train tasks and 80 validation tasks.

Table~\ref{tab:taskgen-audit}(a) reports the candidate-generation and overlap-audit iterations used to build the ASIL RL curriculum, and Table~\ref{tab:taskgen-audit}(b) summarizes the selected task sets after quota selection and internal duplicate checks.

\begin{table*}[!htbp]
\centering

\textbf{(a) Per-iteration candidate generation and overlap audit}\\[3pt]
\scriptsize
\setlength{\tabcolsep}{4pt}
\renewcommand{\arraystretch}{1.12}
\resizebox{\textwidth}{!}{%
\begin{tabular}{@{}llrrrrrrrp{4.2cm}@{}}
\toprule
Run & Split & Cand. & Refs & Accept & Warn & Reject & Acc.\,\% & Non-rej.\,\% & Method change \\
\midrule
\texttt{v1\_baseline} & combined & 16 & 306 & 0 & 0 & 16 & 0.0 & 0.0 & handwritten v1 train/valid baseline \\
\addlinespace[2pt]
\texttt{train\_r1} & train & 1024 & 306 & 65 & 155 & 804 & 6.3 & 21.5 & API slot rewrite, initial prompt \\
\texttt{train\_r2\_heuristic} & train & 1024 & 306 & 978 & 46 & 0 & 95.5 & 100.0 & expanded target-slot extraction, heuristic-only audit \\
\texttt{train\_r2\_calibrated} & train & 1024 & 306 & 60 & 515 & 449 & 5.9 & 56.2 & expanded target slots plus calibrated GPT judge \\
\texttt{train\_r3\_shortfall} & train & 512 & 306 & 26 & 206 & 280 & 5.1 & 45.3 & targeted shortfall generation for low-yield software \\
\texttt{train\_r4\_gimp} & train & 64 & 306 & 0 & 37 & 27 & 0.0 & 57.8 & targeted GIMP repair \\
\texttt{train\_final\_merged} & train & 1600 & 306 & 86 & 758 & 756 & 5.4 & 52.8 & merged v2 final train pool \\
\texttt{train\_r5} & train & 1536 & 306 & 89 & 963 & 484 & 5.8 & 68.5 & risk-aware R5 train pool \\
\texttt{train\_r6\_merged} & train & 1600 & 306 & 91 & 982 & 527 & 5.7 & 67.1 & R6 targeted train repair \\
\addlinespace[2pt]
\texttt{valid\_r2} & valid & 256 & 818 & 5 & 139 & 112 & 2.0 & 56.2 & initial valid pool judged against test plus train \\
\texttt{valid\_r3\_shortfall} & valid & 448 & 818 & 8 & 284 & 156 & 1.8 & 65.2 & targeted valid shortfall generation \\
\texttt{valid\_r4\_blender} & valid & 64 & 818 & 1 & 9 & 54 & 1.6 & 15.6 & targeted Blender repair \\
\texttt{valid\_final\_merged} & valid & 768 & 818 & 14 & 432 & 322 & 1.8 & 58.1 & merged v2 final valid pool \\
\texttt{valid\_r5} & valid & 384 & 818 & 19 & 272 & 93 & 4.9 & 75.8 & risk-aware R5 valid pool \\
\bottomrule
\end{tabular}%
}

\vspace{10pt}

\textbf{(b) Selected task-set summary after quota selection}\\[3pt]
\small
\setlength{\tabcolsep}{6pt}
\renewcommand{\arraystretch}{1.15}
\begin{tabular}{@{}lrrrrr@{}}
\toprule
Set & Selected & Acc. & Acc.\,\% & Rej. & Max risk \\
\midrule
\texttt{v2\_train} & 512 & 86 & 16.8 & 0 & 0.700 \\
\texttt{v3\_train\_r6} & 512 & 91 & 17.8 & 0 & 0.690 \\
\addlinespace[2pt]
\texttt{v2\_valid} & 128 & 14 & 10.9 & 0 & 0.700 \\
\texttt{v3\_valid} & 128 & 19 & 14.8 & 0 & 0.650 \\
\bottomrule
\end{tabular}

\caption{Training task generation overlap audit. \textbf{(a)} Per-iteration candidate generation and overlap audit: candidates with maximum overlap risk below 0.55 are accept, risk in $[0.55, 0.72)$ is warn, and hard rejects or risk $\geq 0.72$ are reject; non-reject rate is accept plus warn. Rows are grouped into baseline, train pools, and validation pools. \textbf{(b)} Selected task-set summary after quota selection and internal duplicate checks. All listed sets are quota-complete with zero quota backfill; the v3 pool is further filtered into the learnable v4 subset used for final RL.}
\label{tab:taskgen-audit}
\end{table*}

\FloatBarrier

\section{Reproducibility}
\label{sec:reproducibility}

We release the ASIL inference and evaluation stack to support reproduction of the benchmark setup and extension to new applications. Public resources are separated into the \href{https://github.com/sharryXR/ASIL}{GitHub code repository}, \href{https://huggingface.co/collections/sharryXR/asil-models-6a1e9faf39fe6ce4eb4626e1}{released model checkpoints}, \href{https://huggingface.co/datasets/sharryXR/asil-benchmark}{benchmark tasks}, \href{https://huggingface.co/datasets/sharryXR/asil-benchmark-images}{benchmark runtime images}, and \href{https://huggingface.co/datasets/sharryXR/asil-training-data}{training data}. The release covers: (i) the adapter library for all 15 applications, the shared observation--action protocol, the evaluator implementations described in Appendix~\ref{sec:asil-implementation-appendix}, and the semi-automatic ASILization pipeline (adapter templates, observation--schema scaffolds, evaluator-rule scaffolds, and validation tooling); (ii) task definitions for the 380-task main benchmark, the 80-task hard suite, the easy60 band, and the prepared 320/80 RL curriculum derived from the v3 generation pool of Appendix~\ref{sec:taskgen-appendix}; (iii) the prepared SFT-v0, agentic-guided-v2, and merged 9B datasets summarized in Table~\ref{tab:sft-data}, together with the selected Qwen3.5-2B and Qwen3.5-9B SFT and RL checkpoints used in Tables~\ref{tab:main-results}, \ref{tab:hard-results}, and \ref{tab:pattern-ablation}; and (iv) the repaired GUI, native-UNO, and draw.io-MCP baseline scripts used for the camera-ready audits. A one-command Docker path builds the runtime from source on x86\_64 Ubuntu 22.04/24.04, launches the desktop/service dependencies, runs a deterministic no-key smoke test, and emits a fail-closed readiness report over the 15 adapter gates.

\end{document}

%% file: figures/styles.tex
\usepackage{xcolor}
\usepackage{graphicx}
\usepackage{tikz}
\usetikzlibrary{arrows.meta, backgrounds, calc, fit, positioning}

\newcommand{\FigureCornerRadius}{2.7pt}
\newcommand{\FigureGroupCorner}{5.4pt}
\newcommand{\FigureArrowWidth}{1.05pt}

\newlength{\FigureModuleWidth}
\newlength{\FigureModuleHeight}
\newlength{\FigureCompactWidth}
\newlength{\FigureCompactHeight}
\newlength{\FigureXGap}
\newlength{\FigureYGap}
\newlength{\FigureGroupPadding}

\definecolor{FigureLegacyBorder}{HTML}{E86F2D}
\definecolor{FigureLegacyFill}{HTML}{FFF7ED}

\definecolor{FigureCoreBorder}{HTML}{2563EB}
\definecolor{FigureCoreFill}{HTML}{EEF6FF}

\definecolor{FigureBackendBorder}{HTML}{F59E0B}
\definecolor{FigureBackendFill}{HTML}{FFF7E2}

\definecolor{FigureBenchBorder}{HTML}{64748B}
\definecolor{FigureBenchFill}{HTML}{F8FAFC}

\definecolor{FigureMintBorder}{HTML}{4E8A63}
\definecolor{FigureMintFill}{HTML}{D9E9D8}

\definecolor{FigureSkyBorder}{HTML}{4F77A8}
\definecolor{FigureSkyFill}{HTML}{D6E4F3}

\definecolor{FigureLilacBorder}{HTML}{6F6A95}
\definecolor{FigureLilacFill}{HTML}{E2DEED}

\definecolor{FigureRoseBorder}{HTML}{9A5F79}
\definecolor{FigureRoseFill}{HTML}{EEDBE4}

\definecolor{FigureLimeBorder}{HTML}{6F8B42}
\definecolor{FigureLimeFill}{HTML}{DDE9CD}

\definecolor{FigurePeachBorder}{HTML}{A66E43}
\definecolor{FigurePeachFill}{HTML}{F0DCC8}

\colorlet{FigureTextDark}{black!82}
\colorlet{FigureTextMuted}{black!60}

\newcommand{\FigureGroupTitleFont}{\sffamily\bfseries\footnotesize}
\newcommand{\FigureLegendFont}{\sffamily\fontsize{6.5}{7.3}\selectfont}

\tikzset{
  moduleBase/.style={
    draw=black!45,
    fill=white,
    rounded corners=\FigureCornerRadius,
    line width=0.8pt,
    align=center,
    text width=\FigureModuleWidth,
    minimum height=\FigureModuleHeight,
    inner xsep=5.5pt,
    inner ysep=4.8pt,
    text=FigureTextDark,
  },
  compactBase/.style={
    draw=black!38,
    fill=white,
    rounded corners=5.4pt,
    line width=0.72pt,
    align=center,
    text width=\FigureCompactWidth,
    minimum height=\FigureCompactHeight,
    inner xsep=4.5pt,
    inner ysep=3.2pt,
    text=FigureTextDark,
  },
  heroBox/.style={
    moduleBase,
    draw=FigureBenchBorder!78,
    fill=FigureBenchFill!85,
  },
  legacyBox/.style={
    moduleBase,
    draw=FigureLegacyBorder!92,
    fill=FigureLegacyFill,
  },
  coreBox/.style={
    moduleBase,
    draw=FigureCoreBorder!95,
    fill=FigureCoreFill,
  },
  backendBox/.style={
    moduleBase,
    draw=FigureBackendBorder!92,
    fill=FigureBackendFill,
  },
  benchBox/.style={
    moduleBase,
    draw=FigureBenchBorder!82,
    fill=FigureBenchFill,
  },
  issueNote/.style={
    compactBase,
    text width=1.56cm,
    minimum height=0.62cm,
    draw=FigureLegacyBorder!80,
    fill=FigureLegacyFill!95!white,
    text=FigureLegacyBorder!95!black,
  },
  principlePill/.style={
    compactBase,
    text width=1.48cm,
    minimum height=0.60cm,
    draw=FigureCoreBorder!75,
    fill=white,
    text=FigureCoreBorder!95!black,
  },
  benchPill/.style={
    compactBase,
    text width=1.92cm,
    minimum height=0.80cm,
    draw=FigureBenchBorder!80,
    fill=white,
    text=FigureBenchBorder!95!black,
  },
  softGroup/.style={
    rounded corners=\FigureGroupCorner,
    line width=0.72pt,
    draw=black!24,
    fill=black!1,
    inner sep=\FigureGroupPadding,
  },
  legacyGroup/.style={
    softGroup,
    draw=FigureLegacyBorder!42,
    fill=FigureLegacyFill!58!white,
  },
  coreGroup/.style={
    softGroup,
    draw=FigureCoreBorder!36,
    fill=FigureCoreFill!54!white,
  },
  backendGroup/.style={
    softGroup,
    draw=FigureBackendBorder!40,
    fill=FigureBackendFill!56!white,
  },
  solidFlow/.style={
    -{Latex[length=2.8mm,width=2.1mm]},
    line width=\FigureArrowWidth,
    draw=black!72,
  },
  legacyFlow/.style={
    solidFlow,
    draw=FigureLegacyBorder!96,
  },
  coreFlow/.style={
    solidFlow,
    draw=FigureCoreBorder!98,
  },
  backendFlow/.style={
    solidFlow,
    draw=FigureBackendBorder!95,
  },
  feedbackFlow/.style={
    -{Latex[length=2.7mm,width=1.9mm]},
    line width=\FigureArrowWidth,
    draw=black!56,
    dashed,
  },
  storyFlow/.style={
    -{Latex[length=3.1mm,width=2.3mm]},
    line width=1.12pt,
    draw=black!50,
  },
  groupTitle/.style={
    font=\FigureGroupTitleFont,
    anchor=west,
    text=FigureTextDark,
    fill=white,
    inner xsep=4pt,
    inner ysep=1pt,
    rounded corners=2pt,
  },
  legendText/.style={
    font=\FigureLegendFont,
    text=FigureTextMuted,
    anchor=west,
  },
  appBadge/.style={
    draw=FigureBenchBorder!64,
    fill=white,
    rounded corners=4.6pt,
    line width=0.74pt,
    font=\sffamily\scriptsize,
    text=FigureTextDark,
    inner xsep=5.0pt,
    inner ysep=2.5pt,
  },
  sectionLabel/.style={
    font=\sffamily\bfseries\footnotesize,
    text=FigureTextMuted,
    anchor=west,
  },
  supportBox/.style={
    rounded corners=9pt,
    line width=0.78pt,
    draw=FigureBenchBorder!38,
    fill=white,
    inner xsep=9pt,
    inner ysep=8pt,
    text=FigureTextDark,
  },
  softRule/.style={
    draw=FigureBenchBorder!36,
    line width=0.64pt,
  },
}

%% file: figures/method_overview.tex

\begin{figure*}[t]
\centering
\resizebox{0.80\textwidth}{!}{%
\begin{tikzpicture}[
  x=1cm,
  y=1cm,
  font=\sffamily,
  panelTitle/.style={
    font=\sffamily\bfseries\fontsize{18.6}{19.6}\selectfont,
    anchor=west,
  },
  panelSub/.style={
    font=\sffamily\fontsize{7.9}{8.7}\selectfont,
    text=FigureTextMuted,
    anchor=west,
    fill=white,
    fill opacity=0.92,
    text opacity=1,
    inner xsep=2.2pt,
    inner ysep=0.9pt,
    rounded corners=2pt,
  },
  badge/.style={
    rounded corners=5pt,
    draw=FigureCoreBorder!48,
    fill=white,
    line width=0.55pt,
    inner xsep=5pt,
    inner ysep=2.2pt,
    font=\sffamily\bfseries\fontsize{7.2}{7.9}\selectfont,
    text=FigureCoreBorder!92!black,
  },
  cardLabel/.style={
    font=\sffamily\bfseries\fontsize{7.6}{8.3}\selectfont,
    text=FigureCoreBorder!94!black,
    fill=white,
    fill opacity=0.88,
    text opacity=1,
    inner xsep=3.0pt,
    inner ysep=1.0pt,
    rounded corners=2pt,
    anchor=center,
    align=center,
  },
  cardLabelTeal/.style={
    cardLabel,
    text=black!72,
  },
  chipText/.style={
    font=\sffamily\bfseries\fontsize{7.15}{7.8}\selectfont,
    anchor=center,
    align=center,
    text height=1.45ex,
    text depth=0.35ex,
    text=FigureCoreBorder!92!black,
  },
  oldText/.style={
    font=\sffamily\bfseries\fontsize{7.3}{8.0}\selectfont,
    anchor=center,
    align=center,
    text=FigureLegacyBorder!95!black,
    fill=white,
    fill opacity=0.82,
    text opacity=1,
    inner xsep=2.0pt,
    inner ysep=0.7pt,
    rounded corners=2pt,
  },
  oldChip/.style={
    font=\sffamily\bfseries\fontsize{6.85}{7.45}\selectfont,
    anchor=center,
    align=center,
    minimum width=1.15cm,
    minimum height=0.36cm,
    text height=1.45ex,
    text depth=0.35ex,
    text=FigureLegacyBorder!95!black,
  },
  backendText/.style={
    font=\sffamily\bfseries\fontsize{7.3}{8.0}\selectfont,
    anchor=west,
    text=FigureBackendBorder!95!black,
    fill=white,
    fill opacity=0.86,
    text opacity=1,
    inner xsep=2.2pt,
    inner ysep=0.7pt,
    rounded corners=2pt,
  },
  evalText/.style={
    font=\sffamily\bfseries\fontsize{7.7}{8.3}\selectfont,
    anchor=center,
    align=center,
    text height=1.45ex,
    text depth=0.35ex,
    text=FigureBenchBorder!94!black,
  },
  evalSubText/.style={
    font=\sffamily\bfseries\fontsize{5.0}{5.6}\selectfont,
    anchor=center,
    align=center,
    text height=1.35ex,
    text depth=0.25ex,
    text=FigureBenchBorder!72!black,
  },
  supportLabel/.style={
    font=\sffamily\bfseries\fontsize{5.7}{6.3}\selectfont,
    anchor=center,
    text=FigureBenchBorder!92!black,
    fill=white,
    fill opacity=0.88,
    text opacity=1,
    inner xsep=2.4pt,
    inner ysep=0.7pt,
    rounded corners=2pt,
  },
]

\path[use as bounding box] (0,0) rectangle (18.0,10.0);
\node[anchor=south west, inner sep=0pt] at (0,0)
  {\includegraphics[width=18cm]{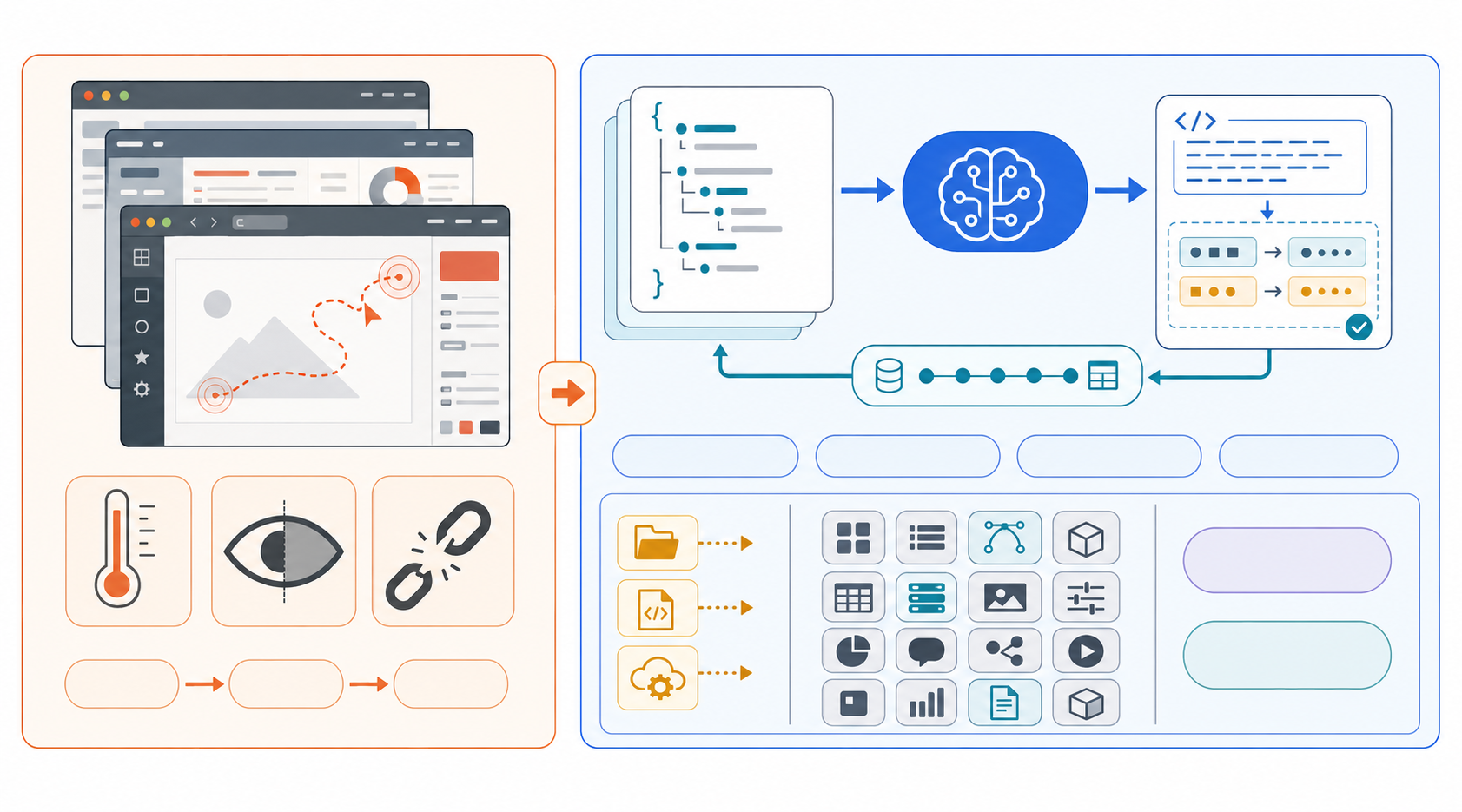}};

\node[panelTitle, text=FigureLegacyBorder!95!black] at (0.50,9.70) {GUI Agent};
\node[panelSub] at (0.53,9.34) {screenshot observation + coordinate clicks};
\node[panelTitle, text=FigureCoreBorder!95!black] at (7.38,9.70) {ASIL};
\node[panelSub] at (7.41,9.34) {structured state + semantic software actions};
\node[badge, anchor=east] at (17.55,9.62) {agent-native interface};

\node[cardLabelTeal] at (8.54,9.00) {Structured state};
\node[cardLabel] at (12.26,8.40) {Planner};
\node[cardLabel] at (15.26,9.00) {Semantic action};
\node[cardLabelTeal] at (12.26,5.72) {State update};

\node[oldText] at (1.56,2.42) {Cost};
\node[oldText] at (3.51,2.42) {Partial};
\node[oldText] at (5.42,2.42) {Fragile};
\node[oldChip] at (1.48,1.58) {surface};
\node[oldChip] at (3.50,1.58) {grounding};
\node[oldChip] at (5.46,1.58) {motor};

\node[chipText] at (8.67,4.45) {complete};
\node[chipText] at (11.16,4.45) {readable};
\node[chipText] at (13.65,4.45) {semantic};
\node[chipText] at (16.10,4.45) {composable};

\node[supportLabel] at (8.12,3.88) {Access paths};
\node[supportLabel] at (11.96,3.88) {15 software envs};
\node[supportLabel] at (15.85,3.88) {Benchmark views};
\node[backendText] at (8.42,3.15) {file};
\node[backendText] at (8.42,2.27) {script};
\node[backendText] at (8.42,1.39) {API};
\node[evalText] at (15.85,3.15) {ASIL run};
\node[evalSubText] at (15.85,2.88) {semantic path};
\node[evalText] at (15.85,2.00) {GUI run};
\node[evalSubText] at (15.85,1.73) {pixel baseline};

\end{tikzpicture}%
}
\caption{%
\textbf{Overview of ASIL.}
Conventional GUI agents bind software operation to a human-native screenshot-and-click loop, requiring visual grounding over a partial pixel surface and brittle motor sequences.
ASIL replaces this interface with structured observations of software state and code-executable semantic actions, realized through file-, script-, and service-level access paths across 15 software environments and 380 benchmark tasks.%
}
\label{fig:method-overview}
\end{figure*}
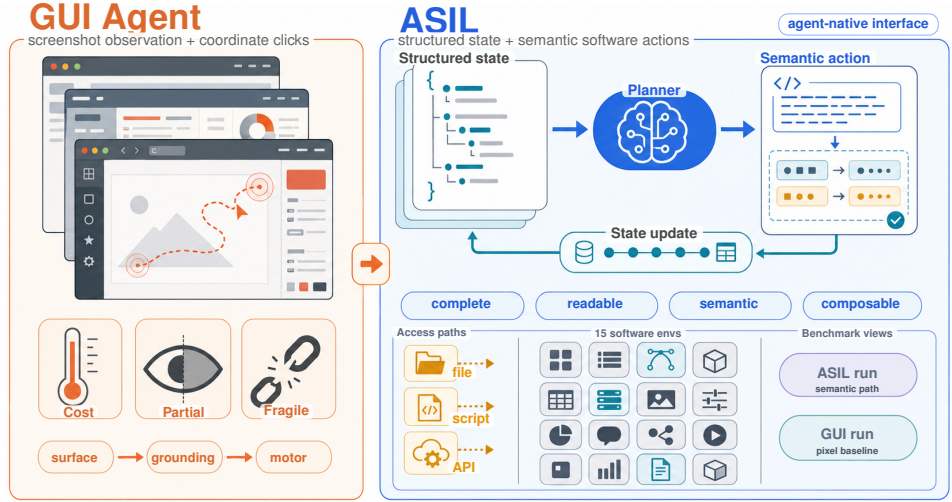

%% file: figures/schema_anatomy.tex

\begin{figure*}[t]
\centering
\resizebox{\textwidth}{!}{%
\begin{tikzpicture}[
  x=1cm,
  y=1cm,
  font=\sffamily,
  title/.style={
    font=\sffamily\bfseries\fontsize{15.6}{16.4}\selectfont,
    text=FigureBenchBorder!94!black,
    anchor=west,
  },
  sub/.style={
    font=\sffamily\fontsize{6.7}{7.4}\selectfont,
    text=FigureTextMuted,
    anchor=west,
  },
  panel/.style={
    rounded corners=11pt,
    line width=0.76pt,
    fill=white,
    inner sep=0pt,
  },
  panelTitle/.style={
    font=\sffamily\bfseries\fontsize{8.4}{9.1}\selectfont,
    anchor=west,
    text=FigureTextDark,
  },
  small/.style={
    font=\sffamily\fontsize{5.7}{6.3}\selectfont,
    text=FigureTextMuted,
  },
  tiny/.style={
    font=\sffamily\fontsize{4.9}{5.5}\selectfont,
    text=FigureTextMuted,
  },
  edgeLabel/.style={
    font=\sffamily\fontsize{4.9}{5.5}\selectfont,
    text=FigureTextMuted,
    inner sep=0pt,
  },
  fieldBase/.style={
    rounded corners=6pt,
    line width=0.62pt,
    minimum width=2.28cm,
    minimum height=0.54cm,
    align=center,
    inner xsep=3pt,
    inner ysep=2pt,
    font=\sffamily\bfseries\fontsize{6.1}{6.7}\selectfont,
    text=FigureTextDark,
    text height=1.05ex,
    text depth=.25ex,
  },
  fieldMint/.style={fieldBase, draw=FigureBenchBorder!58, fill=FigureMintFill},
  fieldSky/.style={fieldBase, draw=FigureBenchBorder!58, fill=FigureCoreFill},
  fieldLilac/.style={fieldBase, draw=FigureBenchBorder!54, fill=FigureMintFill!78!white},
  fieldRose/.style={fieldBase, draw=FigureBenchBorder!54, fill=FigureCoreFill!82!white},
  actionChip/.style={
    rounded corners=6pt,
    draw=FigureBackendBorder!68,
    fill=FigureBackendFill!85!white,
    line width=0.58pt,
    minimum width=1.84cm,
    minimum height=0.48cm,
    align=center,
    inner xsep=2pt,
    inner ysep=1.8pt,
    font=\sffamily\bfseries\fontsize{5.15}{5.8}\selectfont,
    text=FigureBackendBorder!82!black,
    text height=1.05ex,
    text depth=.25ex,
  },
  loopNode/.style={
    rounded corners=9pt,
    line width=0.70pt,
    fill=white,
    minimum width=2.22cm,
    minimum height=1.40cm,
    inner sep=0pt,
  },
  loopTitle/.style={
    font=\sffamily\bfseries\fontsize{6.7}{7.3}\selectfont,
    text=FigureTextDark,
    align=center,
  },
  lanePill/.style={
    rounded corners=10pt,
    line width=0.58pt,
    minimum width=1.84cm,
    minimum height=0.84cm,
    align=center,
    font=\sffamily\bfseries\fontsize{5.7}{6.3}\selectfont,
    text height=1.05ex,
    text depth=.25ex,
  },
  flow/.style={
    -{Latex[length=2.6mm,width=1.95mm]},
    line width=0.88pt,
    draw=FigureBenchBorder!78,
  },
  warmFlow/.style={
    -{Latex[length=2.6mm,width=1.95mm]},
    line width=0.88pt,
    draw=FigureBackendBorder!82,
  },
  feedback/.style={
    -{Latex[length=2.35mm,width=1.70mm]},
    line width=0.74pt,
    draw=FigureBenchBorder!66,
    dashed,
  },
]

\path[use as bounding box] (0,0) rectangle (18.0,10.20);
\fill[white, rounded corners=13pt] (0.18,0.18) rectangle (17.82,9.96);
\draw[FigureBenchBorder!64, rounded corners=13pt, line width=0.92pt] (0.18,0.18) rectangle (17.82,9.96);
\fill[FigureBenchBorder!10, rounded corners=13pt] (0.18,9.12) rectangle (17.82,9.96);
\draw[FigureBenchBorder!34, line width=0.48pt] (0.56,9.12) -- (17.44,9.12);
\node[title] at (0.64,9.62) {ASIL protocol and runtime loop};
\node[sub] at (0.66,9.30) {structured state and semantic actions close through planning, verification, memory, and retry};

\node[panel, draw=FigureBenchBorder!60, fill=FigureMintFill!62!white, minimum width=8.20cm, minimum height=3.18cm] (obsPanel) at (4.50,7.16) {};
\node[panel, draw=FigureBenchBorder!60, fill=FigureBackendFill!68!white, minimum width=8.20cm, minimum height=3.18cm] (actPanel) at (13.50,7.16) {};

\node[panelTitle, text=FigureCoreBorder!82!black] at (0.88,8.48) {Structured observation};
\node[sub] at (0.90,8.24) {machine-readable software state};
\node[panelTitle, text=FigureBackendBorder!82!black] at (9.88,8.48) {Semantic action};
\node[sub] at (9.90,8.24) {code-executable operation envelope};

\draw[rounded corners=7pt, draw=FigureMintBorder!54, fill=FigureMintFill!72!white, line width=0.62pt]
  (0.94,5.92) rectangle (2.68,7.72);
\fill[FigureMintBorder!12, rounded corners=7pt] (0.94,7.42) rectangle (2.68,7.72);
\draw[FigureMintBorder!40, line width=0.40pt] (1.08,7.42) -- (2.54,7.42);
\foreach \x/\c in {1.16/FigureMintBorder,1.32/FigureSkyBorder,1.48/FigureBackendBorder} {
  \fill[\c!68] (\x,7.56) circle (0.036);
}
\draw[rounded corners=2pt, draw=FigureSkyBorder!38, fill=white, line width=0.38pt] (1.16,6.03) rectangle (2.42,6.30);
\foreach \x in {1.28,1.54,1.80,2.06} {
  \draw[FigureSkyBorder!36, line width=0.32pt] (\x,6.03) -- +(0,0.27);
}
\draw[FigureSkyBorder!36, line width=0.32pt] (1.16,6.16) -- +(1.26,0);
\draw[FigureMintBorder!78, line width=0.68pt] (1.20,7.48) -- (1.04,7.48) -- (1.04,6.16) -- (1.20,6.16);
\foreach \y/\w/\c in {7.30/0.48/FigureMintBorder,7.04/0.68/FigureSkyBorder,6.78/0.55/FigureLilacBorder,6.52/0.78/FigureRoseBorder,6.26/0.62/FigureLimeBorder} {
  \fill[\c!74] (1.38,\y) circle (0.038);
  \draw[\c!64, line width=0.58pt] (1.54,\y) -- +(0.65*\w,0);
}
\draw[rounded corners=2pt, draw=FigureMintBorder!42, fill=white, line width=0.36pt] (2.04,6.60) rectangle (2.44,7.12);
\draw[FigureMintBorder!46, line width=0.34pt] (2.12,6.98) -- +(0.22,0);
\draw[FigureMintBorder!38, line width=0.34pt] (2.12,6.84) -- +(0.16,0);
\draw[FigureBackendBorder!38, line width=0.34pt] (2.12,6.70) -- +(0.24,0);
\draw[FigureMintBorder!78, line width=0.68pt] (2.42,7.48) -- (2.58,7.48) -- (2.58,6.16) -- (2.42,6.16);
\draw[FigureBenchBorder!42, line width=0.40pt] (2.76,6.82) -- (3.02,6.82);
\draw[FigureBenchBorder!42, line width=0.40pt] (3.02,7.18) -- (3.30,7.18);
\draw[FigureBenchBorder!42, line width=0.40pt] (3.02,6.46) -- (3.30,6.46);

\node[fieldMint] at (4.10,7.68) {\texttt{meta}};
\node[fieldSky] at (6.64,7.68) {\texttt{app\_state}};
\node[fieldLilac, minimum width=2.52cm] at (4.10,6.98) {\texttt{interactive\_elements}};
\node[fieldRose] at (6.64,6.98) {\texttt{environment}};
\node[fieldLilac] at (4.10,6.28) {\texttt{navigation}};
\node[fieldRose] at (6.64,6.28) {\texttt{data\_summary}};

	\draw[rounded corners=7pt, draw=FigureBackendBorder!62, fill=FigureBackendFill!78!white, line width=0.62pt]
	  (10.02,5.96) rectangle (12.82,7.68);
	\fill[FigureBackendBorder!10, rounded corners=7pt] (10.02,7.38) rectangle (12.82,7.68);
	\foreach \x/\c in {10.18/FigureBackendBorder,10.34/FigureSkyBorder,10.50/FigureMintBorder} {
	  \fill[\c!66] (\x,7.53) circle (0.032);
	}
	\foreach \y in {7.22,6.90,6.58} {
	  \draw[FigureBackendBorder!18, line width=0.30pt] (10.18,\y) -- (12.62,\y);
	}
	\node[font=\sffamily\fontsize{4.35}{4.9}\selectfont, text=FigureBackendBorder!82!black, anchor=west] at (10.20,7.28) {\texttt{action\_type}};
	\node[font=\sffamily\fontsize{4.35}{4.9}\selectfont, text=FigureTextMuted, anchor=west] at (10.20,6.96) {\texttt{target}};
	\node[font=\sffamily\fontsize{4.35}{4.9}\selectfont, text=FigureTextMuted, anchor=west] at (10.20,6.64) {\texttt{params}};
	\node[font=\sffamily\fontsize{4.35}{4.9}\selectfont, text=FigureTextMuted, anchor=west] at (10.20,6.32) {\texttt{expect\_observation}};
		\draw[rounded corners=4pt, draw=FigureBackendBorder!54, fill=white, line width=0.36pt]
		  (11.88,7.17) rectangle (12.64,7.34);
		\node[tiny, text=FigureBackendBorder!82!black] at (12.26,7.255) {\texttt{set}};
		\draw[rounded corners=4pt, draw=FigureSkyBorder!44, fill=FigureSkyFill!40!white, line width=0.34pt]
		  (11.88,6.83) rectangle (12.64,7.04);
		\draw[FigureSkyBorder!50, line width=0.34pt] (12.02,6.935) circle (0.054);
		\draw[FigureSkyBorder!48, line width=0.30pt] (12.10,6.935) -- (12.42,6.935);
		\draw[FigureSkyBorder!42, line width=0.30pt] (12.02,6.99) -- +(0,0.05);
		\draw[FigureSkyBorder!42, line width=0.30pt] (12.02,6.83) -- +(0,-0.05);
		\foreach \x/\c in {11.88/FigureMintBorder,12.16/FigureBackendBorder,12.44/FigureLilacBorder} {
		  \draw[rounded corners=1.6pt, draw=\c!48, fill=white, line width=0.30pt] (\x,6.50) rectangle +(0.20,0.16);
		}
		\draw[rounded corners=3pt, draw=FigureCoreBorder!46, fill=white, line width=0.34pt]
		  (12.22,6.20) rectangle (12.64,6.40);
		\draw[FigureCoreBorder!74, line width=0.50pt] (12.30,6.30) -- (12.38,6.23) -- (12.56,6.38);

\node[actionChip] at (14.02,7.66) {\texttt{set\_value}};
\node[actionChip] at (16.16,7.66) {\texttt{invoke\_function}};
\node[actionChip] at (14.02,6.98) {\texttt{modify\_file}};
\node[actionChip] at (16.16,6.98) {\texttt{api\_call}};
\node[actionChip] at (14.02,6.30) {\texttt{navigate}};
\node[actionChip] at (16.16,6.30) {\texttt{batch}};

\draw[warmFlow] (8.62,7.16) -- (9.38,7.16);
\node[small, anchor=center] at (9.00,7.42) {planner emits};

\node[panel, draw=FigureBenchBorder!60, fill=FigureCoreFill!72!white, minimum width=16.90cm, minimum height=4.82cm] (loopPanel) at (9.00,2.74) {};
\node[panelTitle, text=FigureCoreBorder!82!black] at (0.88,4.86) {Closed execution loop};
\node[sub] at (0.90,4.62) {observe, plan, execute, verify, and repair through the same JSON contract};

\node[loopNode, draw=FigureMintBorder!60, fill=FigureMintFill!68!white] (env) at (2.02,3.16) {};
\node[loopTitle, text=FigureMintBorder!72!black] at (2.02,3.54) {software\\state};
\draw[rounded corners=2pt, draw=FigureMintBorder!42, fill=white, line width=0.45pt] (1.36,2.50) rectangle (2.60,3.08);
\fill[FigureMintBorder!10] (1.36,2.92) rectangle (2.60,3.08);
\foreach \x/\c in {1.48/FigureMintBorder,1.62/FigureSkyBorder,1.76/FigureBackendBorder} {
  \fill[\c!68] (\x,3.00) circle (0.026);
}
\draw[FigureMintBorder!36, line width=0.34pt] (1.52,2.80) rectangle (1.76,2.62);
\draw[FigureMintBorder!36, line width=0.34pt] (1.84,2.80) rectangle (2.18,2.62);
\draw[FigureMintBorder!36, line width=0.34pt] (2.26,2.80) rectangle (2.48,2.62);
\draw[rounded corners=2pt, draw=FigureMintBorder!34, fill=white, line width=0.40pt] (1.68,2.34) rectangle (2.32,2.52);
\draw[FigureMintBorder!46, line width=0.36pt] (1.80,2.43) -- +(0.38,0);
\fill[FigureMintBorder!70] (2.42,2.94) circle (0.040);
\fill[FigureMintBorder!70] (2.42,2.70) circle (0.040);

\node[loopNode, draw=FigureSkyBorder!60, fill=FigureSkyFill!74!white, minimum width=2.42cm] (obs) at (5.14,3.16) {};
\node[loopTitle, text=FigureSkyBorder!72!black] at (5.14,3.54) {observation\\JSON};
\draw[rounded corners=3pt, draw=FigureSkyBorder!36, fill=white, line width=0.38pt] (4.48,2.40) rectangle (5.68,3.02);
\fill[FigureSkyBorder!10] (4.48,2.84) rectangle (5.68,3.02);
\draw[FigureSkyBorder!46, line width=0.36pt] (4.60,2.92) -- +(0.42,0);
\draw[FigureMintBorder!46, line width=0.34pt] (5.12,2.92) -- +(0.32,0);
\draw[FigureSkyBorder!66, line width=0.50pt] (4.43,2.96) -- (4.30,2.96) -- (4.30,2.40) -- (4.43,2.40);
\draw[FigureSkyBorder!66, line width=0.50pt] (5.88,2.96) -- (6.01,2.96) -- (6.01,2.40) -- (5.88,2.40);
\foreach \y/\w/\c in {2.87/0.56/FigureSkyBorder,2.70/0.86/FigureMintBorder,2.53/0.68/FigureLilacBorder,2.36/0.48/FigurePeachBorder} {
  \fill[\c!70] (4.58,\y) circle (0.026);
  \draw[\c!54, line width=0.46pt] (4.72,\y) -- +(0.56*\w,0);
}
\draw[FigureSkyBorder!44, line width=0.34pt] (5.28,2.72) -- (5.50,2.72) -- (5.50,2.54) -- (5.70,2.54);
\draw[FigureSkyBorder!44, line width=0.34pt] (5.28,2.72) -- (5.48,2.88);
\fill[FigureSkyBorder!64] (5.28,2.72) circle (0.024);
\fill[FigureMintBorder!58] (5.70,2.54) circle (0.024);
\draw[rounded corners=1.4pt, draw=FigureSkyBorder!42, fill=FigureSkyFill!54!white, line width=0.28pt]
  (5.48,2.78) rectangle (5.86,2.94);
\draw[FigureSkyBorder!42, line width=0.28pt] (5.56,2.86) -- +(0.20,0);
\fill[FigureSkyBorder!54] (5.56,2.86) circle (0.014);
\fill[FigureMintBorder!48] (5.76,2.86) circle (0.014);

	\node[
	  rounded corners=15pt,
	  draw=FigureLilacBorder!70,
	  fill=FigureLilacFill,
	  line width=0.72pt,
	  minimum width=2.42cm,
	  minimum height=1.28cm,
	  align=center,
	  inner sep=0pt
	] (planner) at (8.48,3.16) {};
		\node[loopTitle, text=FigureLilacBorder!74!black] at (8.48,3.54) {agent\\planner};
			\draw[rounded corners=7pt, draw=FigureLilacBorder!50, fill=white, line width=0.44pt]
			  (7.70,2.54) rectangle (9.26,3.04);
		\foreach \x/\t/\c in {7.98/G/FigureMintBorder,8.48/S/FigureSkyBorder,8.98/A/FigureBackendBorder} {
		  \draw[rounded corners=3pt, draw=\c!48, fill=\c!10!white, line width=0.34pt]
			    (\x-0.18,2.84) rectangle (\x+0.18,3.00);
			  \node[tiny, text=\c!78!black] at (\x,2.92) {\t};
		}
			\draw[FigureLilacBorder!54, line width=0.38pt, -{Latex[length=1.15mm,width=0.85mm]}] (8.18,2.92) -- (8.30,2.92);
			\draw[FigureLilacBorder!54, line width=0.38pt, -{Latex[length=1.15mm,width=0.85mm]}] (8.68,2.92) -- (8.80,2.92);
			\draw[FigureLilacBorder!36, line width=0.34pt] (7.88,2.68) -- (9.08,2.68);
		\foreach \x/\c in {8.00/FigureMintBorder,8.28/FigureSkyBorder,8.56/FigureLilacBorder,8.84/FigureBackendBorder} {
		  \fill[\c!64] (\x,2.68) circle (0.024);
		}
		\draw[rounded corners=1.8pt, draw=FigureLilacBorder!36, fill=FigureLilacFill!34!white, line width=0.30pt]
		  (7.88,2.59) rectangle (8.38,2.72);
		\draw[rounded corners=1.8pt, draw=FigureLilacBorder!36, fill=white, line width=0.30pt]
		  (8.60,2.59) rectangle (9.08,2.72);

\node[loopNode, draw=FigureBackendBorder!70, fill=FigureBackendFill!78!white, minimum width=2.32cm] (action) at (11.78,3.16) {};
\node[loopTitle, text=FigureBackendBorder!82!black] at (11.78,3.54) {action\\JSON};
	\draw[rounded corners=4pt, draw=FigureBackendBorder!44, fill=white, line width=0.45pt] (11.05,2.42) rectangle (12.51,2.98);
	\fill[FigureBackendBorder!10] (11.05,2.80) rectangle (12.51,2.98);
	\draw[rounded corners=2pt, draw=FigureBackendBorder!42, fill=FigureBackendFill!46!white, line width=0.34pt] (11.18,2.50) rectangle (11.56,2.70);
	\draw[FigureBackendBorder!46, line width=0.34pt] (11.30,2.70) -- (11.30,2.78);
	\draw[FigureBackendBorder!46, line width=0.34pt] (11.44,2.70) -- (11.44,2.78);
	\draw[FigureBackendBorder!66, line width=0.50pt] (11.20,2.90) -- +(0.66,0);
	\draw[FigureBackendBorder!50, line width=0.50pt] (11.68,2.68) -- +(0.48,0);
	\draw[FigureBackendBorder!70, line width=0.50pt] (11.68,2.55) -- +(0.42,0);
	\fill[FigureBackendBorder!68] (12.22,2.68) circle (0.038);
	\draw[FigureBackendBorder!54, line width=0.42pt] (12.10,2.68) -- (12.34,2.68);

\node[loopNode, draw=FigureSkyBorder!70, fill=FigureSkyFill!82!white, minimum width=2.40cm] (update) at (15.04,3.16) {};
\node[loopTitle, text=FigureSkyBorder!74!black] at (15.04,3.54) {state update\\+ trace};
	\draw[rounded corners=3pt, draw=FigureSkyBorder!46, fill=white, line width=0.45pt] (14.36,2.42) rectangle (15.74,2.98);
	\fill[FigureSkyBorder!10] (14.36,2.74) rectangle (15.74,2.98);
	\draw[FigureSkyBorder!42, line width=0.34pt] (15.06,2.42) -- +(0,0.56);
	\draw[FigureSkyBorder!62, line width=0.46pt] (14.54,2.86) -- +(0.42,0);
	\draw[FigureBackendBorder!58, line width=0.46pt] (14.54,2.66) -- +(0.58,0);
	\draw[FigureCoreBorder!72, line width=0.62pt] (15.44,2.68) -- (15.54,2.54) -- (15.68,2.88);
	\draw[rounded corners=2pt, draw=FigureSkyBorder!36, fill=white, line width=0.40pt] (14.64,2.28) rectangle (15.34,2.44);
	\fill[FigureSkyBorder!55] (14.80,2.36) circle (0.020);
	\fill[FigureBackendBorder!50] (15.00,2.36) circle (0.020);

\draw[flow] (env.east) -- node[midway, above=3pt, edgeLabel] {observe} (obs.west);
\draw[flow] (obs.east) -- node[midway, above=3pt, edgeLabel] {plan} (planner.west);
\draw[flow] (planner.east) -- node[midway, above=3pt, edgeLabel] {emit} (action.west);
\draw[warmFlow] (action.east) -- node[midway, above=3pt, edgeLabel] {execute} (update.west);
\draw[flow] (update.south) -- ++(0,-0.70) -| (env.south);

	\node[lanePill, draw=FigureLilacBorder!50, fill=FigureLilacFill!72!white] (memory) at (5.14,1.18) {};
	\node[font=\sffamily\bfseries\fontsize{5.7}{6.3}\selectfont, text=FigureLilacBorder!76!black] at (5.14,1.38) {memory};
	\draw[FigureLilacBorder!46, line width=0.36pt] (4.76,1.06) arc[start angle=180,end angle=360,x radius=0.28,y radius=0.06];
	\draw[FigureLilacBorder!46, line width=0.36pt] (4.76,1.06) -- (4.76,0.90);
	\draw[FigureLilacBorder!46, line width=0.36pt] (5.32,1.06) -- (5.32,0.90);
	\draw[FigureLilacBorder!46, line width=0.36pt] (4.76,0.90) arc[start angle=180,end angle=360,x radius=0.28,y radius=0.06];
	\draw[FigureLilacBorder!36, line width=0.32pt] (4.86,0.98) -- (5.22,0.98);
	\draw[rounded corners=1.8pt, draw=FigureLilacBorder!38, fill=FigureLilacFill!38!white, line width=0.32pt] (5.28,0.91) rectangle (5.62,1.07);
	\draw[FigureLilacBorder!40, line width=0.28pt] (5.34,0.99) -- +(0.18,0);

	\node[lanePill, draw=FigureMintBorder!62, fill=FigureMintFill!62!white] (verify) at (8.48,1.18) {};
	\node[font=\sffamily\bfseries\fontsize{5.7}{6.3}\selectfont, text=FigureMintBorder!76!black] at (8.48,1.38) {verifier};
	\draw[rounded corners=2pt, draw=FigureMintBorder!38, fill=FigureMintFill!32!white, line width=0.32pt] (7.90,0.86) rectangle (9.06,1.10);
	\foreach \x/\c in {8.02/FigureCoreBorder,8.40/FigureMintBorder,8.78/FigureBackendBorder} {
	  \draw[\c!58, line width=0.32pt] (\x,0.94) rectangle +(0.11,0.11);
	}
	\draw[FigureCoreBorder!72, line width=0.54pt] (7.92,0.93) -- (8.00,0.86) -- (8.12,1.06);
	\draw[FigureMintBorder!52, line width=0.42pt] (8.50,0.96) circle (0.08);
	\draw[FigureBackendBorder!58, line width=0.46pt] (8.82,0.88) -- (8.98,1.04);
	\draw[FigureBackendBorder!58, line width=0.46pt] (8.98,0.88) -- (8.82,1.04);

	\node[lanePill, draw=FigureBackendBorder!60, fill=FigureBackendFill!82!white] (retry) at (11.78,1.18) {};
	\node[font=\sffamily\bfseries\fontsize{5.7}{6.3}\selectfont, text=FigureBackendBorder!82!black] at (11.78,1.38) {retry};
	\draw[rounded corners=2pt, draw=FigureBackendBorder!40, fill=FigureBackendFill!34!white, line width=0.32pt] (11.26,0.86) rectangle (12.30,1.08);
	\draw[FigureBackendBorder!48, line width=0.30pt] (11.38,1.00) -- +(0.32,0);
	\draw[FigureBackendBorder!38, line width=0.30pt] (11.38,0.92) -- +(0.54,0);
	\draw[FigureBackendBorder!62, line width=0.46pt, -{Latex[length=1.35mm,width=1.00mm]}] (11.48,0.90) arc[start angle=210,end angle=-25,radius=0.22];
	\draw[FigureBackendBorder!62, line width=0.46pt, -{Latex[length=1.35mm,width=1.00mm]}] (12.12,1.08) arc[start angle=25,end angle=210,radius=0.22];

	\node[lanePill, draw=FigureSkyBorder!58, fill=FigureSkyFill!70!white] (trace) at (15.04,1.18) {};
	\node[font=\sffamily\bfseries\fontsize{5.7}{6.3}\selectfont, text=FigureSkyBorder!78!black] at (15.04,1.38) {trace log};
	\draw[rounded corners=2pt, draw=FigureSkyBorder!42, fill=FigureSkyFill!42!white, line width=0.34pt] (14.46,0.86) rectangle (15.54,1.10);
	\foreach \y/\c in {1.04/FigureSkyBorder,0.96/FigureMintBorder,0.88/FigureBackendBorder} {
	  \fill[\c!56] (14.60,\y) circle (0.016);
	  \draw[\c!46, line width=0.30pt] (14.70,\y) -- +(0.58,0);
	}

	\draw[feedback] (update.south) -- (trace.north);
	\draw[feedback] (trace.south) -- ++(0,-0.16) -| node[pos=0.30, above=2pt, edgeLabel] {check} (verify.south);
	\draw[feedback] (verify.west) -- node[midway, above=3pt, edgeLabel] {store} (memory.east);
	\draw[feedback] (memory.north) -- ++(0,0.28) -| node[pos=0.72, left=1pt, edgeLabel] {context} (planner.south);
	\draw[feedback] (verify.east) -- node[midway, above=3pt, edgeLabel] {revise} (retry.west);
	\draw[feedback] (retry.north) -- ++(0,0.28) -| (action.south);

\end{tikzpicture}%
}
\caption{\textbf{ASIL protocol and runtime loop.} ASIL exposes software as structured observations and code-executable semantic actions, then closes the loop through planning, verification, memory, retry, and inspectable state updates under the same JSON contract.}
\label{fig:schema-anatomy}
\label{fig:agent-loop}
\end{figure*}
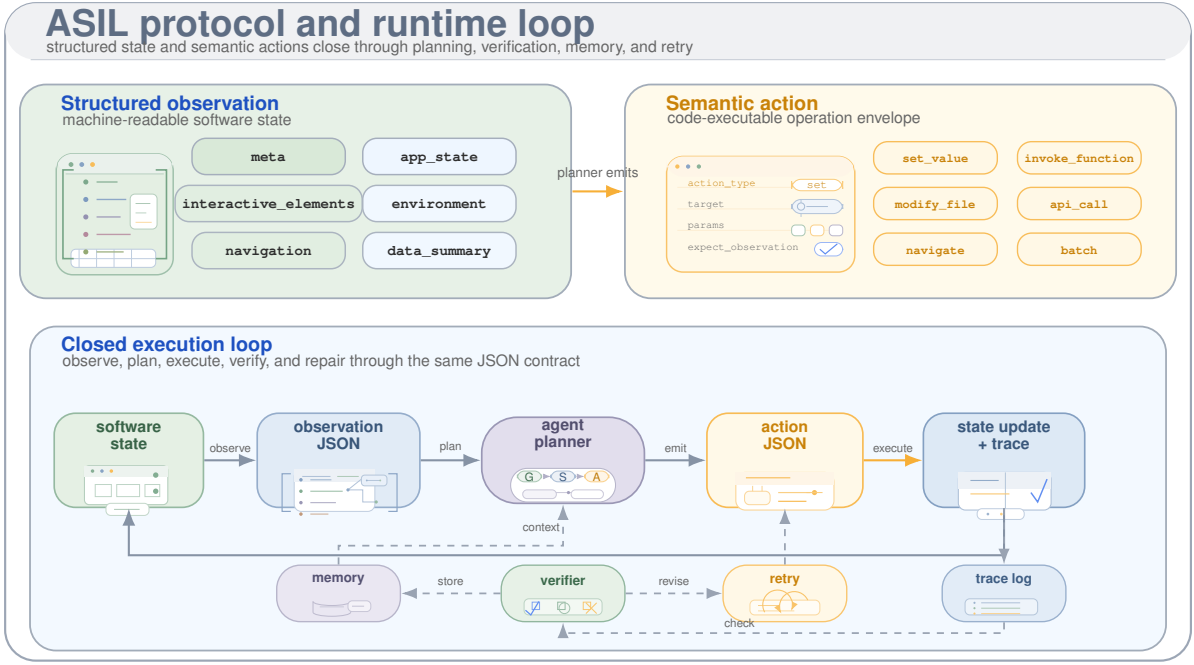

%% file: figures/benchmark_effect.tex
\begin{figure*}[t]
\centering
\begingroup
\newcommand{\BETileW}{3.60}
\newcommand{\BETileH}{2.03}
\newcommand{\BELabelH}{0.30}
\newcommand{\BenchmarkEffectTile}[5]{%
  \begin{scope}
    \clip[rounded corners=3.1pt] (#1,#2) rectangle ++(\BETileW,\BETileH);
    \node[anchor=south west, inner sep=0pt] at (#1,#2) {#3};
    \fill[white, opacity=0.88] (#1,#2) rectangle ++(\BETileW,\BELabelH);
  \end{scope}
  \draw[rounded corners=3.1pt, draw=#5!72!black, line width=0.54pt]
    (#1,#2) rectangle ++(\BETileW,\BETileH);
  \draw[draw=black!18, line width=0.28pt]
    ($ (#1,#2)+(0,\BELabelH) $) -- ($ (#1,#2)+(\BETileW,\BELabelH) $);
  \node[
    font=\sffamily\bfseries\fontsize{5.1}{5.6}\selectfont,
    text=FigureTextDark,
    anchor=center
  ] at ($ (#1,#2)+({0.5*\BETileW},{0.5*\BELabelH}) $) {#4};
}
\newcommand{\BEBar}[6]{%
  \node[
    anchor=east,
    font=\sffamily\bfseries\fontsize{4.9}{5.4}\selectfont,
    text=FigureTextDark
  ] at (#1,#2) {#3};
  \fill[rounded corners=1.8pt, fill=black!7] ($(#1,#2)+(0.14,-0.065)$) rectangle ++(2.55,0.13);
  \fill[rounded corners=1.8pt, fill=FigureCoreBorder!82] ($(#1,#2)+(0.14,-0.065)$) rectangle ++({2.55*#4/100},0.13);
  \fill[rounded corners=1.8pt, fill=black!7] ($(#1,#2)+(3.05,-0.065)$) rectangle ++(2.55,0.13);
  \fill[rounded corners=1.8pt, fill=FigurePeachBorder!78] ($(#1,#2)+(3.05,-0.065)$) rectangle ++({2.55*#5/100},0.13);
  \node[
    anchor=west,
    font=\sffamily\fontsize{4.65}{5.1}\selectfont,
    text=FigureTextDark!82
  ] at ($(#1,#2)+(5.78,0)$) {#6};
}
\resizebox{\textwidth}{!}{%
\begin{tikzpicture}[x=1cm,y=1cm]
  \path[use as bounding box] (0,0) rectangle (16.42,10.82);

  \fill[rounded corners=11pt, fill=FigureBenchFill!88, draw=FigureBenchBorder!25, line width=0.54pt]
    (0.22,0.14) rectangle (16.20,10.70);

  \node[
    anchor=west,
    font=\sffamily\bfseries\fontsize{10.5}{11.8}\selectfont,
    text=FigureTextDark
  ] at (0.58,10.36) {Benchmark coverage and interface effect};
  \node[
    anchor=west,
    font=\sffamily\fontsize{6.2}{7.0}\selectfont,
    text=FigureTextMuted
  ] at (0.60,10.04) {Representative real GUI surfaces from the 380-task benchmark.};

  \node[
    anchor=center,
    rounded corners=6pt,
    draw=FigureSkyBorder!76!black,
    fill=FigureSkyBorder!9,
    line width=0.52pt,
    font=\sffamily\bfseries\fontsize{5.45}{6.1}\selectfont,
    text=FigureTextDark,
    inner xsep=5pt,
    inner ysep=2.3pt
  ] at (9.82,10.25) {15 apps x 20 = 300};
  \node[
    anchor=center,
    rounded corners=6pt,
    draw=FigureMintBorder!76!black,
    fill=FigureMintBorder!9,
    line width=0.52pt,
    font=\sffamily\bfseries\fontsize{5.45}{6.1}\selectfont,
    text=FigureTextDark,
    inner xsep=5pt,
    inner ysep=2.3pt
  ] at (12.12,10.25) {80 multi-app tasks};
  \node[
    anchor=center,
    rounded corners=6pt,
    draw=FigureLilacBorder!76!black,
    fill=FigureLilacBorder!9,
    line width=0.52pt,
    font=\sffamily\bfseries\fontsize{5.45}{6.1}\selectfont,
    text=FigureTextDark,
    inner xsep=5pt,
    inner ysep=2.3pt
  ] at (14.34,10.25) {380 total};

  \draw[draw=FigureBenchBorder!18, line width=0.36pt]
    (0.60,9.86) -- (15.82,9.86);

  \BenchmarkEffectTile{0.57}{7.72}{\includegraphics[width=\BETileW cm,height=\BETileH cm]{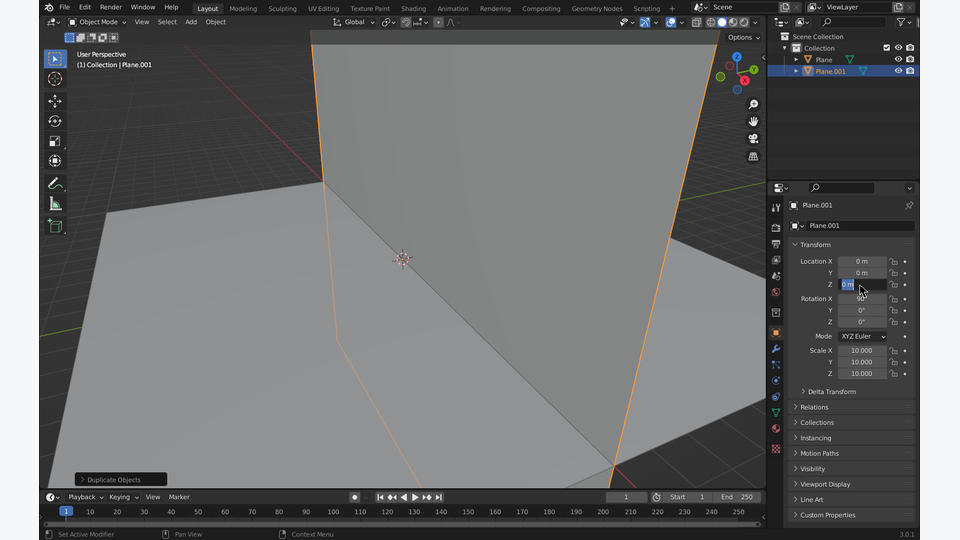}}{Blender}{FigureSkyBorder}
  \BenchmarkEffectTile{4.47}{7.72}{\includegraphics[width=\BETileW cm,height=\BETileH cm]{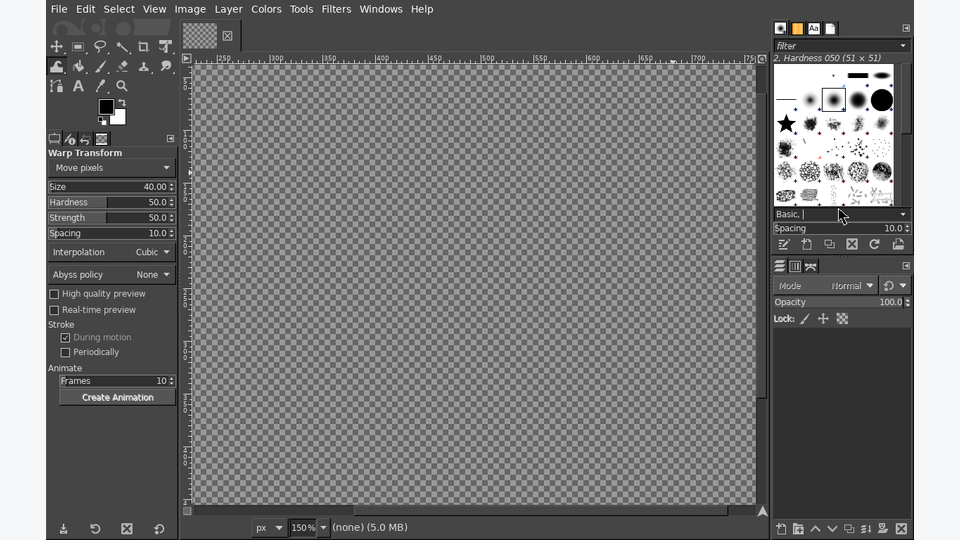}}{GIMP}{FigureRoseBorder}
  \BenchmarkEffectTile{8.37}{7.72}{\includegraphics[width=\BETileW cm,height=\BETileH cm]{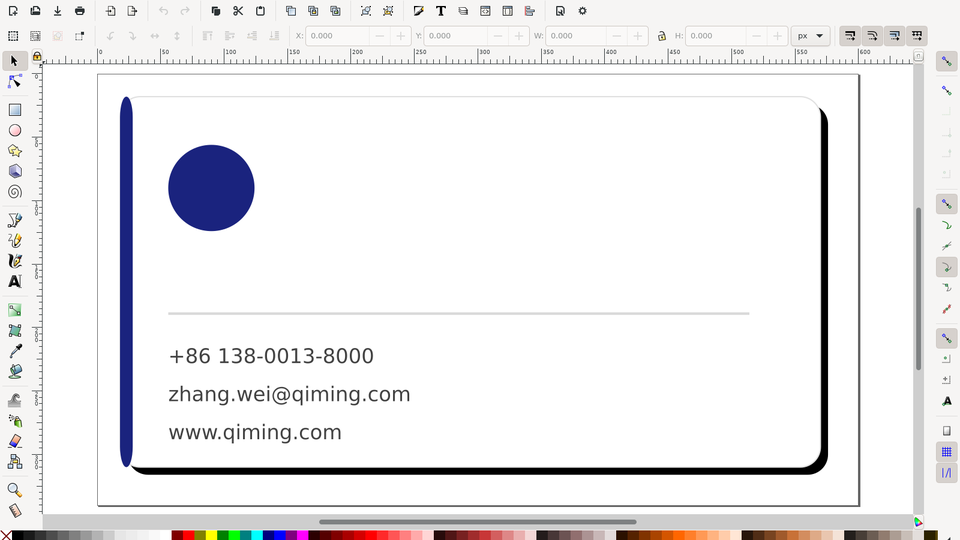}}{Inkscape}{FigureLilacBorder}
  \BenchmarkEffectTile{12.27}{7.72}{\includegraphics[width=\BETileW cm,height=\BETileH cm]{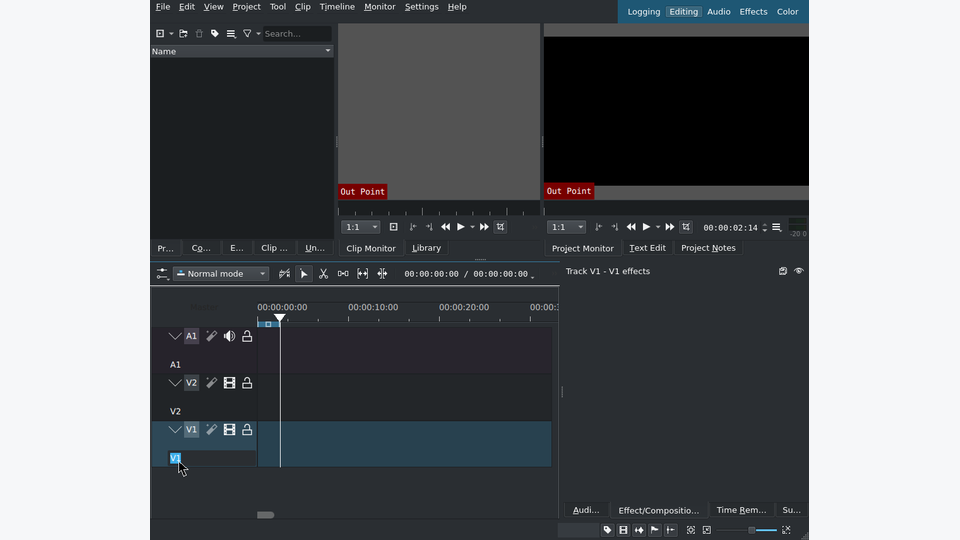}}{Kdenlive}{FigurePeachBorder}

  \BenchmarkEffectTile{0.57}{5.55}{\includegraphics[width=\BETileW cm,height=\BETileH cm]{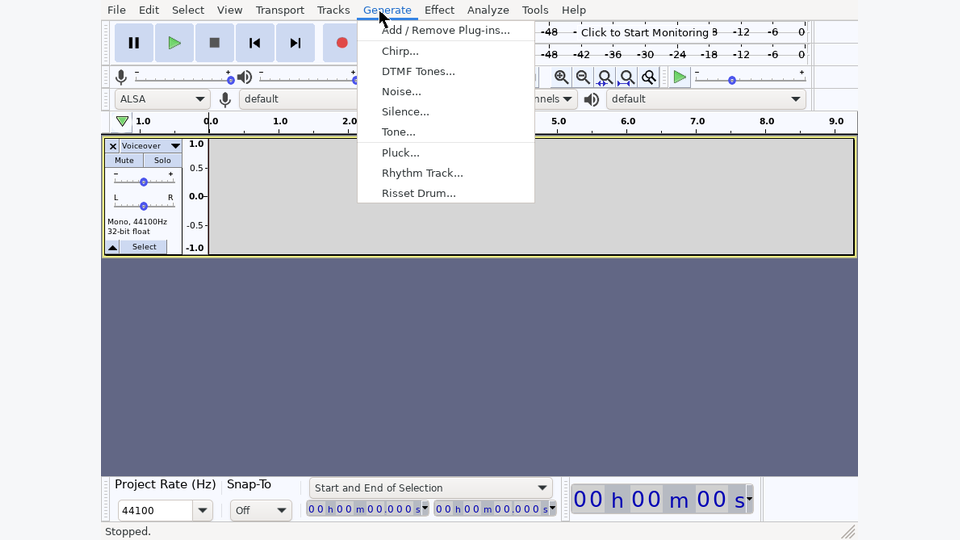}}{Audacity}{FigureMintBorder}
  \BenchmarkEffectTile{4.47}{5.55}{\includegraphics[width=\BETileW cm,height=\BETileH cm]{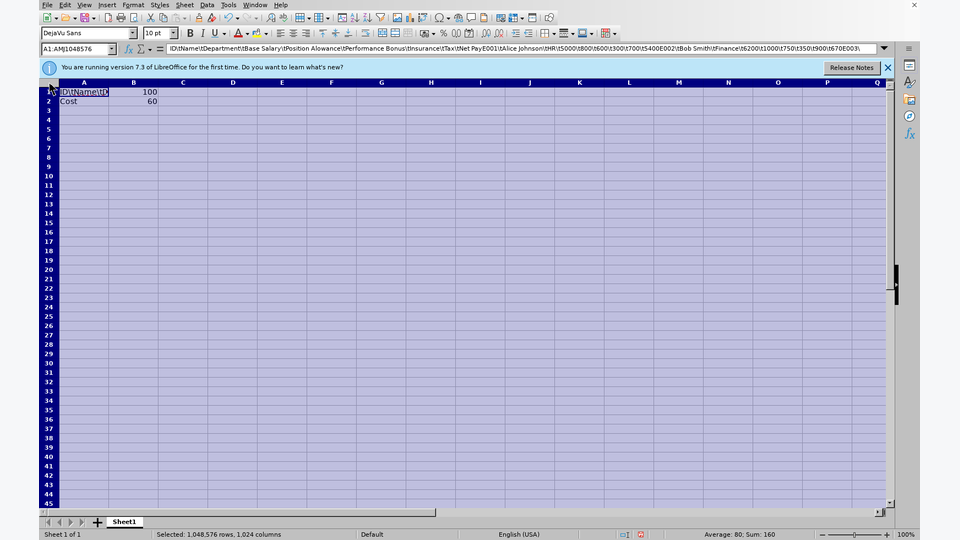}}{Calc}{FigureSkyBorder}
  \BenchmarkEffectTile{8.37}{5.55}{\includegraphics[width=\BETileW cm,height=\BETileH cm]{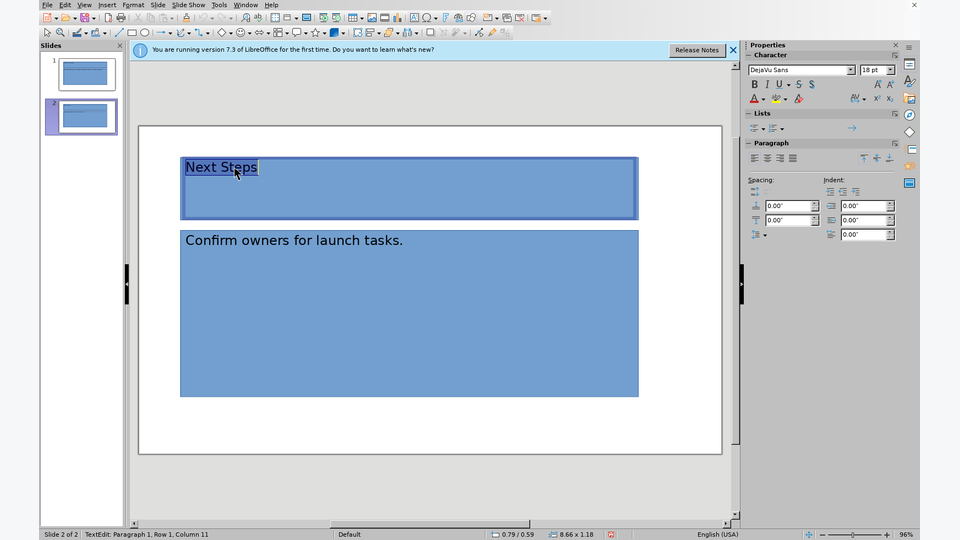}}{Impress}{FigureLilacBorder}
  \BenchmarkEffectTile{12.27}{5.55}{\includegraphics[width=\BETileW cm,height=\BETileH cm]{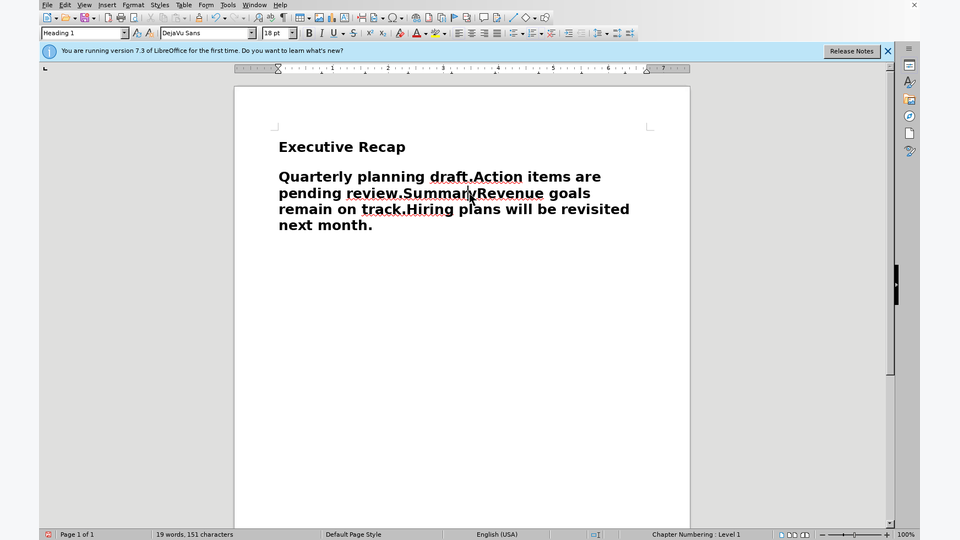}}{Writer}{FigureMintBorder}

  \BenchmarkEffectTile{0.57}{3.38}{\includegraphics[width=\BETileW cm,height=\BETileH cm]{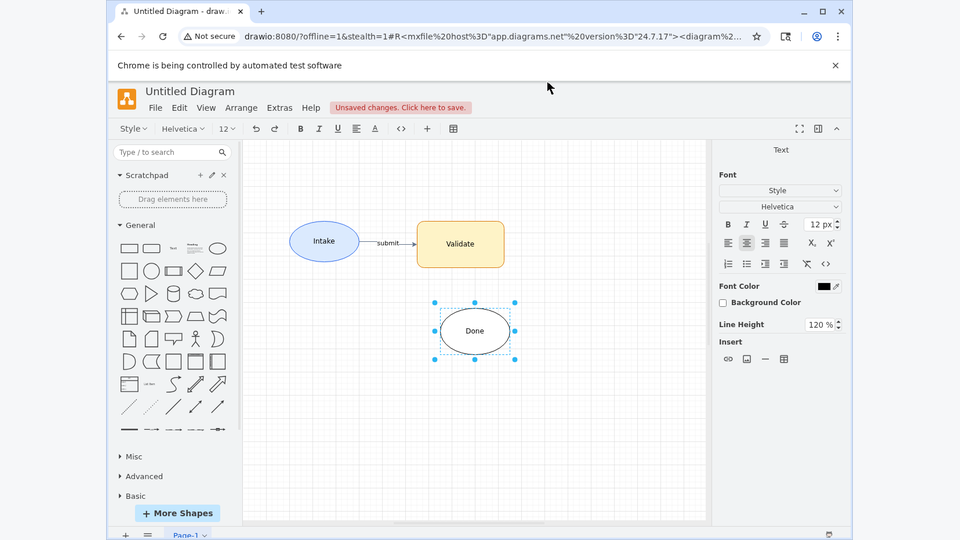}}{Draw.io}{FigurePeachBorder}
  \BenchmarkEffectTile{4.47}{3.38}{\includegraphics[width=\BETileW cm,height=\BETileH cm]{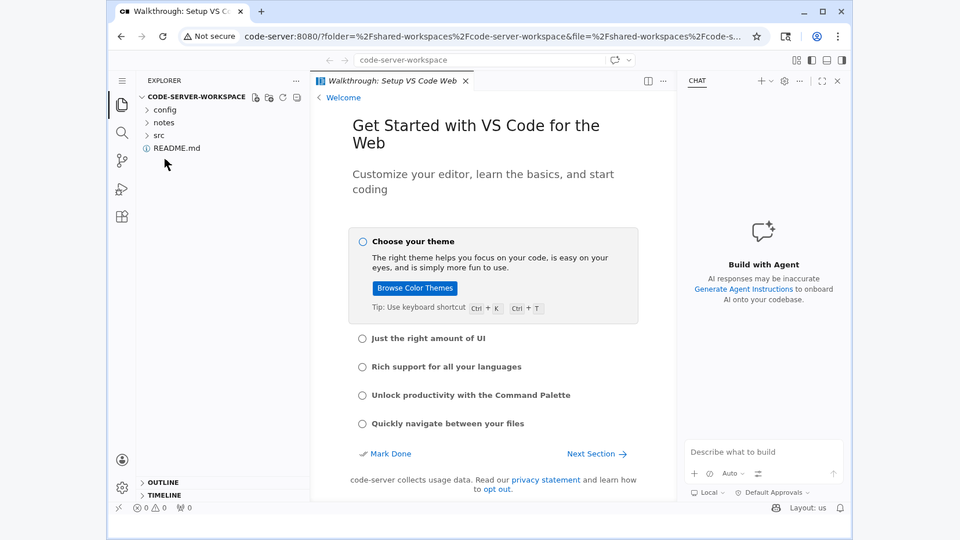}}{Code}{FigureSkyBorder}
  \BenchmarkEffectTile{8.37}{3.38}{\includegraphics[width=\BETileW cm,height=\BETileH cm]{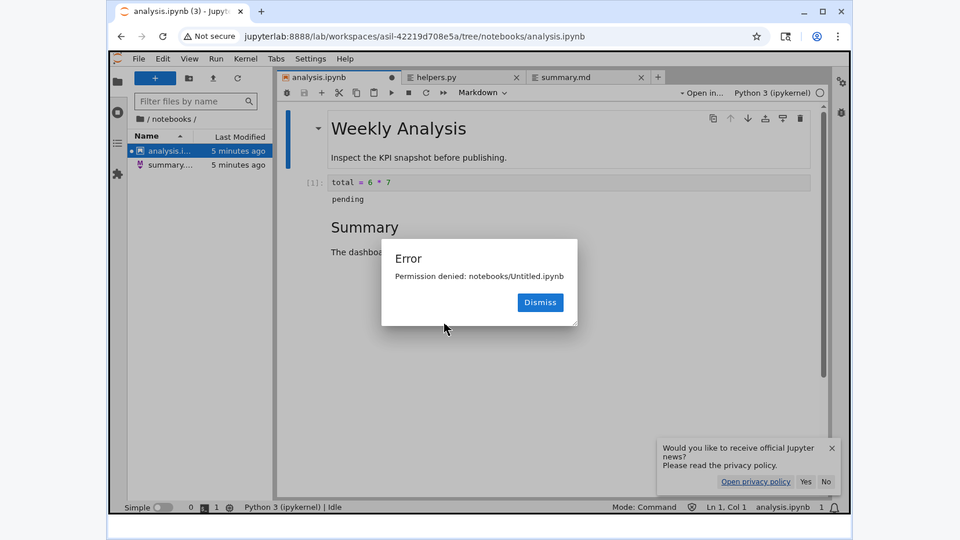}}{Jupyter}{FigureMintBorder}
  \BenchmarkEffectTile{12.27}{3.38}{\includegraphics[width=\BETileW cm,height=\BETileH cm]{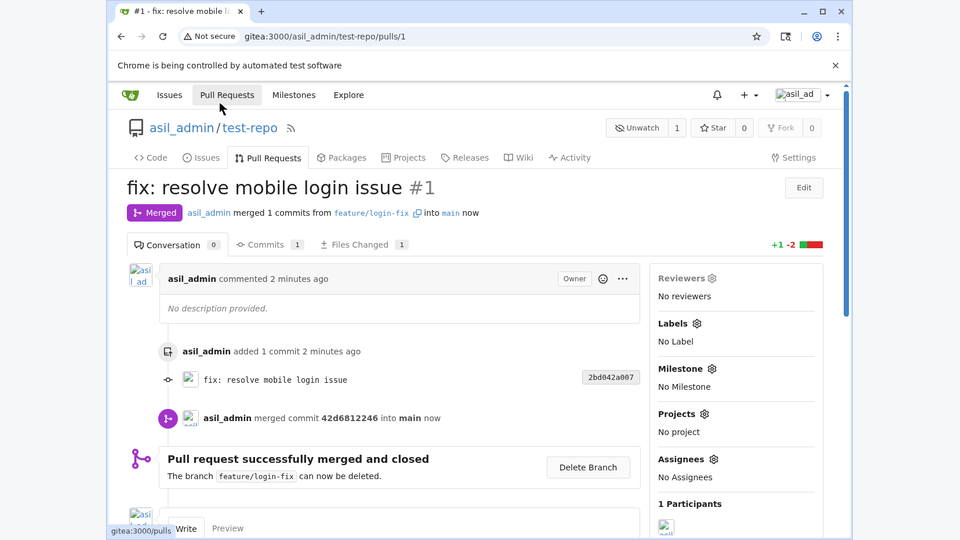}}{Gitea}{FigureRoseBorder}

  \BenchmarkEffectTile{0.57}{1.21}{\includegraphics[width=\BETileW cm,height=\BETileH cm]{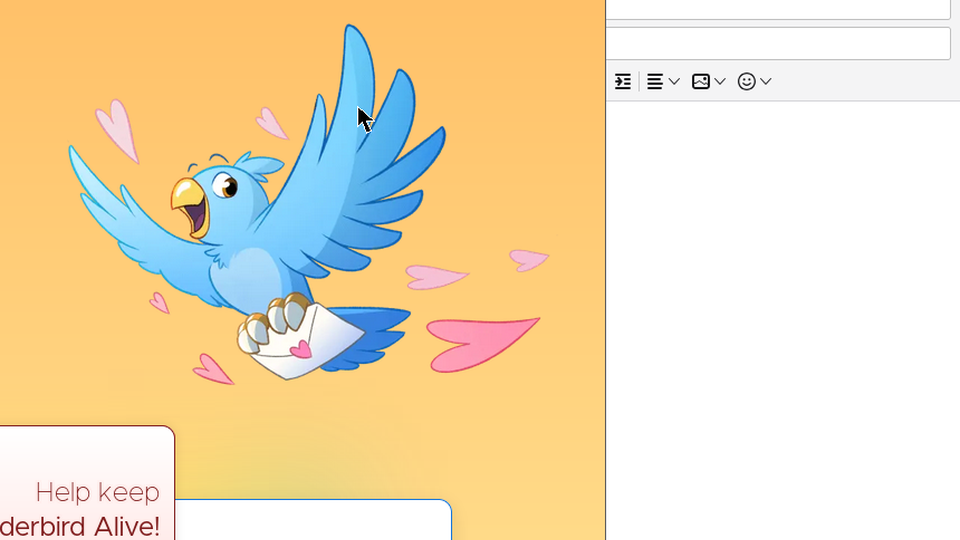}}{Thunderbird}{FigureLilacBorder}
  \BenchmarkEffectTile{4.47}{1.21}{\includegraphics[width=\BETileW cm,height=\BETileH cm]{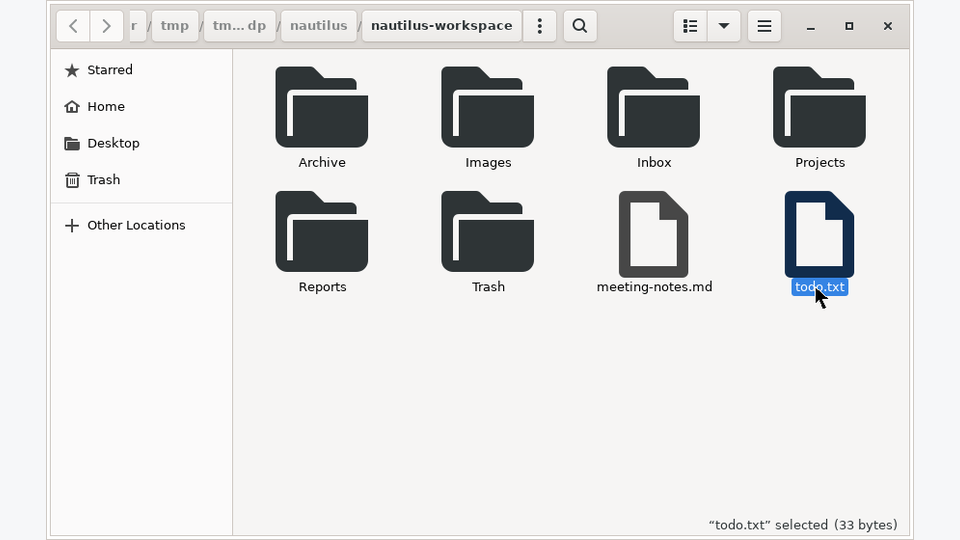}}{Nautilus}{FigureMintBorder}
  \BenchmarkEffectTile{8.37}{1.21}{\includegraphics[width=\BETileW cm,height=\BETileH cm]{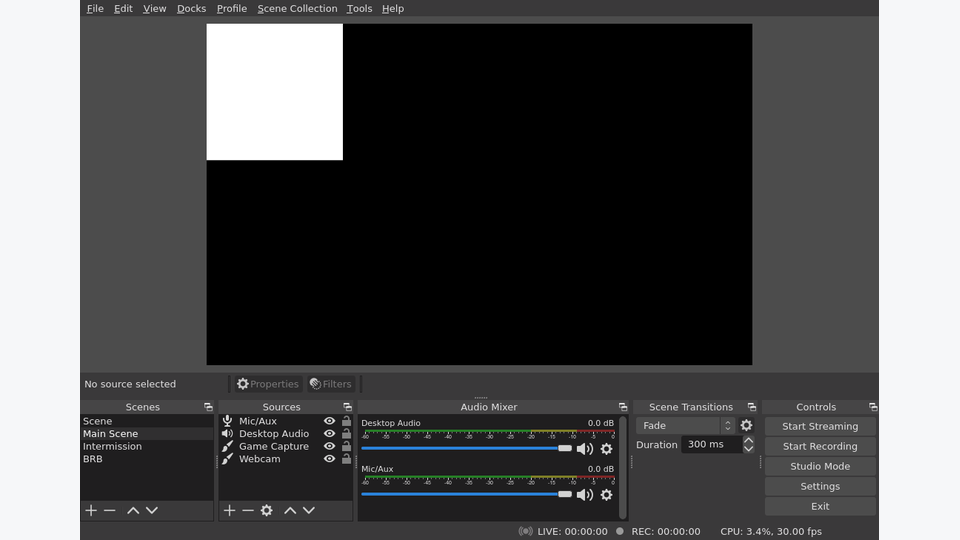}}{OBS}{FigurePeachBorder}
  \BenchmarkEffectTile{12.27}{1.21}{\includegraphics[width=\BETileW cm,height=\BETileH cm]{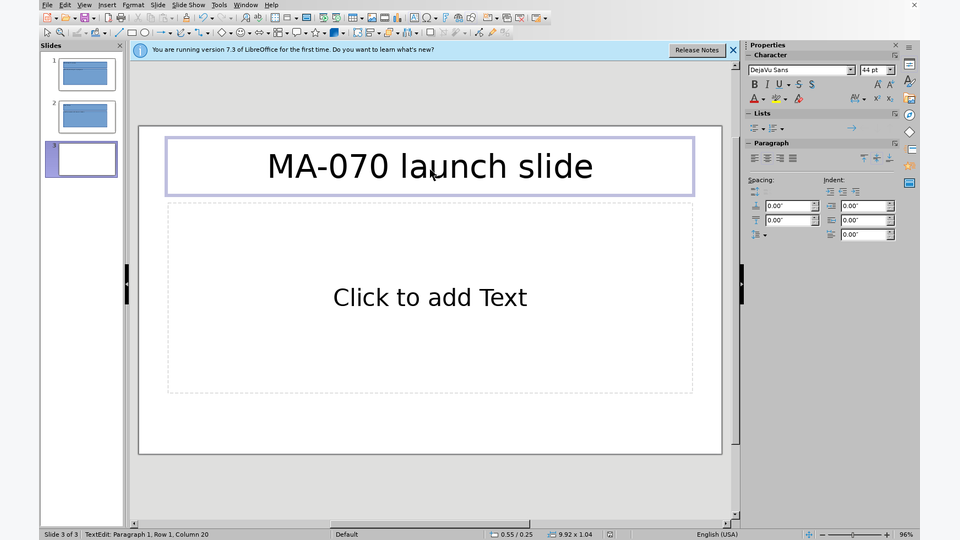}}{Multi-App}{FigureSkyBorder}

  \fill[rounded corners=7.5pt, fill=white, draw=FigureBenchBorder!34, line width=0.52pt]
    (0.57,0.16) rectangle (15.87,1.00);

  \node[
    anchor=west,
    font=\sffamily\bfseries\fontsize{5.3}{5.9}\selectfont,
    text=FigureTextDark
  ] at (0.84,0.82) {ASIL contract};
  \fill[rounded corners=3.3pt, fill=FigureSkyBorder!10, draw=FigureSkyBorder!76!black, line width=0.42pt]
    (2.28,0.53) rectangle ++(0.86,0.30);
  \node[font=\sffamily\bfseries\fontsize{4.55}{5.0}\selectfont, text=FigureTextDark]
    at (2.71,0.68) {state};
  \fill[rounded corners=3.3pt, fill=FigurePeachBorder!10, draw=FigurePeachBorder!76!black, line width=0.42pt]
    (3.68,0.53) rectangle ++(0.86,0.30);
  \node[font=\sffamily\bfseries\fontsize{4.55}{5.0}\selectfont, text=FigureTextDark]
    at (4.11,0.68) {action};
  \fill[rounded corners=3.3pt, fill=FigureMintBorder!10, draw=FigureMintBorder!76!black, line width=0.42pt]
    (5.08,0.53) rectangle ++(0.86,0.30);
  \node[font=\sffamily\bfseries\fontsize{4.55}{5.0}\selectfont, text=FigureTextDark]
    at (5.51,0.68) {check};
  \draw[-{Latex[length=1.45mm,width=1.05mm]}, draw=FigureBenchBorder!70, line width=0.46pt]
    (3.16,0.68) -- (3.50,0.68);
  \draw[-{Latex[length=1.45mm,width=1.05mm]}, draw=FigureBenchBorder!70, line width=0.46pt]
    (4.56,0.68) -- (4.90,0.68);
  \node[
    anchor=west,
    font=\sffamily\fontsize{4.7}{5.2}\selectfont,
    text=FigureTextMuted
  ] at (0.84,0.36) {structured state, semantic actions, evaluator checks};

  \draw[draw=FigureBenchBorder!24, line width=0.42pt]
    (6.12,0.30) -- (6.12,0.94);

  \node[
    anchor=east,
    font=\sffamily\bfseries\fontsize{5.2}{5.8}\selectfont,
    text=FigureTextDark
  ] at (7.75,0.82) {Overall score};
  \node[
    anchor=west,
    font=\sffamily\fontsize{4.75}{5.2}\selectfont,
    text=FigureCoreBorder!92!black
  ] at (8.32,0.82) {ASIL};
  \fill[rounded corners=0.8pt, fill=FigureCoreBorder!82]
    (8.15,0.775) rectangle ++(0.10,0.09);
  \node[
    anchor=west,
    font=\sffamily\fontsize{4.75}{5.2}\selectfont,
    text=FigurePeachBorder!92!black
  ] at (11.23,0.82) {GUI};
  \fill[rounded corners=0.8pt, fill=FigurePeachBorder!78]
    (11.06,0.775) rectangle ++(0.10,0.09);

  \BEBar{7.92}{0.58}{GPT-5.4}{81.6}{8.4}{81.6 vs. 8.4}
  \BEBar{7.92}{0.34}{Qwen3.6+}{81.1}{2.9}{81.1 vs. 2.9}
\end{tikzpicture}%
}
\endgroup
\caption{\textbf{Benchmark coverage and interface effect.} The benchmark spans 15 single-application environments and 80 multi-application workflows. The main panel shows representative real GUI screenshots from GPT-5.4 and Qwen3.6-plus benchmark result directories; the compact summary highlights the same-task ASIL-vs-GUI interface gap under the combined 380-task evaluation.}
\label{fig:benchmark-effect}
\end{figure*}
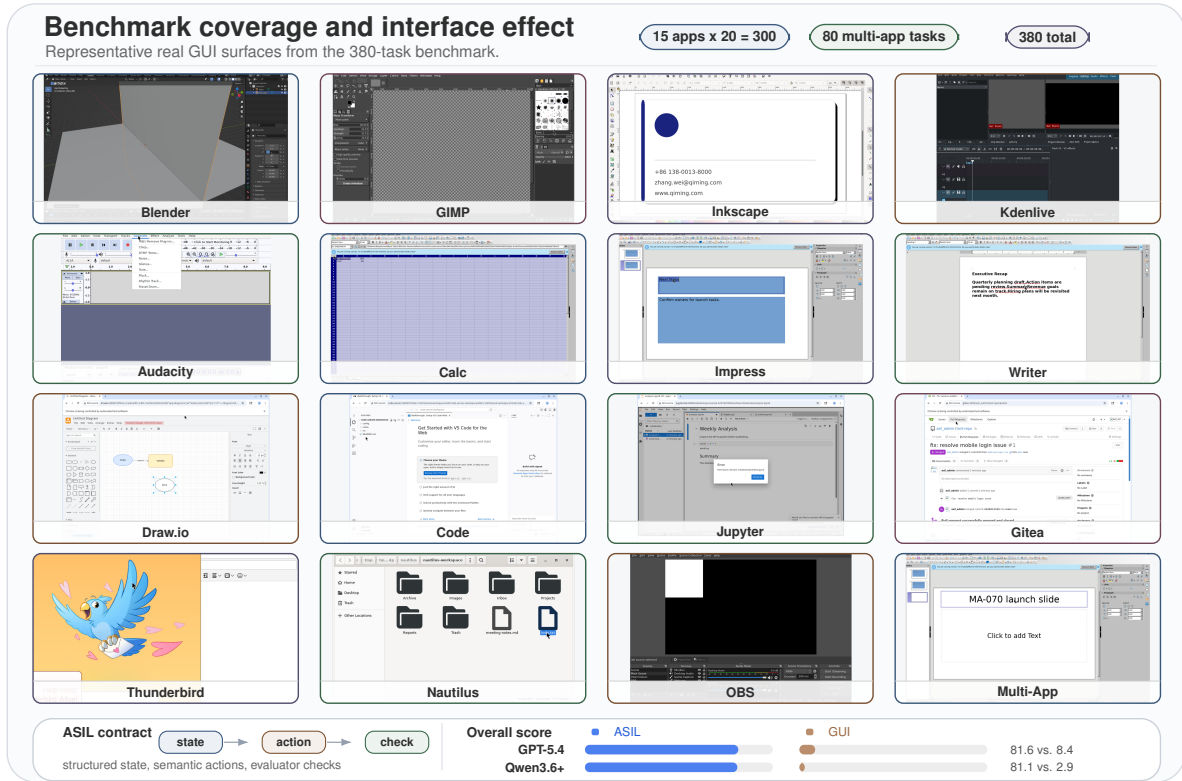

%% file: figures/training_pipeline.tex

\begin{figure*}[t]
\centering
\resizebox{0.88\textwidth}{!}{%
\begin{tikzpicture}[
  x=1cm,
  y=1cm,
  font=\sffamily,
  title/.style={
    font=\sffamily\bfseries\fontsize{14.4}{15.7}\selectfont,
    text=FigureBenchBorder!96!black,
    anchor=west,
  },
  sub/.style={
    font=\sffamily\fontsize{7.2}{8.0}\selectfont,
    text=FigureTextMuted,
    anchor=west,
  },
  laneLabel/.style={
    font=\sffamily\bfseries\fontsize{6.0}{6.6}\selectfont,
    text=FigureTextMuted,
    anchor=center,
  },
  card/.style={
    rounded corners=9pt,
    line width=0.70pt,
    fill=white,
    inner sep=0pt,
  },
  softCard/.style={
    card,
    fill=FigureBenchFill!80!white,
    draw=FigureBenchBorder!34,
  },
  cardTitle/.style={
    font=\sffamily\bfseries\fontsize{7.4}{8.1}\selectfont,
    anchor=west,
    text=FigureTextDark,
  },
  bodyText/.style={
    font=\sffamily\fontsize{5.7}{6.3}\selectfont,
    text=FigureTextMuted,
    anchor=west,
  },
  badge/.style={
    rounded corners=6pt,
    line width=0.52pt,
    fill=white,
    minimum height=0.36cm,
    align=center,
    inner xsep=4pt,
    inner ysep=1.6pt,
    font=\sffamily\bfseries\fontsize{5.35}{5.95}\selectfont,
    text height=1.0ex,
    text depth=.22ex,
  },
  flow/.style={
    -{Latex[length=2.45mm,width=1.82mm]},
    line width=0.86pt,
    draw=FigureBenchBorder!66,
  },
  dataFlow/.style={
    flow,
    draw=FigureMintBorder!82,
  },
  trainFlow/.style={
    flow,
    draw=FigurePeachBorder!88,
  },
  feedback/.style={
    -{Latex[length=2.25mm,width=1.65mm]},
    line width=0.76pt,
    draw=FigureBenchBorder!58,
    dashed,
  },
]

\path[use as bounding box] (0,0) rectangle (18.0,8.75);
\fill[FigureBenchFill, rounded corners=13pt] (0.18,0.18) rectangle (17.82,8.57);
\draw[FigureBenchBorder!32, rounded corners=13pt, line width=0.76pt] (0.18,0.18) rectangle (17.82,8.57);
\fill[FigureBenchBorder!5, rounded corners=13pt] (0.18,7.74) rectangle (17.82,8.57);
\draw[FigureBenchBorder!18, line width=0.42pt] (0.58,7.74) -- (17.42,7.74);

\node[title] at (0.64,8.19) {ASIL traces as reusable learning artifacts};
\node[sub] at (0.66,7.93) {structured observations, semantic actions, artifacts, and evaluator outcomes become reusable training material};
\node[
  badge,
  anchor=east,
  draw=FigureMintBorder!58,
  fill=FigureMintFill!46!white,
  text=FigureMintBorder!82!black,
  minimum width=2.08cm
] at (17.30,8.18) {SFT + RL evaluated};

\node[laneLabel] at (2.30,7.43) {verified trace source};
\node[laneLabel] at (6.64,7.43) {reusable datasets};
\node[laneLabel] at (10.58,7.43) {learning updates};
\node[laneLabel] at (14.72,7.43) {policy + evaluation};

\node[card, draw=FigureMintBorder!62, fill=FigureMintFill!44!white, minimum width=4.00cm, minimum height=5.76cm] (source) at (2.50,4.46) {};
\node[cardTitle, text=FigureMintBorder!80!black] at (0.86,6.92) {Verified ASIL traces};
\node[bodyText] at (0.88,6.65) {state, action, artifact, score};

\draw[rounded corners=6pt, draw=FigureMintBorder!42, fill=white, line width=0.52pt] (0.86,5.87) rectangle (4.14,6.30);
\foreach \x/\c/\lab in {1.12/FigureSkyBorder/obs,1.82/FigurePeachBorder/act,2.52/FigureLilacBorder/chk,3.22/FigureMintBorder/scr} {
  \fill[\c!78] (\x,6.085) circle (0.075);
  \node[font=\sffamily\bfseries\fontsize{3.7}{4.0}\selectfont, text=\c!88!black] at (\x,5.93) {\lab};
}
\draw[-{Latex[length=1.55mm,width=1.10mm]}, draw=FigureMintBorder!72, line width=0.52pt]
  (1.20,6.085) -- (3.72,6.085);

\draw[rounded corners=5pt, draw=FigureSkyBorder!54, fill=white, line width=0.55pt] (0.92,4.80) rectangle (2.18,5.54);
\draw[FigureSkyBorder!72, line width=0.50pt] (1.12,5.36) -- (1.04,5.36) -- (1.04,5.02) -- (1.12,5.02);
\draw[FigureSkyBorder!48, line width=0.46pt] (1.32,5.31) -- +(0.54,0);
\draw[FigureSkyBorder!38, line width=0.46pt] (1.32,5.18) -- +(0.66,0);
\draw[FigureSkyBorder!30, line width=0.46pt] (1.32,5.05) -- +(0.45,0);

\draw[rounded corners=5pt, draw=FigurePeachBorder!58, fill=FigurePeachFill!64!white, line width=0.55pt] (2.82,4.80) rectangle (4.08,5.54);
\draw[FigurePeachBorder!74, line width=0.50pt] (3.06,5.35) -- (3.24,5.18) -- (3.06,5.01);
\draw[FigurePeachBorder!74, line width=0.50pt] (3.84,5.35) -- (3.66,5.18) -- (3.84,5.01);
\draw[FigurePeachBorder!50, line width=0.46pt] (3.34,5.30) -- +(0.32,0);
\draw[FigurePeachBorder!38, line width=0.46pt] (3.34,5.12) -- +(0.24,0);

\foreach \dy/\c in {0.18/FigureLilacBorder,0.09/FigureSkyBorder,0/FigureMintBorder} {
  \draw[rounded corners=4pt, draw=\c!54, fill=white, line width=0.46pt]
    (1.04+\dy,3.82+\dy) rectangle (2.08+\dy,4.30+\dy);
  \draw[\c!44, line width=0.38pt] (1.20+\dy,4.16+\dy) -- +(0.55,0);
  \draw[\c!34, line width=0.38pt] (1.20+\dy,4.02+\dy) -- +(0.68,0);
}

\draw[rounded corners=12pt, draw=FigureMintBorder!54, fill=white, line width=0.55pt] (2.72,3.76) rectangle (4.02,4.36);
\draw[FigureMintBorder!78, line width=0.70pt] (3.04,4.06) -- (3.20,3.90) -- (3.54,4.22);
\node[font=\sffamily\bfseries\fontsize{4.8}{5.2}\selectfont, text=FigureMintBorder!82!black] at (3.46,3.92) {PASS};

\node[badge, draw=FigureMintBorder!52, fill=white, text=FigureMintBorder!82!black, minimum width=1.30cm] at (1.52,2.94) {verified};
\node[badge, draw=FigureSkyBorder!46, fill=FigureSkyFill!58!white, text=FigureSkyBorder!76!black, minimum width=1.38cm] at (3.20,2.94) {replayable};
\draw[rounded corners=4pt, draw=FigureBenchBorder!30, fill=white, line width=0.42pt] (1.72,2.18) rectangle (3.28,2.52);
\draw[FigureBenchBorder!36, line width=0.36pt] (1.94,2.42) -- +(0.94,0);
\draw[FigureBenchBorder!28, line width=0.36pt] (1.94,2.32) -- +(1.10,0);
\draw[FigureBenchBorder!22, line width=0.36pt] (1.94,2.22) -- +(0.72,0);
\node[font=\sffamily\bfseries\fontsize{4.8}{5.2}\selectfont, text=FigureBenchBorder!72!black] at (2.50,2.00) {trace log};

\node[card, draw=FigureSkyBorder!58, fill=FigureSkyFill!55!white, minimum width=3.52cm, minimum height=1.48cm] (sftdata) at (6.70,6.10) {};
\node[cardTitle, text=FigureSkyBorder!76!black] at (5.38,6.62) {SFT datasets};
\node[bodyText] at (5.40,6.37) {SFT-v0 + guided-v2};
\foreach \dx/\dy/\c in {0/0/FigureSkyBorder,0.20/-0.06/FigureMintBorder,0.40/-0.12/FigureCoreBorder} {
  \draw[rounded corners=3pt, draw=\c!52, fill=white, line width=0.45pt] (5.64+\dx,5.70+\dy) rectangle (6.35+\dx,6.08+\dy);
}
\draw[FigureSkyBorder!54, line width=0.42pt] (7.06,5.89) -- (7.92,5.89);
\node[badge, draw=FigureSkyBorder!48, fill=white, text=FigureSkyBorder!78!black, minimum width=1.08cm] at (8.04,5.89) {complete};

\node[card, draw=FigureLilacBorder!58, fill=FigureLilacFill!50!white, minimum width=3.52cm, minimum height=1.48cm] (verifierdata) at (6.70,4.42) {};
\node[cardTitle, text=FigureLilacBorder!76!black] at (5.38,4.94) {Verifier data};
\node[bodyText] at (5.40,4.69) {checks, failures, recoveries};
\draw[FigureMintBorder!76, line width=0.72pt] (5.78,4.22) -- (5.98,4.00) -- (6.38,4.42);
\draw[FigureRoseBorder!72, line width=0.68pt] (6.86,4.38) -- (7.20,4.04);
\draw[FigureRoseBorder!72, line width=0.68pt] (7.20,4.38) -- (6.86,4.04);
\node[badge, draw=FigureLilacBorder!48, fill=white, text=FigureLilacBorder!78!black, minimum width=1.08cm] at (8.04,4.21) {scored};

\node[card, draw=FigureLilacBorder!64, fill=FigureLilacFill!44!white, minimum width=3.52cm, minimum height=1.48cm] (rolloutdata) at (6.70,2.74) {};
\node[cardTitle, text=FigureLilacBorder!78!black] at (5.38,3.26) {Rollout data};
\node[bodyText] at (5.40,3.01) {evaluator-backed sampling};
\draw[rounded corners=8pt, draw=FigureLilacBorder!50, fill=white, line width=0.50pt] (5.78,2.44) rectangle (6.92,2.84);
\draw[-{Latex[length=1.3mm,width=1.0mm]}, draw=FigureLilacBorder!58, line width=0.44pt] (5.96,2.63) arc (180:20:0.28);
\draw[-{Latex[length=1.3mm,width=1.0mm]}, draw=FigureLilacBorder!58, line width=0.44pt] (6.72,2.65) arc (0:-160:0.28);
\draw[FigureLilacBorder!38, line width=0.40pt] (7.10,2.64) -- +(0.62,0);
\node[badge, draw=FigureLilacBorder!48, fill=white, text=FigureLilacBorder!78!black, minimum width=1.08cm] at (8.04,2.53) {collected};

\draw[dataFlow] (4.50,6.10) -- (sftdata.west);
\draw[dataFlow] (4.50,4.42) -- (verifierdata.west);
\draw[dataFlow] (4.50,2.74) -- (rolloutdata.west);

\node[card, draw=FigureSkyBorder!58, fill=white, minimum width=2.82cm, minimum height=1.48cm] (sft) at (10.62,6.10) {};
\node[cardTitle, text=FigureSkyBorder!76!black] at (9.58,6.62) {SFT update};
\node[bodyText] at (9.60,6.37) {behavior cloning};
\draw[rounded corners=8pt, draw=FigureSkyBorder!48, fill=FigureSkyFill!52!white, line width=0.52pt] (9.82,5.72) rectangle (11.42,6.04);
\foreach \x/\h in {10.02/0.10,10.32/0.17,10.62/0.24,10.92/0.14,11.22/0.21} {
  \fill[FigureSkyBorder!64] (\x,5.76) rectangle ++(0.12,\h);
}
\node[badge, draw=FigureMintBorder!50, fill=FigureMintFill!44!white, text=FigureMintBorder!80!black, minimum width=0.96cm] at (10.62,5.45) {done};

\node[card, draw=FigurePeachBorder!58, fill=white, minimum width=2.82cm, minimum height=1.48cm] (reward) at (10.62,4.42) {};
\node[cardTitle, text=FigurePeachBorder!78!black] at (9.58,4.94) {Reward signal};
\node[bodyText] at (9.60,4.69) {rank, score, replay};
\draw[rounded corners=7pt, draw=FigurePeachBorder!42, fill=FigurePeachFill!44!white, line width=0.50pt] (9.82,4.10) rectangle (11.42,4.44);
\draw[FigurePeachBorder!72, line width=0.62pt] (10.02,4.19) -- (10.26,4.32) -- (10.54,4.15) -- (10.86,4.34) -- (11.20,4.22);
\node[badge, draw=FigurePeachBorder!48, fill=white, text=FigurePeachBorder!78!black, minimum width=0.96cm] at (10.62,3.77) {checks};

\node[card, draw=FigurePeachBorder!68, fill=FigurePeachFill!22!white, minimum width=2.82cm, minimum height=1.48cm] (rl) at (10.62,2.74) {};
\draw[rounded corners=9pt, draw=FigurePeachBorder!52, line width=0.62pt, dashed] (9.30,2.08) rectangle (11.94,3.40);
\node[cardTitle, text=FigurePeachBorder!82!black] at (9.58,3.26) {RL update};
\node[bodyText] at (9.60,3.01) {on-policy rollout};
\foreach \x/\y/\lab/\c in {10.14/2.60/p/FigureSkyBorder,10.62/2.82/e/FigureMintBorder,11.10/2.60/r/FigurePeachBorder} {
  \fill[\c!18] (\x,\y) circle (0.13);
  \draw[\c!64, line width=0.44pt] (\x,\y) circle (0.13);
  \node[font=\sffamily\bfseries\fontsize{3.9}{4.2}\selectfont, text=\c!84!black] at (\x,\y) {\lab};
}
\draw[-{Latex[length=1.0mm,width=0.75mm]}, draw=FigurePeachBorder!60, line width=0.40pt] (10.26,2.66) -- (10.50,2.76);
\draw[-{Latex[length=1.0mm,width=0.75mm]}, draw=FigurePeachBorder!60, line width=0.40pt] (10.74,2.76) -- (10.98,2.66);
\draw[-{Latex[length=1.0mm,width=0.75mm]}, draw=FigurePeachBorder!60, line width=0.40pt] (10.96,2.52) .. controls (10.62,2.32) .. (10.28,2.52);
\node[badge, draw=FigureMintBorder!50, fill=FigureMintFill!44!white, text=FigureMintBorder!80!black, minimum width=0.96cm] at (10.62,2.23) {done};

\draw[trainFlow] (sftdata.east) -- (sft.west);
\draw[trainFlow] (verifierdata.east) -- (reward.west);
\draw[trainFlow] (rolloutdata.east) -- (rl.west);

\node[card, draw=FigureCoreBorder!58, fill=FigureCoreFill!42!white, minimum width=3.22cm, minimum height=2.32cm] (policy) at (14.72,5.72) {};
\node[cardTitle, text=FigureCoreBorder!80!black] at (13.48,6.58) {ASIL policy};
\node[bodyText] at (13.50,6.32) {planner over same interface};
\draw[rounded corners=13pt, draw=FigureCoreBorder!46, fill=white, line width=0.58pt] (13.72,5.52) rectangle (15.72,6.02);
\foreach \x/\c in {14.02/FigureSkyBorder,14.48/FigureLilacBorder,14.94/FigurePeachBorder,15.40/FigureMintBorder} {
  \fill[\c!76] (\x,5.77) circle (0.070);
}
\draw[FigureCoreBorder!42, line width=0.46pt] (14.09,5.77) -- (14.41,5.77) -- (14.87,5.77) -- (15.33,5.77);
\draw[FigureCoreBorder!36, line width=0.42pt] (14.48,5.77) -- (14.94,5.95);
\draw[FigureCoreBorder!36, line width=0.42pt] (14.48,5.77) -- (14.94,5.59);
\node[badge, draw=FigureCoreBorder!44, fill=white, text=FigureCoreBorder!80!black, minimum width=1.22cm] at (14.72,5.18) {planner};

\node[card, draw=FigurePeachBorder!58, fill=FigurePeachFill!42!white, minimum width=3.42cm, minimum height=2.02cm] (results) at (14.72,3.28) {};
\node[cardTitle, text=FigurePeachBorder!82!black] at (13.48,3.98) {380-task scores};
\node[bodyText] at (13.50,3.73) {Base -> SFT -> RL};
\foreach \y/\lab/\a/\b/\c in {3.30/2B/58.0/72.1/74.4,2.82/9B/66.6/80.4/82.2} {
  \node[font=\sffamily\bfseries\fontsize{5.0}{5.5}\selectfont, text=FigureTextDark, anchor=east] at (13.82,\y) {\lab};
  \fill[rounded corners=1.4pt, fill=FigureBenchBorder!12] (14.02,\y-0.055) rectangle ++(1.92,0.11);
  \fill[rounded corners=1.4pt, fill=FigureSkyBorder!72] (14.02,\y-0.055) rectangle ++({1.92*\a/100},0.11);
  \fill[rounded corners=1.4pt, fill=FigureMintBorder!72] (14.02,\y-0.205) rectangle ++({1.92*\b/100},0.11);
  \fill[rounded corners=1.4pt, fill=FigurePeachBorder!72] (14.02,\y-0.355) rectangle ++({1.92*\c/100},0.11);
}

\node[card, draw=FigureBenchBorder!54, fill=white, minimum width=3.22cm, minimum height=1.00cm] (eval) at (14.72,1.42) {};
\node[cardTitle, text=FigureBenchBorder!88!black] at (13.48,1.60) {Evaluator replay};
\node[bodyText] at (13.50,1.34) {validate, relabel, reuse};
\draw[FigureMintBorder!76, line width=0.62pt] (15.88,1.45) -- (16.02,1.31) -- (16.26,1.58);

\draw[trainFlow] (sft.east) -- ++(0.38,0) -- (13.08,6.16);
\draw[trainFlow] (reward.east) -- ++(0.32,0) -- (13.08,5.72);
\draw[trainFlow] (rl.east) -- ++(0.38,0) -- (13.08,5.28);
\draw[dataFlow] (policy.south) -- (results.north);
\draw[dataFlow] (results.south) -- (eval.north);
\draw[feedback] (eval.west) -- ++(-9.84,0) |- (source.south);
\node[font=\sffamily\fontsize{5.0}{5.5}\selectfont, text=FigureTextMuted, fill=FigureBenchFill, inner xsep=2.0pt, inner ysep=0.8pt] at (7.38,1.12) {replay / relabel};

\fill[rounded corners=8pt, fill=white, draw=FigureBenchBorder!26, line width=0.48pt] (0.58,0.42) rectangle (17.42,0.92);
\draw[dataFlow] (0.94,0.67) -- +(0.52,0);
\node[bodyText] at (1.62,0.63) {verified data};
\draw[trainFlow] (3.02,0.67) -- +(0.52,0);
\node[bodyText] at (3.70,0.63) {training update};
\draw[feedback] (5.68,0.67) -- +(0.52,0);
\node[bodyText] at (6.36,0.63) {evaluation reuse};
\node[
  badge,
  draw=FigureSkyBorder!42,
  fill=FigureSkyFill!44!white,
  text=FigureCoreBorder!86!black,
  minimum width=2.78cm,
  font=\sffamily\bfseries\fontsize{4.95}{5.45}\selectfont
] at (10.56,0.67) {2B 58.0 -> 72.1 -> 74.4};
\node[
  badge,
  draw=FigureMintBorder!42,
  fill=FigureMintFill!38!white,
  text=FigureMintBorder!84!black,
  minimum width=2.88cm,
  font=\sffamily\bfseries\fontsize{4.95}{5.45}\selectfont
] at (13.84,0.67) {9B 66.6 -> 80.4 -> 82.2};

\end{tikzpicture}%
}
\caption{\textbf{ASIL traces as reusable learning artifacts.} Verified ASIL observations, actions, artifacts, and evaluator outcomes are reusable across supervised fine-tuning and evaluator-backed rollout training. The same loop drives both stages: SFT updates the policy from replayed and guided traces, while on-policy RL updates it from rollouts whose rewards come from the same ASIL evaluator. Both stages produce final 380-task gains on Qwen3.5-2B and Qwen3.5-9B, and the 9B family's training deltas further amplify by 3--4 points on the curated 80-task hard suite reported in Section~\ref{sec:hard-results}.}
\label{fig:training-pipeline}
\end{figure*}
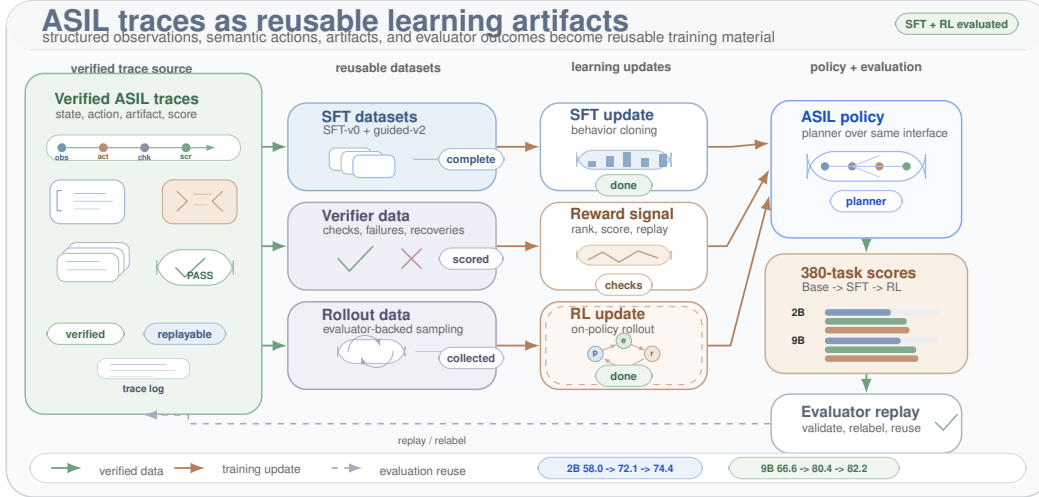

%% file: figures/success_case.tex

\begin{figure*}[t]
\centering
\begingroup
\newcommand{\SuccessLargeImage}[8]{%
  \begin{scope}
    \clip[rounded corners=6pt] (#1,#2) rectangle ++(#3,#4);
    \node[anchor=south west, inner sep=0pt] at (#1,#2)
      {#5};
    \fill[white, opacity=0.90] ($(#1,#2)+(0.13,#4-0.52)$) rectangle ++(#3-0.26,0.38);
    \node[
      anchor=west,
      font=\sffamily\bfseries\fontsize{5.55}{6.1}\selectfont,
      text=#8!88!black
    ] at ($(#1,#2)+(0.26,#4-0.30)$) {#6};
    \node[
      anchor=east,
      font=\sffamily\bfseries\fontsize{4.95}{5.45}\selectfont,
      text=#8!90!black
    ] at ($(#1,#2)+(#3-0.25,#4-0.30)$) {#7};
  \end{scope}
  \draw[rounded corners=6pt, draw=#8!62!black, line width=0.72pt]
    (#1,#2) rectangle ++(#3,#4);
}
\newcommand{\TraceCheck}[3]{%
  \draw[#3!88!black, line width=0.88pt, line cap=round, line join=round]
    ($(#1,#2)+(0.38,0.36)$) -- ($(#1,#2)+(0.48,0.24)$) -- ($(#1,#2)+(0.70,0.50)$);
}
\newcommand{\TraceCross}[3]{%
  \draw[#3!88!black, line width=0.82pt, line cap=round]
    ($(#1,#2)+(0.35,0.23)$) -- ($(#1,#2)+(0.65,0.49)$);
  \draw[#3!88!black, line width=0.82pt, line cap=round]
    ($(#1,#2)+(0.35,0.49)$) -- ($(#1,#2)+(0.65,0.23)$);
}
\newcommand{\SuccessTraceCard}[7]{%
  \fill[rounded corners=6pt, fill=white, draw=#3!68!black, line width=0.64pt]
    (#1,#2) rectangle ++(7.56,1.66);
  \node[
    anchor=west,
    font=\sffamily\bfseries\fontsize{4.80}{5.30}\selectfont,
    text=#3!92!black
  ] at ($(#1,#2)+(0.24,1.36)$) {Thought};
  \node[
    anchor=west,
    font=\sffamily\bfseries\fontsize{4.58}{5.08}\selectfont,
    text=FigureTextDark,
    text width=5.94cm
  ] at ($(#1,#2)+(1.30,1.36)$) {#4};
  \node[
    anchor=west,
    font=\sffamily\bfseries\fontsize{4.80}{5.30}\selectfont,
    text=#3!92!black
  ] at ($(#1,#2)+(0.24,0.82)$) {Action};
  \node[
    anchor=west,
    font=\sffamily\bfseries\fontsize{4.58}{5.08}\selectfont,
    text=FigureTextDark,
    text width=5.94cm
  ] at ($(#1,#2)+(1.30,0.82)$) {#5};
  \fill[rounded corners=4pt, fill=#3!14, draw=#3!68!black, line width=0.48pt]
    ($(#1,#2)+(0.24,0.15)$) rectangle ++(7.08,0.42);
  \ifnum#7=1
    \TraceCheck{#1}{#2}{#3}
  \else
    \TraceCross{#1}{#2}{#3}
  \fi
  \node[
    anchor=west,
    font=\sffamily\bfseries\fontsize{4.55}{5.05}\selectfont,
    text=#3!92!black,
    text width=6.10cm
  ] at ($(#1,#2)+(0.84,0.36)$) {#6};
}
\resizebox{0.96\textwidth}{!}{%
\begin{tikzpicture}[
  x=1cm,
  y=1cm,
  title/.style={font=\sffamily\bfseries\fontsize{15.6}{16.4}\selectfont, text=FigureBenchBorder!95!black, anchor=west},
  sub/.style={font=\sffamily\fontsize{6.35}{7.05}\selectfont, text=FigureTextMuted, anchor=west},
  panelTitle/.style={font=\sffamily\bfseries\fontsize{8.0}{8.7}\selectfont, anchor=west},
  badge/.style={
    rounded corners=6pt,
    line width=0.52pt,
    fill=white,
    inner xsep=4pt,
    inner ysep=2pt,
    font=\sffamily\bfseries\fontsize{5.05}{5.65}\selectfont,
    align=center,
    text height=1.0ex,
    text depth=.22ex
  },
]
\path[use as bounding box] (0,0) rectangle (18.0,8.62);
\fill[rounded corners=13pt, fill=FigureBenchFill!92, draw=FigureBenchBorder!25, line width=0.62pt]
  (0.18,0.16) rectangle (17.82,8.46);
\fill[rounded corners=13pt, fill=FigureBenchBorder!6] (0.18,7.63) rectangle (17.82,8.46);
\draw[FigureBenchBorder!20, line width=0.42pt] (0.56,7.63) -- (17.44,7.63);

\node[title] at (0.62,8.11) {Same task, two interfaces};
\node[sub] at (0.64,7.82) {Instruction: create a payroll sheet with 10 employee rows, unique IDs, non-negative salary fields, and correct Net Pay formula};

\fill[rounded corners=10pt, fill=FigureLegacyFill!82!white, draw=FigureLegacyBorder!52, line width=0.70pt]
  (0.58,0.52) rectangle (8.78,7.36);
\node[panelTitle, text=FigureLegacyBorder!86!black] at (0.92,7.02) {GUI control};
\node[badge, draw=FigureLegacyBorder!55, text=FigureLegacyBorder!88!black, minimum width=2.05cm]
  at (7.42,7.00) {fail after 15};

\SuccessLargeImage{0.90}{2.68}{7.56}{3.86}{\includegraphics[width=7.56cm,height=3.86cm]{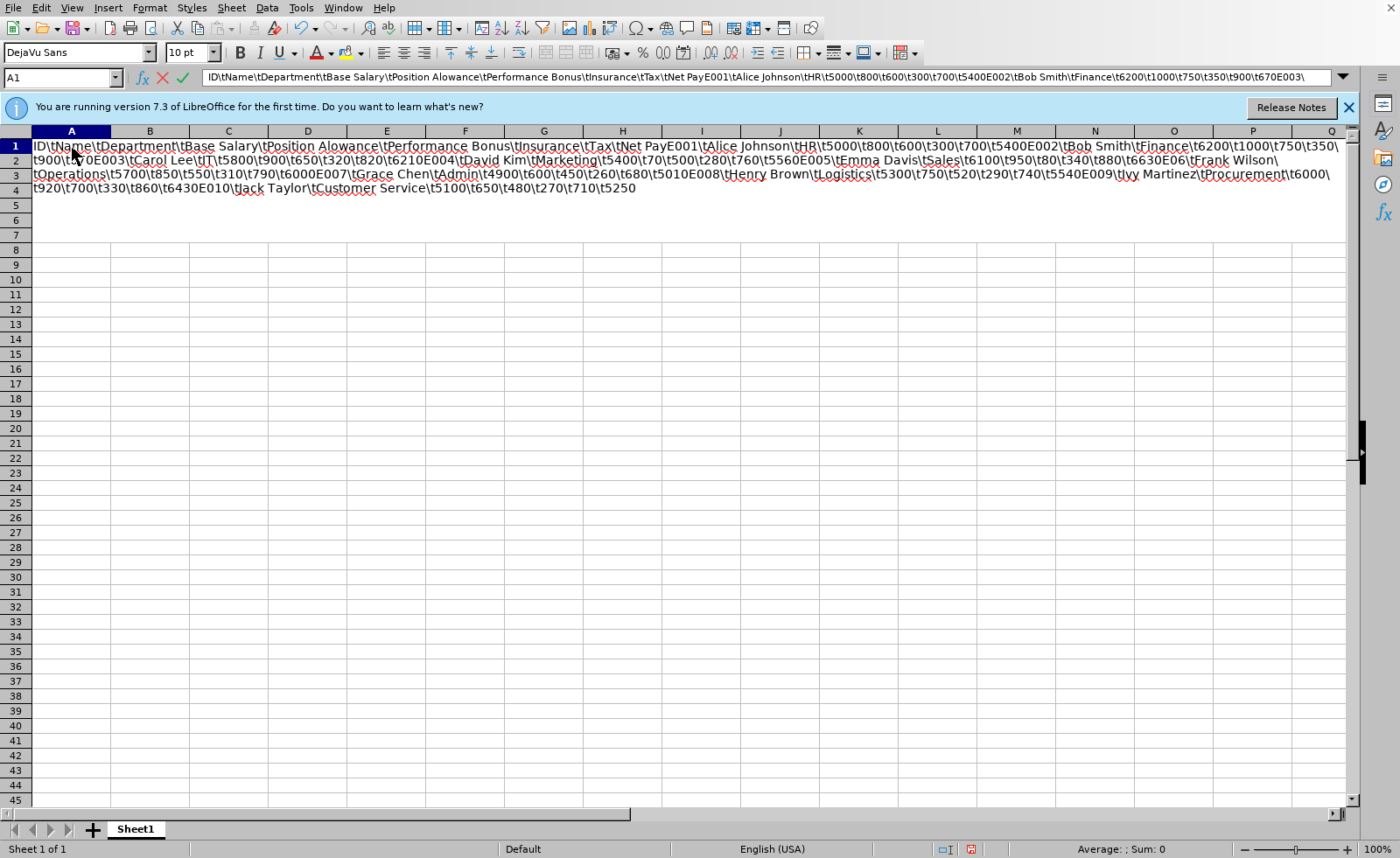}}{GUI result screen}{literal-tab paste state}{FigureLegacyBorder}
\SuccessTraceCard{0.90}{0.74}{FigureLegacyBorder}
  {``A1 is selected; I can paste the full payroll table as tab-separated data.'' Repeats this plan at step 8 after the first paste leaves literal \texttt{\textbackslash t} characters.}
  {\textbf{\texttt{TYPING}} of \texttt{"ID\textbackslash tName\textbackslash t...\textbackslash nE001\textbackslash t..."} into A1; literal \texttt{\textbackslash t} and \texttt{\textbackslash n} are stored as characters, evaluator headers check fails.}
  {Fail reason: GUI typing stores literal escape characters in cells; score 0.0 after 15 events.}
  {0}

\fill[rounded corners=10pt, fill=FigureMintFill!60!white, draw=FigureMintBorder!54, line width=0.70pt]
  (9.22,0.52) rectangle (17.42,7.36);
\node[panelTitle, text=FigureMintBorder!84!black] at (9.56,7.02) {ASIL control};
\node[badge, draw=FigureMintBorder!54, text=FigureMintBorder!84!black, minimum width=2.05cm]
  at (16.10,7.00) {pass in 1};

\SuccessLargeImage{9.54}{2.68}{7.56}{3.86}{\includegraphics[width=7.56cm,height=3.86cm]{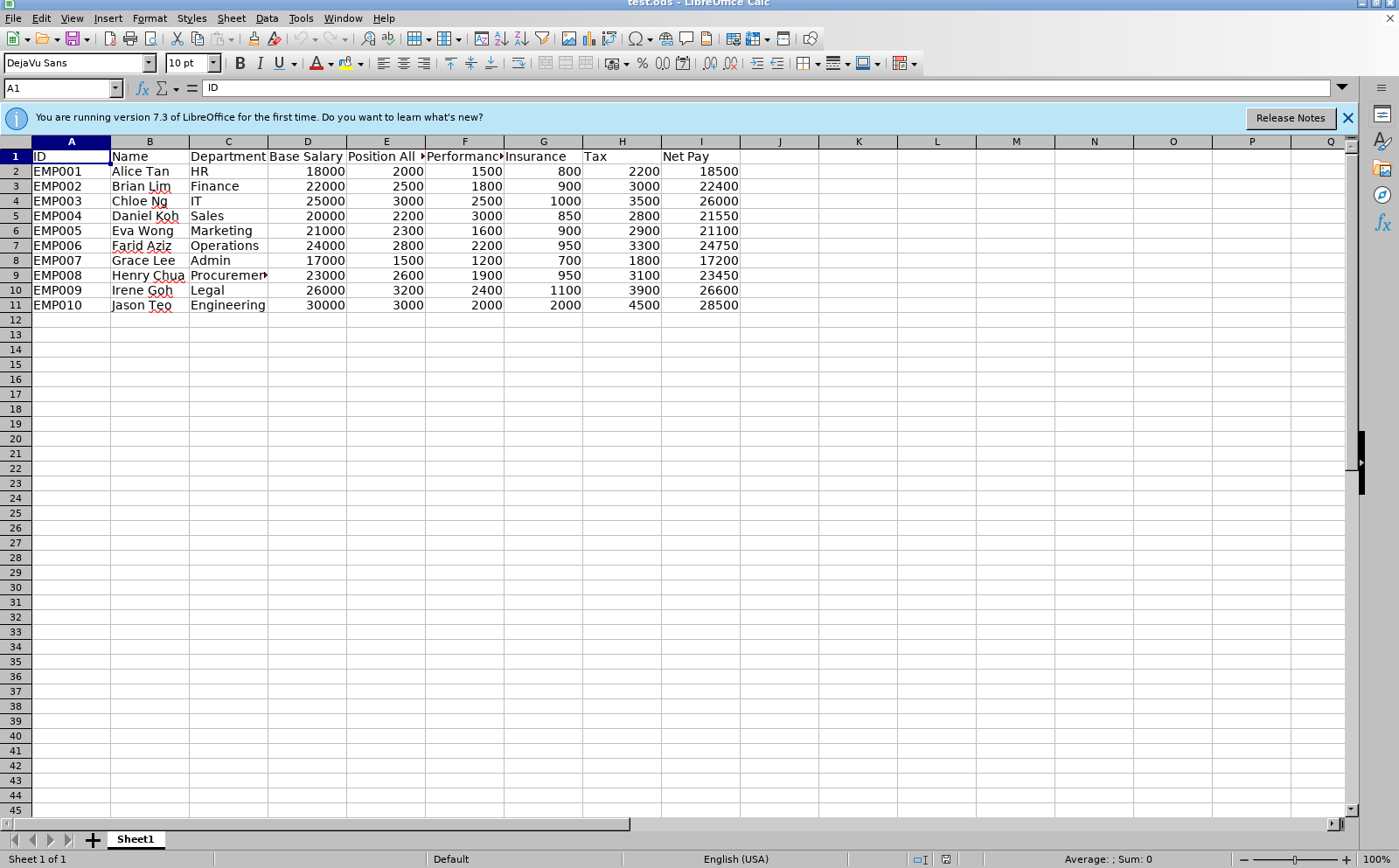}}{ASIL result screen}{validated spreadsheet}{FigureMintBorder}
\SuccessTraceCard{9.54}{0.74}{FigureMintBorder}
  {``Current state does not meet success criteria; write the full payroll table to Sheet1 in one batch with headers and 10 employee rows.''}
  {\textbf{\texttt{modify\_file}} writes 99 cells across \texttt{A1:I11}, with Net Pay $=$ Base $+$ Allowance $+$ Bonus $-$ Insurance $-$ Tax.}
  {Pass reason: semantic action writes spreadsheet state directly; evaluator's 6 checkpoints all satisfied.}
  {1}

\end{tikzpicture}%
}
\endgroup
\caption{\textbf{Same task under GUI control and ASIL control.} GPT-5.4 is run on the libreoffice\_19 payroll task in both modes; thoughts and actions are excerpted verbatim from the official GUI and ASIL evaluation runs. The GUI run identifies a literal-tab paste problem but cannot escape the low-level typing loop and fails after 15 events; ASIL writes the spreadsheet state directly with one \texttt{modify\_file} action and passes.}
\label{fig:success-case}
\end{figure*}
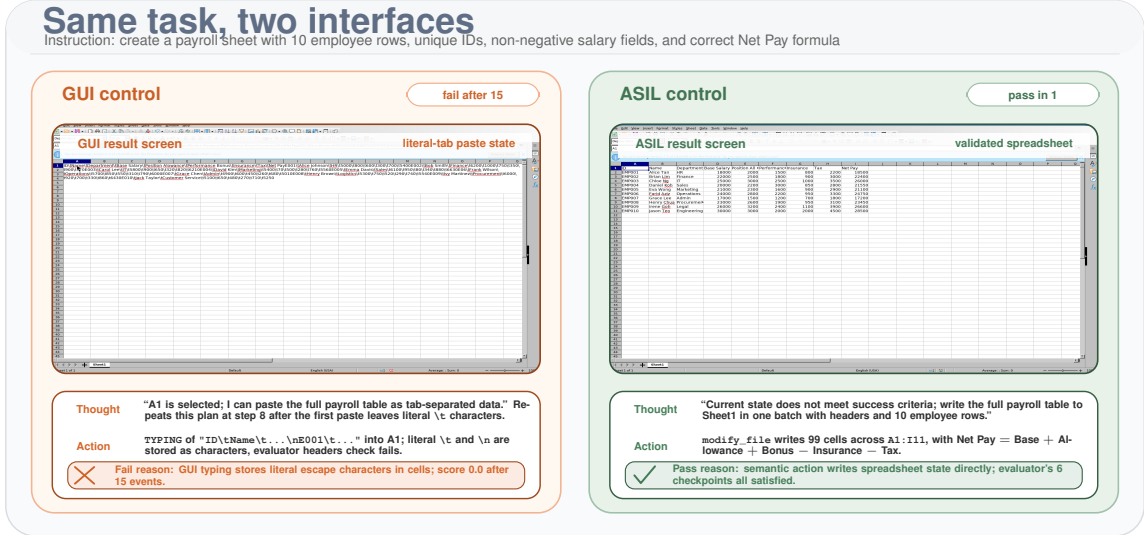

%% file: figures/failure_cases.tex

\begin{figure*}[t]
\centering
\begingroup
\newcommand{\FailurePanel}[4]{%
  \fill[rounded corners=10pt, fill=white, draw=#4!48!black, line width=0.66pt]
    (#1,#2) rectangle ++(5.22,7.28);
  \fill[rounded corners=10pt, fill=#4!10] ($(#1,#2)+(0,6.70)$) rectangle ++(5.22,0.58);
  \node[
    anchor=west,
    font=\sffamily\bfseries\fontsize{5.75}{6.35}\selectfont,
    text=#4!86!black
  ] at ($(#1,#2)+(0.24,7.02)$) {#3};
}
\newcommand{\FailurePreview}[7]{%
  \begin{scope}
    \clip[rounded corners=6pt] (#1,#2) rectangle ++(#3,#4);
    \node[anchor=south west, inner sep=0pt] at (#1,#2)
      {#5};
    \fill[white, opacity=0.91] ($(#1,#2)+(0.13,#4-0.47)$) rectangle ++(#3-0.26,0.34);
    \node[
      anchor=west,
      font=\sffamily\bfseries\fontsize{4.85}{5.35}\selectfont,
      text=#7!88!black
    ] at ($(#1,#2)+(0.25,#4-0.28)$) {#6};
  \end{scope}
  \draw[rounded corners=6pt, draw=#7!62!black, line width=0.58pt]
    (#1,#2) rectangle ++(#3,#4);
}
\newcommand{\FailureCheck}[3]{%
  \draw[#3!88!black, line width=0.76pt, line cap=round, line join=round]
    ($(#1,#2)+(0.18,0.24)$) -- ($(#1,#2)+(0.27,0.13)$) -- ($(#1,#2)+(0.46,0.36)$);
}
\newcommand{\FailureCross}[3]{%
  \draw[#3!88!black, line width=0.72pt, line cap=round]
    ($(#1,#2)+(0.17,0.13)$) -- ($(#1,#2)+(0.43,0.35)$);
  \draw[#3!88!black, line width=0.72pt, line cap=round]
    ($(#1,#2)+(0.17,0.35)$) -- ($(#1,#2)+(0.43,0.13)$);
}
\newcommand{\FailureMetric}[6]{%
  \fill[rounded corners=5pt, fill=#4!14, draw=#4!70!black, line width=0.52pt]
    (#1,#2) rectangle ++(#3,0.46);
  \ifnum#6=1
    \FailureCheck{#1}{#2}{#4}
  \else
    \FailureCross{#1}{#2}{#4}
  \fi
  \node[
    anchor=west,
    font=\sffamily\bfseries\fontsize{4.72}{5.22}\selectfont,
    text=#4!92!black,
    align=center
  ] at ($(#1,#2)+(0.58,0.26)$) {#5};
}
\newcommand{\FailureInstruction}[3]{%
  \node[
    anchor=north west,
    font=\sffamily\bfseries\fontsize{4.35}{4.92}\selectfont,
    text=FigureTextDark,
    text width=4.54cm
  ] at (#1,#2) {\textbf{Instruction:} #3};
}
\newcommand{\FailureTrace}[5]{%
  \fill[rounded corners=6pt, fill=#3!9, draw=#3!62!black, line width=0.56pt]
    (#1,#2) rectangle ++(4.62,1.42);
  \node[
    anchor=west,
    font=\sffamily\bfseries\fontsize{4.45}{4.95}\selectfont,
    text=#3!92!black
  ] at ($(#1,#2)+(0.16,1.14)$) {Thought};
  \node[
    anchor=west,
    font=\sffamily\bfseries\fontsize{4.18}{4.70}\selectfont,
    text=FigureTextDark,
    text width=3.48cm
  ] at ($(#1,#2)+(0.98,1.14)$) {#4};
  \node[
    anchor=west,
    font=\sffamily\bfseries\fontsize{4.45}{4.95}\selectfont,
    text=#3!92!black
  ] at ($(#1,#2)+(0.16,0.48)$) {Action};
  \node[
    anchor=west,
    font=\sffamily\bfseries\fontsize{4.18}{4.70}\selectfont,
    text=FigureTextDark,
    text width=3.48cm
  ] at ($(#1,#2)+(0.98,0.48)$) {#5};
}
\resizebox{\textwidth}{!}{%
\begin{tikzpicture}[
  x=1cm,
  y=1cm,
  title/.style={font=\sffamily\bfseries\fontsize{15.6}{16.4}\selectfont, text=FigureBenchBorder!95!black, anchor=west},
  sub/.style={font=\sffamily\fontsize{6.45}{7.15}\selectfont, text=FigureTextMuted, anchor=west},
]
\path[use as bounding box] (0,0) rectangle (18.0,9.40);
\fill[rounded corners=13pt, fill=FigureBenchFill!92, draw=FigureBenchBorder!25, line width=0.62pt]
  (0.18,0.16) rectangle (17.82,9.24);
\fill[rounded corners=13pt, fill=FigureBenchBorder!6] (0.18,8.43) rectangle (17.82,9.24);
\draw[FigureBenchBorder!20, line width=0.42pt] (0.56,8.43) -- (17.44,8.43);
\node[title] at (0.62,8.90) {Failure modes under ASIL};
\node[sub] at (0.64,8.61) {Each panel pairs a concrete task instruction, thought/action evidence, result preview, and outcome pattern};

\FailurePanel{0.58}{0.64}{(a) Access-path limit}{FigureRoseBorder}
\node[
  anchor=east,
  font=\sffamily\fontsize{4.35}{4.85}\selectfont,
  text=FigureTextMuted
] at (5.56,7.66) {GIMP real-photo};
\FailurePreview{0.82}{4.64}{4.74}{2.76}{\includegraphics[width=4.74cm,height=2.76cm]{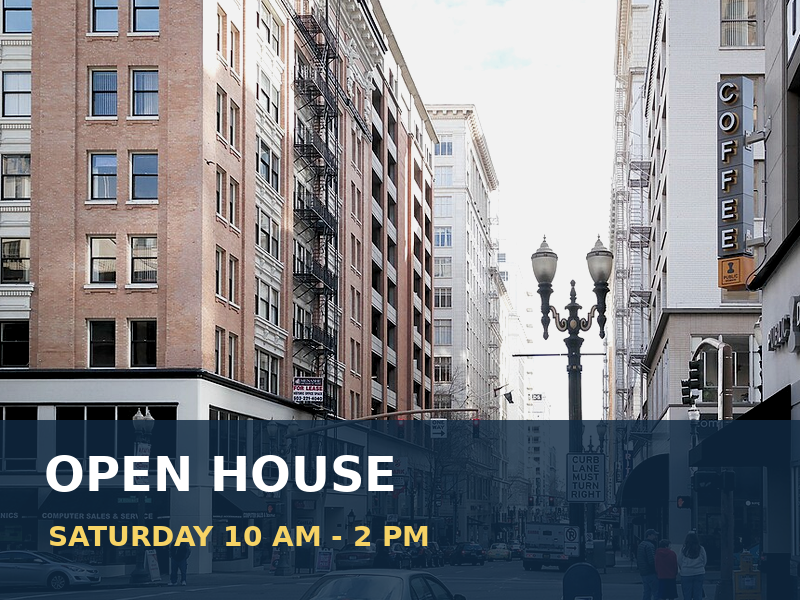}}{banner result preview}{FigureRoseBorder}
\FailureInstruction{0.86}{4.32}{Create an 800x600 cafe banner from a real street photo: crop, reduce saturation, add a navy band, title ``OPEN HOUSE'', and detail ``SATURDAY 10 AM - 2 PM''.}
\FailureTrace{0.86}{2.16}{FigureRoseBorder}
  {Plan covers crop, color correction, band, title, and detail; final correctness depends on pixel-level composition.}
  {\textbf{\texttt{invoke\_function}} sequence: street-photo layer, navy band, title text, detail text.}
\FailureMetric{0.86}{1.44}{2.18}{FigureRoseBorder}{0/24 full passes}{0}
\FailureMetric{3.20}{1.44}{2.28}{FigureRoseBorder}{best avg 0.1515}{0}
\FailureMetric{0.86}{0.84}{4.62}{FigureRoseBorder}{reason: pixel-level perception}{0}

\FailurePanel{6.40}{0.64}{(b) Data-coverage gap}{FigurePeachBorder}
\node[
  anchor=east,
  font=\sffamily\fontsize{4.35}{4.85}\selectfont,
  text=FigureTextMuted
] at (11.38,7.66) {LibreOffice Calc};
\FailurePreview{6.64}{4.64}{4.74}{2.76}{\includegraphics[width=4.74cm,height=2.76cm]{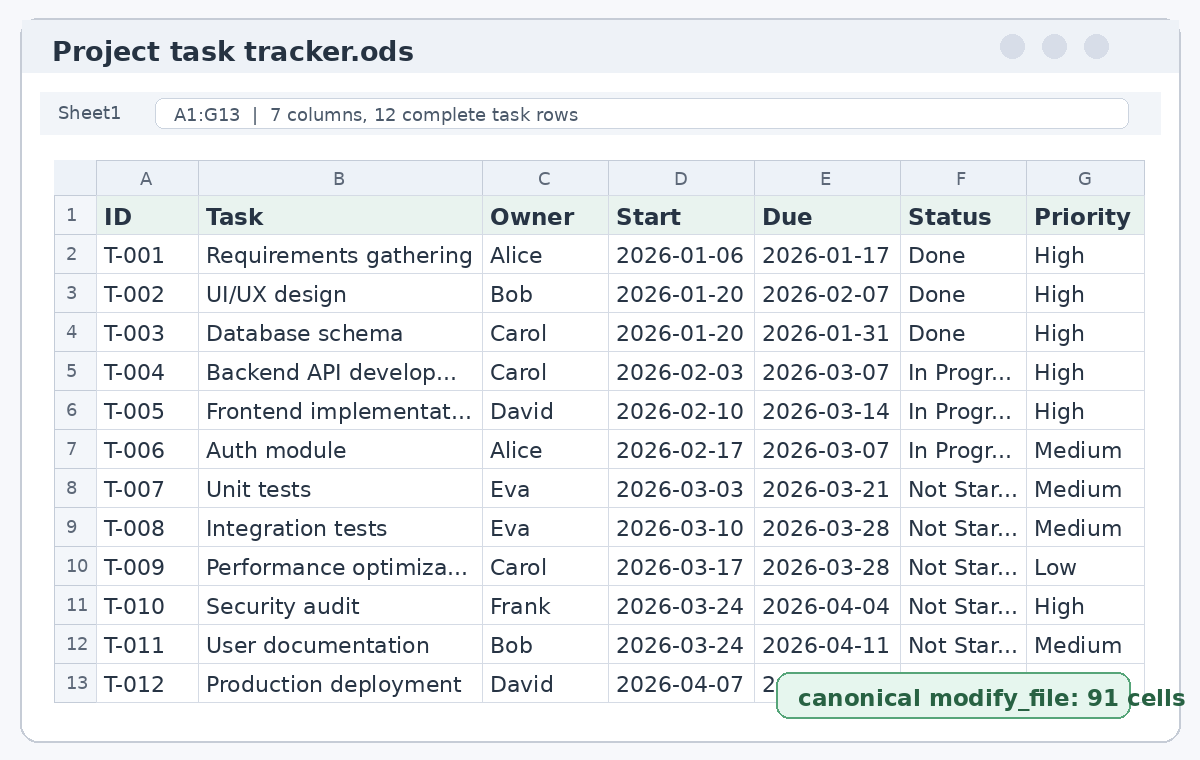}}{tracker result preview}{FigurePeachBorder}
\FailureInstruction{6.68}{4.32}{Create a 12-row task tracker (ID, Task, Owner, Start, Due, Status, Priority) with unique IDs, ISO-formatted dates, and Due on or after Start.}
\FailureTrace{6.68}{2.16}{FigurePeachBorder}
  {Need unique IDs, complete rows, ISO dates, and \texttt{Due $\geq$ Start}.}
  {\textbf{\texttt{modify\_file}} writes 91 cells across \texttt{A1:G13}; Qwen rows produce malformed dates or missing IDs.}
\FailureMetric{6.68}{1.44}{2.18}{FigureMintBorder}{GPT-5.4: 3/3}{1}
\FailureMetric{9.02}{1.44}{2.28}{FigurePeachBorder}{Qwen-9B: 0/3}{0}
\FailureMetric{6.68}{0.84}{4.62}{FigurePeachBorder}{reason: spreadsheet pattern gap}{0}

\FailurePanel{12.22}{0.64}{(c) Training regression}{FigureLimeBorder}
\node[
  anchor=east,
  font=\sffamily\fontsize{4.35}{4.85}\selectfont,
  text=FigureTextMuted
] at (17.20,7.66) {Thunderbird};
\FailurePreview{12.46}{4.64}{4.74}{2.76}{\includegraphics[width=4.74cm,height=2.76cm]{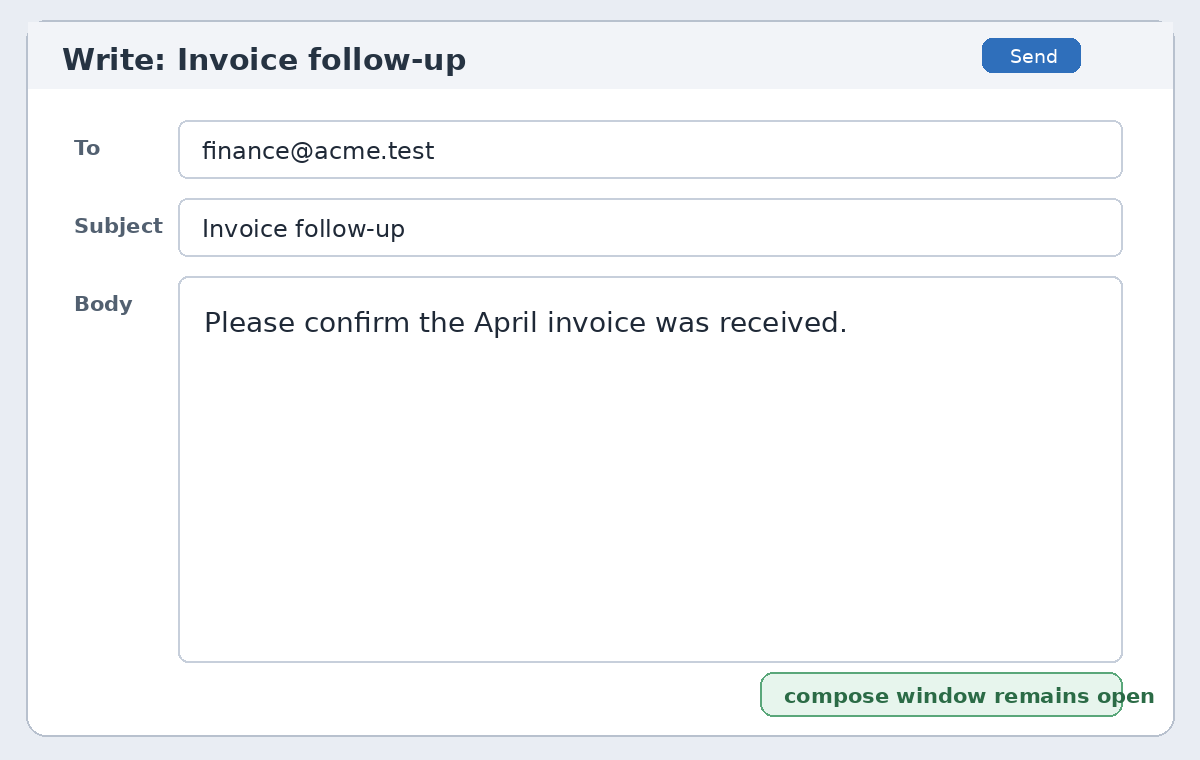}}{compose result preview}{FigureLimeBorder}
\FailureInstruction{12.50}{4.32}{Update the open draft (to: \texttt{finance@acme.test}, subject: ``Invoice follow-up'', body: confirm April invoice) and keep the compose window open.}
\FailureTrace{12.50}{2.16}{FigureLimeBorder}
  {One \texttt{update\_draft} call sets recipient, subject, body and keeps the window open.}
  {\textbf{\texttt{invoke\_function}} with op \texttt{update\_draft} sets all three fields in one call.}
\FailureMetric{12.50}{1.44}{1.54}{FigureMintBorder}{Base 5/5}{1}
\FailureMetric{14.06}{1.44}{1.54}{FigureLimeBorder}{SFT 0/5}{0}
\FailureMetric{15.62}{1.44}{1.54}{FigureMintBorder}{RL 5/5}{1}
\FailureMetric{12.50}{0.84}{4.62}{FigureLimeBorder}{reason: SFT regression}{0}

\end{tikzpicture}%
}
\endgroup
\caption{\textbf{Representative failure modes under ASIL.} Each panel pairs a real hard-suite task instruction with a canonical thought/action sketch and the aggregate pass counts observed in the seven-model hard-suite evaluation. Residual errors remain after screenshot grounding is removed: real-photo editing exposes access-path limits, spreadsheet tasks expose data-coverage gaps, and Thunderbird shows a supervised-training regression that RL later recovers. Thought/action snippets are abbreviated from the canonical task action specifications; aggregate pass counts come from the hard-suite evaluation report (Section~\ref{sec:hard-results}).}
\label{fig:failure-cases}
\end{figure*}
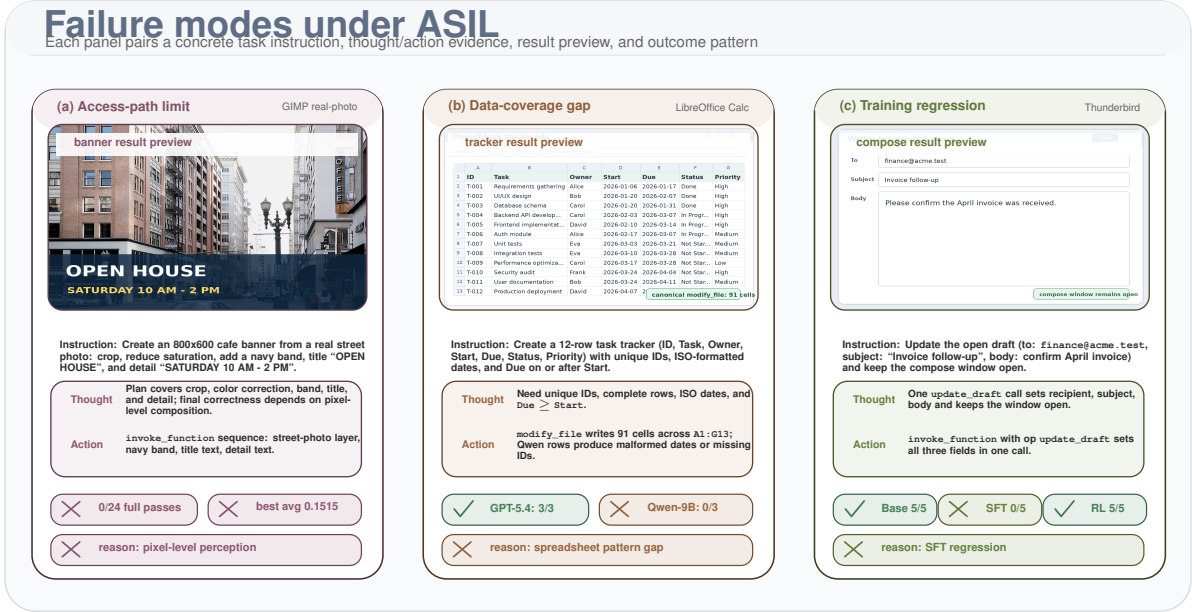

%% file: figures/realization_patterns.tex

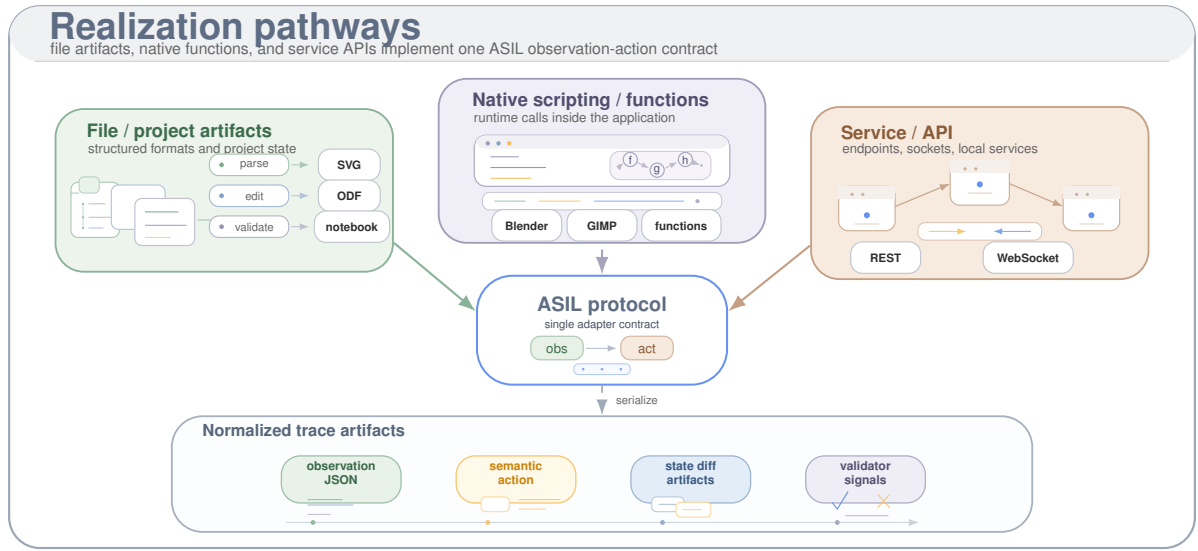
\begin{figure*}[t]
\centering
\resizebox{\textwidth}{!}{%
\begin{tikzpicture}[
  x=1cm,
  y=1cm,
  font=\sffamily,
  title/.style={
    font=\sffamily\bfseries\fontsize{15.6}{16.4}\selectfont,
    text=FigureBenchBorder!94!black,
    anchor=west,
  },
  sub/.style={
    font=\sffamily\fontsize{6.7}{7.4}\selectfont,
    text=FigureTextMuted,
    anchor=west,
  },
  pathCard/.style={
    rounded corners=12pt,
    line width=0.78pt,
    fill=white,
    inner sep=0pt,
  },
  pathTitle/.style={
    font=\sffamily\bfseries\fontsize{7.8}{8.5}\selectfont,
    anchor=west,
    text=FigureTextDark,
  },
  small/.style={
    font=\sffamily\fontsize{5.6}{6.2}\selectfont,
    text=FigureTextMuted,
  },
  tiny/.style={
    font=\sffamily\fontsize{4.8}{5.4}\selectfont,
    text=FigureTextMuted,
  },
  badge/.style={
    rounded corners=5pt,
    draw=FigureBenchBorder!42,
    fill=white,
    line width=0.48pt,
    minimum width=1.04cm,
    minimum height=0.34cm,
    align=center,
    inner xsep=2pt,
    font=\sffamily\bfseries\fontsize{4.9}{5.5}\selectfont,
    text=FigureTextDark,
    text height=1.0ex,
    text depth=.22ex,
  },
  hub/.style={
    rounded corners=15pt,
    draw=FigureCoreBorder!70,
    fill=white,
    line width=0.86pt,
    minimum width=3.72cm,
    minimum height=1.62cm,
    inner sep=0pt,
  },
  outputBox/.style={
    rounded corners=8pt,
    line width=0.56pt,
    fill=white,
    minimum width=1.78cm,
    minimum height=0.70cm,
    align=center,
    font=\sffamily\bfseries\fontsize{5.15}{5.8}\selectfont,
    text=FigureTextDark,
    text height=1.0ex,
    text depth=.22ex,
  },
  flow/.style={
    -{Latex[length=2.65mm,width=1.95mm]},
    line width=0.90pt,
    draw=FigureBenchBorder!76,
  },
  softFlow/.style={
    -{Latex[length=2.35mm,width=1.70mm]},
    line width=0.74pt,
    draw=FigureBenchBorder!62,
    dashed,
  },
]

\path[use as bounding box] (0,0) rectangle (18.0,8.48);
\fill[white, rounded corners=13pt] (0.18,0.18) rectangle (17.82,8.24);
\draw[FigureBenchBorder!64, rounded corners=13pt, line width=0.92pt] (0.18,0.18) rectangle (17.82,8.24);
\fill[FigureBenchBorder!10, rounded corners=13pt] (0.18,7.42) rectangle (17.82,8.24);
\draw[FigureBenchBorder!34, line width=0.48pt] (0.56,7.42) -- (17.44,7.42);
\node[title] at (0.64,7.90) {Realization pathways};
\node[sub] at (0.66,7.59) {file artifacts, native functions, and service APIs implement one ASIL observation-action contract};

\node[hub] (hub) at (9.00,3.42) {};
\node[
  font=\sffamily\bfseries\fontsize{8.4}{9.1}\selectfont,
  text=FigureBenchBorder!94!black,
  align=center
] at (9.00,3.78) {ASIL protocol};
\node[tiny, text=FigureBenchBorder!70!black] at (9.00,3.50) {single adapter contract};
\draw[rounded corners=4pt, draw=FigureMintBorder!46, fill=FigureMintFill!64!white, line width=0.50pt]
  (7.94,2.98) rectangle (8.72,3.33);
\draw[rounded corners=4pt, draw=FigurePeachBorder!50, fill=FigurePeachFill!62!white, line width=0.50pt]
  (9.28,2.98) rectangle (10.06,3.33);
\node[small, text=FigureMintBorder!78!black] at (8.33,3.15) {obs};
\node[small, text=FigurePeachBorder!82!black] at (9.67,3.15) {act};
\draw[FigureBenchBorder!48, line width=0.42pt, -{Latex[length=1.35mm,width=1.0mm]}]
  (8.76,3.15) -- (9.22,3.15);
\draw[rounded corners=2pt, draw=FigureCoreBorder!36, fill=FigureCoreFill!55!white, line width=0.34pt]
  (8.58,2.76) rectangle (9.42,2.92);
\foreach \x in {8.72,9.00,9.28} {
  \fill[FigureCoreBorder!55] (\x,2.84) circle (0.020);
}

\node[pathCard, draw=FigureMintBorder!66, fill=FigureMintFill!50!white, minimum width=5.02cm, minimum height=2.42cm] (file) at (3.38,5.50) {};
\node[pathTitle, text=FigureMintBorder!78!black] at (1.20,6.37) {File / project artifacts};
\node[small, anchor=west] at (1.22,6.11) {structured formats and project state};
\draw[rounded corners=3pt, draw=FigureMintBorder!48, fill=white, line width=0.46pt]
  (1.10,4.80) rectangle (1.82,5.62);
\draw[rounded corners=2pt, draw=FigureMintBorder!42, fill=FigureMintFill!55!white, line width=0.36pt]
  (1.22,5.46) rectangle (1.52,5.70);
\draw[FigureMintBorder!46, line width=0.34pt] (1.22,5.46) -- (1.10,5.46);
\draw[FigureMintBorder!34, line width=0.34pt] (1.28,5.30) -- (1.28,4.94);
\foreach \y/\w/\c in {5.30/0.34/FigureMintBorder,5.10/0.46/FigureMintBorder,4.94/0.30/FigureSkyBorder} {
  \fill[\c!56] (1.28,\y) circle (0.016);
  \draw[\c!46, line width=0.34pt] (1.36,\y) -- +(0.74*\w,0);
}
\draw[rounded corners=3pt, draw=FigureSkyBorder!42, fill=white, line width=0.50pt] (1.70,4.92) rectangle (2.50,5.58);
\draw[rounded corners=3pt, draw=FigureLilacBorder!42, fill=white, line width=0.50pt] (2.05,4.68) rectangle (2.94,5.38);
\draw[FigureMintBorder!68, line width=0.48pt] (2.20,5.18) -- +(0.50,0);
\draw[FigureSkyBorder!58, line width=0.48pt] (2.20,4.98) -- +(0.62,0);
\draw[FigureBackendBorder!50, line width=0.42pt] (2.20,4.80) -- +(0.38,0);
\draw[FigureMintBorder!54, line width=0.42pt] (2.98,5.06) -- (3.18,5.06);
\foreach \y/\t/\c in {5.88/parse/FigureMintBorder,5.42/edit/FigureSkyBorder,4.96/validate/FigureLilacBorder} {
  \draw[rounded corners=4pt, draw=\c!50, fill=white, line width=0.48pt] (3.16,\y-0.16) rectangle (4.34,\y+0.16);
  \fill[\c!72] (3.34,\y) circle (0.035);
  \node[tiny, anchor=center] at (3.83,\y) {\t};
}
\draw[-{Latex[length=1.35mm,width=0.98mm]}, draw=FigureMintBorder!48, line width=0.36pt] (4.38,5.88) -- (4.64,5.88);
\draw[-{Latex[length=1.35mm,width=0.98mm]}, draw=FigureSkyBorder!44, line width=0.36pt] (4.38,5.42) -- (4.64,5.42);
\draw[-{Latex[length=1.35mm,width=0.98mm]}, draw=FigureLilacBorder!44, line width=0.36pt] (4.38,4.96) -- (4.64,4.96);
\node[badge, minimum width=0.92cm] at (5.24,5.88) {SVG};
\node[badge, minimum width=0.92cm] at (5.24,5.42) {ODF};
\node[badge, minimum width=1.12cm] at (5.28,4.96) {notebook};

\node[pathCard, draw=FigureLilacBorder!66, fill=FigureLilacFill!52!white, minimum width=4.86cm, minimum height=2.44cm] (script) at (9.00,5.94) {};
\node[pathTitle, text=FigureLilacBorder!74!black] at (6.94,6.82) {Native scripting / functions};
\node[small, anchor=west] at (6.96,6.56) {runtime calls inside the application};
	\draw[rounded corners=5pt, draw=FigureLilacBorder!54, fill=white, line width=0.54pt] (7.10,5.56) rectangle (10.90,6.32);
	\fill[FigureLilacBorder!10, rounded corners=5pt] (7.10,6.12) rectangle (10.90,6.32);
	\foreach \x/\c in {7.30/FigureLilacBorder,7.46/FigureSkyBorder,7.62/FigureBackendBorder} {
	  \fill[\c!68] (\x,6.20) circle (0.032);
	}
	\foreach \y/\w/\c in {6.00/0.54/FigureLilacBorder,5.84/1.06/FigureCoreBorder,5.68/0.78/FigureBackendBorder} {
	  \draw[\c!62, line width=0.48pt] (7.34,\y) -- +(0.78*\w,0);
	}
		\draw[rounded corners=4pt, draw=FigureLilacBorder!36, fill=FigureLilacFill!32!white, line width=0.38pt]
		  (9.12,5.66) rectangle (10.62,6.08);
		\fill[FigureLilacBorder!42] (9.26,5.87) circle (0.022);
		\fill[FigureLilacBorder!42] (10.48,5.87) circle (0.022);
		\foreach \x/\y/\t in {9.42/5.96/f,9.82/5.80/g,10.24/5.96/h} {
		  \draw[FigureLilacBorder!58, line width=0.52pt] (\x,\y) circle (0.105);
		  \node[tiny, text=FigureLilacBorder!78!black] at (\x,\y) {\t};
		}
		\draw[FigureLilacBorder!46, line width=0.42pt, -{Latex[length=1.15mm,width=0.85mm]}] (9.28,5.87) -- (9.32,5.91);
		\draw[FigureLilacBorder!46, line width=0.42pt, -{Latex[length=1.15mm,width=0.85mm]}] (9.52,5.92) -- (9.70,5.83);
		\draw[FigureLilacBorder!46, line width=0.42pt, -{Latex[length=1.15mm,width=0.85mm]}] (9.94,5.83) -- (10.12,5.92);
		\draw[FigureLilacBorder!46, line width=0.42pt, -{Latex[length=1.15mm,width=0.85mm]}] (10.34,5.92) -- (10.46,5.88);
		\draw[rounded corners=3pt, draw=FigureLilacBorder!38, fill=white, line width=0.42pt] (7.22,5.21) rectangle (10.78,5.47);
		\draw[FigureMintBorder!54, line width=0.38pt] (7.40,5.34) -- +(0.46,0);
		\draw[FigureBackendBorder!54, line width=0.38pt] (8.06,5.34) -- +(0.62,0);
		\draw[FigureCoreBorder!54, line width=0.38pt] (8.88,5.34) -- +(1.34,0);
		\fill[FigureLilacBorder!56] (10.42,5.34) circle (0.030);
		\node[badge] at (7.88,4.98) {Blender};
		\node[badge] at (9.00,4.98) {GIMP};
		\node[badge, minimum width=1.18cm] at (10.18,4.98) {functions};

\node[pathCard, draw=FigurePeachBorder!66, fill=FigurePeachFill!56!white, minimum width=5.02cm, minimum height=2.56cm] (api) at (14.62,5.46) {};
\node[pathTitle, text=FigurePeachBorder!84!black] at (12.42,6.37) {Service / API};
\node[small, anchor=west] at (12.44,6.11) {endpoints, sockets, local services};
\draw[rounded corners=4pt, draw=FigurePeachBorder!54, fill=white, line width=0.50pt] (12.52,4.90) rectangle (13.36,5.56);
\draw[rounded corners=4pt, draw=FigurePeachBorder!54, fill=white, line width=0.50pt] (14.18,5.28) rectangle (15.06,5.94);
\draw[rounded corners=4pt, draw=FigurePeachBorder!54, fill=white, line width=0.50pt] (15.86,4.90) rectangle (16.70,5.56);
\fill[FigurePeachBorder!10] (12.52,5.38) rectangle (13.36,5.56);
\fill[FigurePeachBorder!10] (14.18,5.76) rectangle (15.06,5.94);
\fill[FigurePeachBorder!10] (15.86,5.38) rectangle (16.70,5.56);
\foreach \x/\y in {12.72/5.44,12.88/5.44,14.40/5.82,14.56/5.82,16.06/5.44,16.22/5.44} {
  \fill[FigurePeachBorder!54] (\x,\y) circle (0.024);
}
\foreach \x/\y in {12.94/5.13,14.62/5.59,16.28/5.13} {
  \fill[FigureCoreBorder!70] (\x,\y) circle (0.045);
  \draw[FigurePeachBorder!34, line width=0.30pt] (\x-0.18,\y-0.13) -- +(0.36,0);
}
\draw[-{Latex[length=1.55mm,width=1.10mm]}, draw=FigurePeachBorder!62, line width=0.52pt] (13.36,5.26) -- (14.18,5.60);
\draw[-{Latex[length=1.55mm,width=1.10mm]}, draw=FigurePeachBorder!62, line width=0.52pt] (15.06,5.60) -- (15.86,5.26);
\draw[rounded corners=3pt, draw=FigurePeachBorder!42, fill=white, line width=0.40pt] (13.70,4.76) rectangle (15.54,5.02);
\draw[FigureBackendBorder!58, line width=0.40pt, -{Latex[length=1.32mm,width=0.95mm]}] (13.86,4.89) -- (14.42,4.89);
\draw[FigureCoreBorder!58, line width=0.40pt, -{Latex[length=1.32mm,width=0.95mm]}] (15.38,4.89) -- (14.82,4.89);
\node[badge] at (13.22,4.50) {REST};
\node[badge, minimum width=1.34cm] at (15.34,4.50) {WebSocket};

	\draw[flow, draw=FigureMintBorder!66] ([yshift=-0.78cm]file.east) -- ([yshift=0.28cm]hub.west);
	\draw[flow, draw=FigureLilacBorder!62] (9.00,4.62) -- (hub.north);
	\draw[flow, draw=FigurePeachBorder!66] ([yshift=-0.78cm]api.west) -- ([yshift=0.28cm]hub.east);

\node[
  rounded corners=13pt,
  draw=FigureBenchBorder!56,
  fill=FigureBenchFill!46!white,
  line width=0.78pt,
  minimum width=12.80cm,
  minimum height=1.72cm,
  inner sep=0pt
] (contract) at (9.00,1.28) {};
\node[font=\sffamily\bfseries\fontsize{7.0}{7.6}\selectfont, text=FigureBenchBorder!88!black, anchor=west]
  at (2.92,1.94) {Normalized trace artifacts};

	\draw[FigureBenchBorder!32, line width=0.42pt, -{Latex[length=1.55mm,width=1.10mm]}]
	  (4.30,0.56) -- (13.72,0.56);
	\foreach \x/\c in {4.70/FigureMintBorder,7.30/FigureBackendBorder,9.90/FigureSkyBorder,12.50/FigureLilacBorder} {
	  \fill[\c!60] (\x,0.56) circle (0.036);
	}

\node[outputBox, draw=FigureMintBorder!54, fill=FigureMintFill!58!white, text=FigureMintBorder!76!black] (obsOut) at (5.12,1.20) {observation\\JSON};
\draw[FigureMintBorder!58, line width=0.40pt] (4.62,0.90) -- +(0.52,0);
\draw[FigureSkyBorder!54, line width=0.40pt] (4.62,0.82) -- +(0.68,0);
\draw[FigureLilacBorder!44, line width=0.36pt] (4.62,0.68) -- +(0.34,0);

\node[outputBox, draw=FigureBackendBorder!58, fill=FigureBackendFill!72!white, text=FigureBackendBorder!82!black] (actOut) at (7.72,1.20) {semantic\\action};
\draw[rounded corners=2pt, draw=FigureBackendBorder!42, fill=white, line width=0.34pt] (7.20,0.72) rectangle (7.62,0.94);
\draw[FigureBackendBorder!60, line width=0.42pt] (7.76,0.90) -- +(0.62,0);
\draw[FigureBackendBorder!48, line width=0.42pt] (7.76,0.76) -- +(0.40,0);

\node[outputBox, draw=FigureSkyBorder!54, fill=FigureSkyFill!68!white, text=FigureSkyBorder!76!black] (artOut) at (10.32,1.20) {state diff\\artifacts};
\draw[rounded corners=2pt, draw=FigureSkyBorder!42, fill=white, line width=0.36pt] (9.74,0.72) rectangle (10.22,0.94);
\draw[rounded corners=2pt, draw=FigureBackendBorder!42, fill=white, line width=0.36pt] (10.10,0.64) rectangle (10.62,0.86);
\draw[FigureSkyBorder!48, line width=0.36pt] (9.84,0.83) -- +(0.26,0);
\draw[FigureBackendBorder!44, line width=0.34pt] (10.20,0.75) -- +(0.28,0);

\node[outputBox, draw=FigureLilacBorder!54, fill=FigureLilacFill!54!white, text=FigureLilacBorder!76!black] (valOut) at (12.92,1.20) {validator\\signals};
\draw[FigureCoreBorder!70, line width=0.48pt] (12.42,0.86) -- (12.50,0.76) -- (12.66,1.02);
\draw[FigureBackendBorder!58, line width=0.42pt] (13.10,0.78) -- (13.28,0.98);
\draw[FigureBackendBorder!58, line width=0.42pt] (13.28,0.78) -- (13.10,0.98);
\draw[FigureLilacBorder!50, line width=0.36pt] (12.62,0.66) -- +(0.64,0);

\draw[softFlow] (hub.south) -- node[midway, right=2pt, tiny] {serialize} (contract.north);

\end{tikzpicture}%
}
\caption{\textbf{Realization pathways.} ASIL selects the deepest feasible access path for each software environment while keeping one observation-action protocol; file artifacts, native functions, and service APIs all serialize to the same reusable trace artifacts.}
\label{fig:realization-patterns}
\end{figure*}

%% file: refs.bib
@inproceedings{sweagent,
  title = {{SWE-agent}: Agent-Computer Interfaces Enable Automated Software Engineering},
  author = {Yang, John and Jimenez, Carlos E. and Wettig, Alexander and Lieret, Kilian and Yao, Shunyu and Narasimhan, Karthik and Press, Ofir},
  booktitle = {Advances in Neural Information Processing Systems},
  volume = {37},
  year = {2024},
  doi = {10.52202/079017-1601},
  url = {https://proceedings.neurips.cc/paper_files/paper/2024/hash/5a7c947568c1b1328ccc5230172e1e7c-Abstract-Conference.html},
}

@misc{codeact,
  title = {Executable Code Actions Elicit Better {LLM} Agents},
  author = {Wang, Xingyao and Chen, Yangyi and Yuan, Lifan and Zhang, Yizhe and Li, Yunzhu and Peng, Hao and Ji, Heng},
  year = {2024},
  eprint = {2402.01030},
  archivePrefix = {arXiv},
  primaryClass = {cs.CL},
  doi = {10.48550/arXiv.2402.01030},
  url = {https://arxiv.org/abs/2402.01030},
}

@inproceedings{osworld,
  title = {{OSWorld}: Benchmarking Multimodal Agents for Open-Ended Tasks in Real Computer Environments},
  author = {Xie, Tianbao and Zhang, Danyang and Chen, Jixuan and Li, Xiaochuan and Zhao, Siheng and Cao, Ruisheng and Hua, Toh Jing and Cheng, Zhoujun and Shin, Dongchan and Lei, Fangyu and Liu, Yitao and Xu, Yiheng and Zhou, Shuyan and Savarese, Silvio and Xiong, Caiming and Zhong, Victor and Yu, Tao},
  booktitle = {Advances in Neural Information Processing Systems},
  volume = {37},
  note = {Datasets and Benchmarks Track},
  year = {2024},
  doi = {10.52202/079017-1650},
  url = {https://papers.nips.cc/paper_files/paper/2024/hash/5d413e48f84dc61244b6be550f1cd8f5-Abstract-Datasets_and_Benchmarks_Track.html},
}

@misc{osworldverifiedweb,
  title = {Introducing {OSWorld-Verified}},
  author = {{XLANG Lab}},
  year = {2025},
  howpublished = {\url{https://xlang.ai/blog/osworld-verified}},
  note = {Accessed: 2026-05-26},
}

@inproceedings{agents,
  title = {Agent {S}: An Open Agentic Framework that Uses Computers Like a Human},
  author = {Agashe, Saaket and Han, Jiuzhou and Gan, Shuyu and Yang, Jiachen and Li, Ang and Wang, Xin},
  booktitle = {International Conference on Learning Representations},
  year = {2025},
  url = {https://proceedings.iclr.cc/paper_files/paper/2025/hash/394c7c30ea87b5c3521b4d9e9d419071-Abstract-Conference.html},
}

@misc{agents2,
  title = {Agent {S2}: A Compositional Generalist-Specialist Framework for Computer Use Agents},
  author = {Agashe, Saaket and Wong, Kyle and Tu, Vincent and Yang, Jiachen and Li, Ang and Wang, Xin Eric},
  year = {2025},
  eprint = {2504.00906},
  archivePrefix = {arXiv},
  primaryClass = {cs.AI},
  doi = {10.48550/arXiv.2504.00906},
  url = {https://arxiv.org/abs/2504.00906},
}

@misc{guide,
  title = {{GUIDE}: Resolving Domain Bias in {GUI} Agents through Real-Time Web Video Retrieval and Plug-and-Play Annotation},
  author = {Xie, Rui and Gao, Zhi and Shi, Chenrui and Shang, Zirui and Chen, Lu and Li, Qing},
  year = {2026},
  eprint = {2603.26266},
  archivePrefix = {arXiv},
  primaryClass = {cs.AI},
  doi = {10.48550/arXiv.2603.26266},
  url = {https://arxiv.org/abs/2603.26266},
  note = {Accepted to ECCV 2026},
}

@misc{ufo,
  title = {{UFO}: A {UI}-Focused Agent for {Windows} {OS} Interaction},
  author = {Zhang, Chaoyun and Li, Liqun and He, Shilin and Zhang, Xu and Qiao, Bo and Qin, Si and Ma, Minghua and Kang, Yu and Lin, Qingwei and Rajmohan, Saravan and Zhang, Dongmei and Zhang, Qi},
  year = {2024},
  eprint = {2402.07939},
  archivePrefix = {arXiv},
  primaryClass = {cs.AI},
  doi = {10.48550/arXiv.2402.07939},
  url = {https://arxiv.org/abs/2402.07939},
}

@inproceedings{cradle,
  title = {{Cradle}: Empowering Foundation Agents towards General Computer Control},
  author = {Tan, Weihao and Zhang, Wentao and Xu, Xinrun and Xia, Haochong and Ding, Ziluo and Li, Boyu and Zhou, Bohan and Yue, Junpeng and Jiang, Jiechuan and Li, Yewen and An, Ruyi and Qin, Molei and Zong, Chuqiao and Zheng, Longtao and Wu, Yujie and Chai, Xiaoqiang and Bi, Yifei and Xie, Tianbao and Gu, Pengjie and Li, Xiyun and Zhang, Ceyao and Tian, Long and Wang, Chaojie and Wang, Xinrun and Karlsson, B{\"o}rje F. and An, Bo and Yan, Shuicheng and Lu, Zongqing},
  booktitle = {Proceedings of the 42nd International Conference on Machine Learning},
  pages = {58658--58725},
  year = {2025},
  volume = {267},
  series = {Proceedings of Machine Learning Research},
  publisher = {PMLR},
  url = {https://proceedings.mlr.press/v267/tan25h.html},
}

@misc{uitars,
  title = {{UI-TARS}: Pioneering Automated {GUI} Interaction with Native Agents},
  author = {Qin, Yujia and Ye, Yining and Fang, Junjie and Wang, Haoming and Liang, Shihao and Tian, Shizuo and Zhang, Junda and Li, Jiahao and Li, Yunxin and Huang, Shijue and Zhong, Wanjun and Li, Kuanye and Yang, Jiale and Miao, Yu and Lin, Woyu and Liu, Longxiang and Jiang, Xu and Ma, Qianli and Li, Jingyu and Xiao, Xiaojun and Cai, Kai and Li, Chuang and Zheng, Yaowei and Jin, Chaolin and Li, Chen and Zhou, Xiao and Wang, Minchao and Chen, Haoli and Li, Zhaojian and Yang, Haihua and Liu, Haifeng and Lin, Feng and Peng, Tao and Liu, Xin and Shi, Guang},
  year = {2025},
  eprint = {2501.12326},
  archivePrefix = {arXiv},
  primaryClass = {cs.AI},
  doi = {10.48550/arXiv.2501.12326},
  url = {https://arxiv.org/abs/2501.12326},
}

@misc{opencua,
  title = {{OpenCUA}: Open Foundations for Computer-Use Agents},
  author = {Wang, Xinyuan and Wang, Bowen and Lu, Dunjie and Yang, Junlin and Xie, Tianbao and Wang, Junli and Deng, Jiaqi and Guo, Xiaole and Xu, Yiheng and Wu, Chen Henry and Shen, Zhennan and Li, Zhuokai and Li, Ryan and Li, Xiaochuan and Chen, Junda and Zheng, Boyuan and Li, Peihang and Lei, Fangyu and Cao, Ruisheng and Fu, Yeqiao and Shin, Dongchan and Shin, Martin and Hu, Jiarui and Wang, Yuyan and Chen, Jixuan and Ye, Yuxiao and Zhang, Danyang and Du, Dikang and Hu, Hao and Chen, Huarong and Zhou, Zaida and Yao, Haotian and Chen, Ziwei and Gu, Qizheng and Wang, Yipu and Wang, Heng and Yang, Diyi and Zhong, Victor and Sung, Flood and {Y. Charles} and Yang, Zhilin and Yu, Tao},
  year = {2025},
  eprint = {2508.09123},
  archivePrefix = {arXiv},
  primaryClass = {cs.AI},
  doi = {10.48550/arXiv.2508.09123},
  url = {https://arxiv.org/abs/2508.09123},
}

@inproceedings{cogagent,
  title = {{CogAgent}: A Visual Language Model for {GUI} Agents},
  author = {Hong, Wenyi and Wang, Weihan and Lv, Qingsong and Xu, Jiazheng and Yu, Wenmeng and Ji, Junhui and Wang, Yan and Wang, Zihan and Dong, Yuxiao and Ding, Ming and Tang, Jie},
  booktitle = {Proceedings of the IEEE/CVF Conference on Computer Vision and Pattern Recognition},
  pages = {14281--14290},
  month = {June},
  year = {2024},
  url = {https://openaccess.thecvf.com/content/CVPR2024/html/Hong_CogAgent_A_Visual_Language_Model_for_GUI_Agents_CVPR_2024_paper.html},
}

@inproceedings{seeclick,
  title = {{SeeClick}: Harnessing {GUI} Grounding for Advanced Visual {GUI} Agents},
  author = {Cheng, Kanzhi and Sun, Qiushi and Chu, Yougang and Xu, Fangzhi and YanTao, Li and Zhang, Jianbing and Wu, Zhiyong},
  booktitle = {Proceedings of the 62nd Annual Meeting of the Association for Computational Linguistics (Volume 1: Long Papers)},
  pages = {9313--9332},
  year = {2024},
  month = {August},
  address = {Bangkok, Thailand},
  publisher = {Association for Computational Linguistics},
  doi = {10.18653/v1/2024.acl-long.505},
  url = {https://aclanthology.org/2024.acl-long.505/},
}

@inproceedings{showui,
  title = {{ShowUI}: One Vision-Language-Action Model for {GUI} Visual Agent},
  author = {Lin, Kevin Qinghong and Li, Linjie and Gao, Difei and Yang, Zhengyuan and Wu, Shiwei and Bai, Zechen and Lei, Stan Weixian and Wang, Lijuan and Shou, Mike Zheng},
  booktitle = {Proceedings of the IEEE/CVF Conference on Computer Vision and Pattern Recognition},
  pages = {19498--19508},
  month = {June},
  year = {2025},
  url = {https://openaccess.thecvf.com/content/CVPR2025/html/Lin_ShowUI_One_Vision-Language-Action_Model_for_GUI_Visual_Agent_CVPR_2025_paper.html},
}

@misc{setofmark,
  title = {Set-of-Mark Prompting Unleashes Extraordinary Visual Grounding in {GPT-4V}},
  author = {Yang, Jianwei and Zhang, Hao and Li, Feng and Zou, Xueyan and Li, Chunyuan and Gao, Jianfeng},
  year = {2023},
  eprint = {2310.11441},
  archivePrefix = {arXiv},
  primaryClass = {cs.CV},
  doi = {10.48550/arXiv.2310.11441},
  url = {https://arxiv.org/abs/2310.11441},
}

@misc{omniparser,
  title = {{OmniParser} for Pure Vision Based {GUI} Agent},
  author = {Lu, Yadong and Yang, Jianwei and Shen, Yelong and Awadallah, Ahmed},
  year = {2024},
  eprint = {2408.00203},
  archivePrefix = {arXiv},
  primaryClass = {cs.CV},
  doi = {10.48550/arXiv.2408.00203},
  url = {https://arxiv.org/abs/2408.00203},
}

@inproceedings{guisurvey,
  title = {{GUI} Agents: A Survey},
  author = {Nguyen, Dang and Chen, Jian and Wang, Yu and Wu, Gang and Park, Namyong and Hu, Zhengmian and Lyu, Hanjia and Wu, Junda and Aponte, Ryan and Xia, Yu and Li, Xintong and Shi, Jing and Chen, Hongjie and Lai, Viet Dac and Xie, Zhouhang and Kim, Sungchul and Zhang, Ruiyi and Yu, Tong and Tanjim, Mehrab and Ahmed, Nesreen K. and Mathur, Puneet and Yoon, Seunghyun and Yao, Lina and Kveton, Branislav and Kil, Jihyung and Nguyen, Thien Huu and Bui, Trung and Zhou, Tianyi and Rossi, Ryan A. and Dernoncourt, Franck},
  booktitle = {Findings of the Association for Computational Linguistics: ACL 2025},
  pages = {22522--22538},
  year = {2025},
  month = {July},
  address = {Vienna, Austria},
  publisher = {Association for Computational Linguistics},
  doi = {10.18653/v1/2025.findings-acl.1158},
  url = {https://aclanthology.org/2025.findings-acl.1158/},
}

@inproceedings{axis,
  title = {{AXIS}: Efficient Human-Agent-Computer Interaction with {API}-First {LLM}-Based Agents},
  author = {Lu, Junting and Zhang, Zhiyang and Yang, Fangkai and Zhang, Jue and Wang, Lu and Du, Chao and Lin, Qingwei and Rajmohan, Saravan and Zhang, Dongmei and Zhang, Qi},
  booktitle = {Proceedings of the 63rd Annual Meeting of the Association for Computational Linguistics (Volume 1: Long Papers)},
  pages = {7711--7743},
  year = {2025},
  month = {July},
  address = {Vienna, Austria},
  publisher = {Association for Computational Linguistics},
  doi = {10.18653/v1/2025.acl-long.381},
  url = {https://aclanthology.org/2025.acl-long.381/},
}

@inproceedings{beyondbrowsing,
  title = {Beyond Browsing: {API}-Based Web Agents},
  author = {Song, Yueqi and Xu, Frank F. and Zhou, Shuyan and Neubig, Graham},
  booktitle = {Findings of the Association for Computational Linguistics: ACL 2025},
  pages = {11066--11085},
  year = {2025},
  month = {July},
  address = {Vienna, Austria},
  publisher = {Association for Computational Linguistics},
  doi = {10.18653/v1/2025.findings-acl.577},
  url = {https://aclanthology.org/2025.findings-acl.577/},
}

@misc{goi,
  title = {From Imperative to Declarative: Towards {LLM}-friendly {OS} Interfaces for Boosted Computer-Use Agents},
  author = {Wang, Yuan and Li, Mingyu and Chen, Haibo},
  year = {2026},
  eprint = {2510.04607},
  archivePrefix = {arXiv},
  primaryClass = {cs.OS},
  doi = {10.48550/arXiv.2510.04607},
  url = {https://arxiv.org/abs/2510.04607},
}

@misc{ufo2,
  title = {{UFO2}: The Desktop {AgentOS}},
  author = {Zhang, Chaoyun and Huang, He and Ni, Chiming and Mu, Jian and Qin, Si and He, Shilin and Wang, Lu and Yang, Fangkai and Zhao, Pu and Du, Chao and Li, Liqun and Kang, Yu and Jiang, Zhao and Zheng, Suzhen and Wang, Rujia and Qian, Jiaxu and Ma, Minghua and Lou, Jian-Guang and Lin, Qingwei and Rajmohan, Saravan and Zhang, Dongmei},
  year = {2025},
  eprint = {2504.14603},
  archivePrefix = {arXiv},
  primaryClass = {cs.AI},
  doi = {10.48550/arXiv.2504.14603},
  url = {https://arxiv.org/abs/2504.14603},
}

@misc{oscopilot,
  title = {{OS-Copilot}: Towards Generalist Computer Agents with Self-Improvement},
  author = {Wu, Zhiyong and Han, Chengcheng and Ding, Zichen and Weng, Zhenmin and Liu, Zhoumianze and Yao, Shunyu and Yu, Tao and Kong, Lingpeng},
  year = {2024},
  eprint = {2402.07456},
  archivePrefix = {arXiv},
  primaryClass = {cs.AI},
  doi = {10.48550/arXiv.2402.07456},
  url = {https://arxiv.org/abs/2402.07456},
}

@inproceedings{dynasaur,
  title = {{DynaSaur}: Large Language Agents Beyond Predefined Actions},
  author = {Nguyen, Dang and Lai, Viet Dac and Yoon, Seunghyun and Rossi, Ryan A. and Zhao, Handong and Zhang, Ruiyi and Mathur, Puneet and Lipka, Nedim and Wang, Yu and Bui, Trung and Dernoncourt, Franck and Zhou, Tianyi},
  booktitle = {Conference on Language Modeling},
  year = {2025},
  url = {https://openreview.net/forum?id=lv0cJ2pWVd},
}

@article{clianything,
  title = {{CLI-Anything}: Towards Agent-Native Computer Use},
  author = {Yang, Yuhao and Fan, Tianyu and Huang, Chao},
  journal = {arXiv preprint arXiv:2606.03854},
  year = {2026},
  doi = {10.48550/arXiv.2606.03854},
  url = {https://arxiv.org/abs/2606.03854},
}

@misc{opencli,
  title = {{OpenCLI}},
  author = {{jackwener}},
  year = {2026},
  howpublished = {\url{https://github.com/jackwener/opencli}},
  note = {Accessed: 2026-05-26},
}

@misc{claudecode,
  title = {{Claude Code} Documentation: Overview},
  author = {{Anthropic}},
  year = {2026},
  howpublished = {\url{https://code.claude.com/docs/en/overview}},
  note = {Accessed: 2026-05-26},
}

@misc{openaicodex,
  title = {{Codex}: {OpenAI}'s Coding Agent for Software Development},
  author = {{OpenAI}},
  year = {2026},
  howpublished = {\url{https://developers.openai.com/api/docs/guides/code-generation#use-codex}},
  note = {Accessed: 2026-05-26},
}
